\documentclass{fairmeta}
\usepackage{amsmath,amssymb,booktabs,array,tabularx,graphicx,microtype,enumitem,hyperref,url}
\usepackage{wrapfig,tcolorbox,capt-of,placeins,needspace}
\usepackage{pgfplots}
\pgfplotsset{compat=1.18}
\newcommand{\rc}{\textbf{JudgeProfile}}

\title{JudgeProfile: Understanding and Steering Subjectivity in LLM Judges}
\author[1,2,*]{Qi Cao}
\author[1]{Kangning Liu}
\author[1]{Xuan Kan}
\author[1]{Shunwen Tan}
\author[1]{Yang Pei}
\author[1]{Dake Chen}
\author[1,3,*]{Yatai Ji}
\author[1,4,*]{Zixuan Ye}
\author[1,3,*]{Yuanpeng Tu}
\author[1]{Daniel Li}
\author[1]{Junbiao Tang}
\author[2]{Pengtao Xie}
\author[1]{Zihao He}
\affiliation[1]{Meta}
\affiliation[2]{University of California San Diego}
\affiliation[3]{The University of Hong Kong}
\affiliation[4]{The Hong Kong University of Science and Technology}
\contribution[*]{Work done during an internship at Meta}
\abstract{
LLM judges are inherently subjective, often favoring different responses in pairwise comparison when neither option is objectively wrong. 
To study this subjectivity, we introduce \textbf{JudgeProfile}, a framework that dissects LLM evaluation into \textit{perception} (how a judge compares two responses across specific attributes like  clarity, correctness, and detail) and \textit{prioritization} (how much each attribute influences the final choice). 
We curate \textbf{SubjectiveSet}, a dataset of 50,013 response pairs from 17 public data sources, evaluated by 21 LLM judges across 87 attributes. We find a hidden consensus in perception: judges frequently agree on attribute judgments even when their overall choices diverge.
Building on this separation, we first characterize each judge’s prioritization using attribute weights estimated from its own overall choices. These weights differ across judges even when estimated from the same attribute judgments. We then learn new weights from reference labels to adapt their decisions to a target evaluation standard.
Reweighting perceived attributes improves 
average held-out agreement with reference labels from 66.48\% to 71.97\%, outperforming fine-tuning and rubric prompting. 
Our findings show that understanding and steering the subjectivity of LLM judges requires attention not only to what they perceive, but also to how they prioritize it.

}
\date{September 28, 2026}
\begin{document}
\raggedbottom
\maketitle
\section{Introduction}

Large language models (LLMs) are often used to judge which of two responses is better~\citep{zheng2023judging,liu2023geval,kim2024prometheus2}. For open-ended tasks, more than one response can be good. One response may be richer in details, while another gives a short, direct answer. Even if both are correct, different users may prefer different responses~\citep{sorensen2024pluralistic}. We call such differences \emph{subjectivity} when there is no single objectively correct judgment. LLM judges also exhibit such subjectivity, favoring one of several acceptable responses. However, it remains unclear what drives these preferences.

\begin{figure}[t]
\centering
\includegraphics[width=\linewidth]{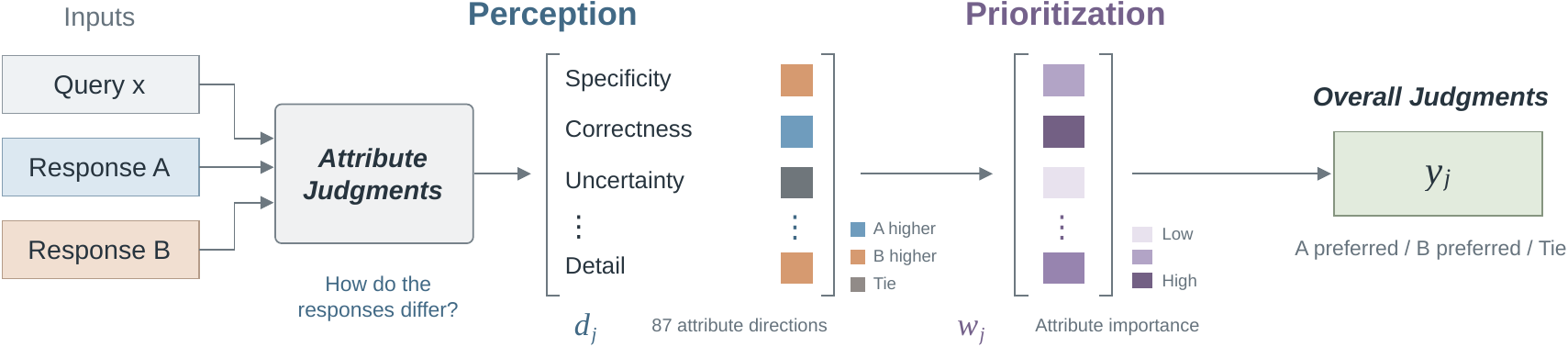}
\caption{
\textbf{Perception and prioritization underlying LLM judgment.}
Given a query and two responses, attribute-level comparisons $\mathbf d_j$ describe
how the responses differ.
Prioritization assigns relative importance to these differences through
weights $\mathbf w_j$, modeling an overall preference $y_j$.
}
\label{fig:decompose}
\end{figure}

We introduce \rc{}, a framework to study what drives these preferences, which decomposes overall judgments into perception and prioritization (Figure~\ref{fig:decompose}). In \emph{perception}, a judge compares the responses on individual qualities, which we call \emph{attributes}, such as clarity, correctness, and detail level. In \emph{prioritization}, it weighs these attributes to reach an overall judgment. Subjectivity between judges may therefore arise from either or both stages: judges may perceive the responses differently, prioritize attributes differently, or both.

To examine both stages at scale, we curate \textbf{SubjectiveSet}, a dataset of 50,013 response pairs from 17 public data sources, each with two responses to the same query. We ask 21 LLM judges to make overall judgments and compare the responses on 87 attributes. Evaluating each attribute separately in both response orders yields 182.7 million inference calls. Building on attribute evaluation and preference analysis~\citep{ye2023flask,wang2023helpsteer,dunlap2024vibecheck,movva2025wimhf}, we study perception by comparing these attribute judgments across judges. For prioritization, we estimate weights that predict each judge's overall choices from its attribute judgments. Larger weights mark more influential attributes.

We find that perception is largely shared. Even when their overall judgments differ, judges agree on which response has more of a given attribute in 85.9\% of the comparisons analyzed. We call this \emph{hidden consensus}. A human study shows the same pattern.

If judges largely agree on attributes, why do they still disagree overall? We next examine whether they also differ in prioritization. Even when all judges' weights are fitted on the same attribute judgments, the weights vary substantially across judges, yet remain reproducible within each judge across data splits. Judges therefore assign different importance to qualities they perceive similarly, identifying prioritization as another source of subjectivity.

These findings suggest a way to adapt judges to different evaluation standards: changing how their existing attribute judgments contribute to overall decisions. We implement this through \emph{reweighting}, which learns new attribute weights from a target dataset's training labels while keeping each model and its attribute judgments fixed. Reweighting raises average agreement with reference labels from 66.48\% to 71.97\% on held-out data, with improvements for all 21 judges. It also works with limited supervision: with 256 labels per source, it outperforms LoRA fine-tuning~\citep{hu2021lora} on all four judges tested. When both methods use six attributes, reweighting outperforms rubric prompting~\citep{roy2026premise,chen2026calibratedrubric} on three judges and performs similarly on the fourth. Beyond adaptation, researchers can inspect attribute judgments and fitted weights to investigate why judges disagree about the same responses.

Our contributions are:
\begin{itemize}[leftmargin=*]
    \item \textbf{A framework and dataset for studying subjectivity.} JudgeProfile separates perception from prioritization. SubjectiveSet supports a large study of LLM judgments and a smaller study of human judgments.
    \item \textbf{Hidden consensus in perception and differences in prioritization.} Judges often agree on attribute judgments despite different overall judgments. Their fitted weights differ even when estimated using the same attribute judgments, providing evidence of differences in prioritization across judges.
    \item \textbf{Lightweight steering of subjectivity.} Learning a small set of attribute weights adapts overall judgments to target preferences without updating the language model.
\end{itemize}

\section{Subjectivity in LLM Judges}

\subsection{What is subjectivity?}
We use \emph{subjectivity} to refer to variation in judgments for which there is no unique objective ground truth. Disagreement alone does not establish subjectivity. For verifiable factual claims, external evidence may determine which judgment is correct. Judgments about whether a response is interesting, sufficiently detailed, or appropriately cautious may instead depend on evaluative standards, with no single objectively correct answer. Subjectivity can therefore appear not only in which response a judge ultimately prefers, but also in how the judge evaluates individual qualities of the responses.

\subsection{Perception and prioritization.}
An overall preference tells us which response a judge favors,
but not why judges disagree. For example, one response may
provide more concrete details, while another expresses more
uncertainty. Judges may disagree about these characteristics, about their relative importance, or both.
We refer to separately assessable characteristics---such as
specificity, factual correctness, and explicit uncertainty---as \emph{attributes} of responses.

Given a query $x$ and two candidate responses $r_A$ and $r_B$, we
introduce an analytical distinction between perception and
prioritization:
\begin{equation}
(x,r_A,r_B)
\xrightarrow{\text{perception}}
\mathbf d_j(x,r_A,r_B)
\xrightarrow{\text{prioritization}}
y_j.
\label{eq:overview}
\end{equation}
\emph{Perception} captures how judge $j$ compares the responses
along individual attributes, represented by $\mathbf d_j$.
These comparisons describe response differences without
directly deciding which response is better overall.
\emph{Prioritization} captures the relative importance assigned
to those differences when forming the overall preference $y_j$.
By eliciting attribute comparisons separately from holistic
verdicts, we ask whether conflicting preferences reflect
different perceptions, or whether different priorities can
explain disagreement even when perceptions are largely shared.

\begin{figure}[!t]
\centering
\includegraphics[width=\linewidth]{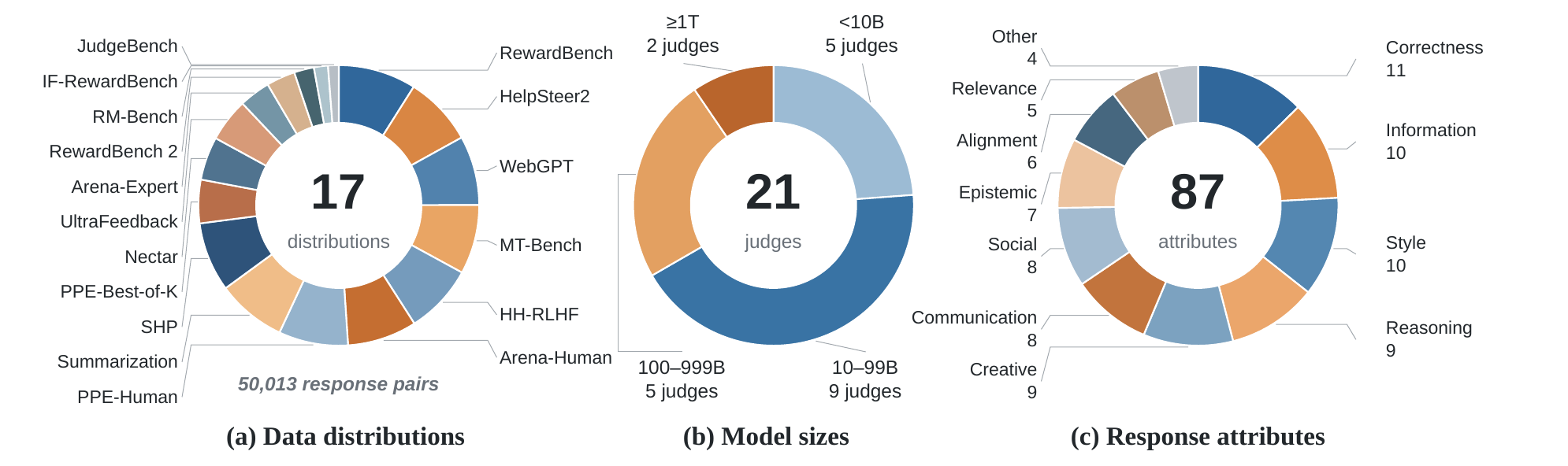}
\caption{
\textbf{Evaluation setup.} 50,013 response pairs from 17 public distributions, 21 LLM judges ranging from 0.8B to 2.8T total parameters, and 87 response attributes. Independent evaluation of each attribute in both response orders yields 182.7 million attribute-level inference calls.
}
\label{fig:scale}

\end{figure}

\subsection{Studying subjectivity at scale.}
We curate \emph{SubjectiveSet}, a dataset comprising 50,013 response pairs from 17 public distributions~\citep{stiennon2020learning,bai2022training,nakano2021webgpt,ethayarajh2022understanding,wang2024helpsteer2,cui2023ultrafeedback,chiang2024arena,lambert2024rewardbench}. For each pair, 21 LLM judges provide overall preferences and separate judgments along 87 attributes. Evaluating each attribute independently in both response orders yields 182.7 million attribute-level inference calls (Figure~\ref{fig:scale}).

SubjectiveSet includes both \emph{preference-labeled} data, based on human or AI preferences, and \emph{benchmark-labeled} data, based on benchmark evaluation criteria. Figure~\ref{fig:label-type-examples} in Appendix~\ref{app:datasets} illustrates both types. A response with a correct final answer may still be harder to read or provide less complete reasoning than the alternative. We use disagreement among LLM judges to select pairs for closer attribute-level analysis.

To examine whether judges agree on attributes despite disagreeing overall, we focus on response pairs with substantial overall disagreement. \emph{SubjectiveSet--LLM} contains 11,524 response pairs (23.0\% of the full collection) on which at least one third of judges giving a definite A/B verdict choose the minority response. To examine whether humans show a similar separation between attribute and overall judgments, we collect \emph{SubjectiveSet--Human}. It contains 23 curated cases evaluated by seven raters: 21 with model disagreement and two controls, covering 55 item--attribute comparisons. The full collection supports broader agreement, priority, and adaptation analyses. Subset definitions and source counts appear in Appendix~\ref{app:jc-subsets}; human cases and judgments appear in Appendix~\ref{app:human-cases}.


\section{Perception and Subjectivity}
\label{sec:hidden-consensus}

\begin{figure}[t]
\centering
\includegraphics[width=1\linewidth]{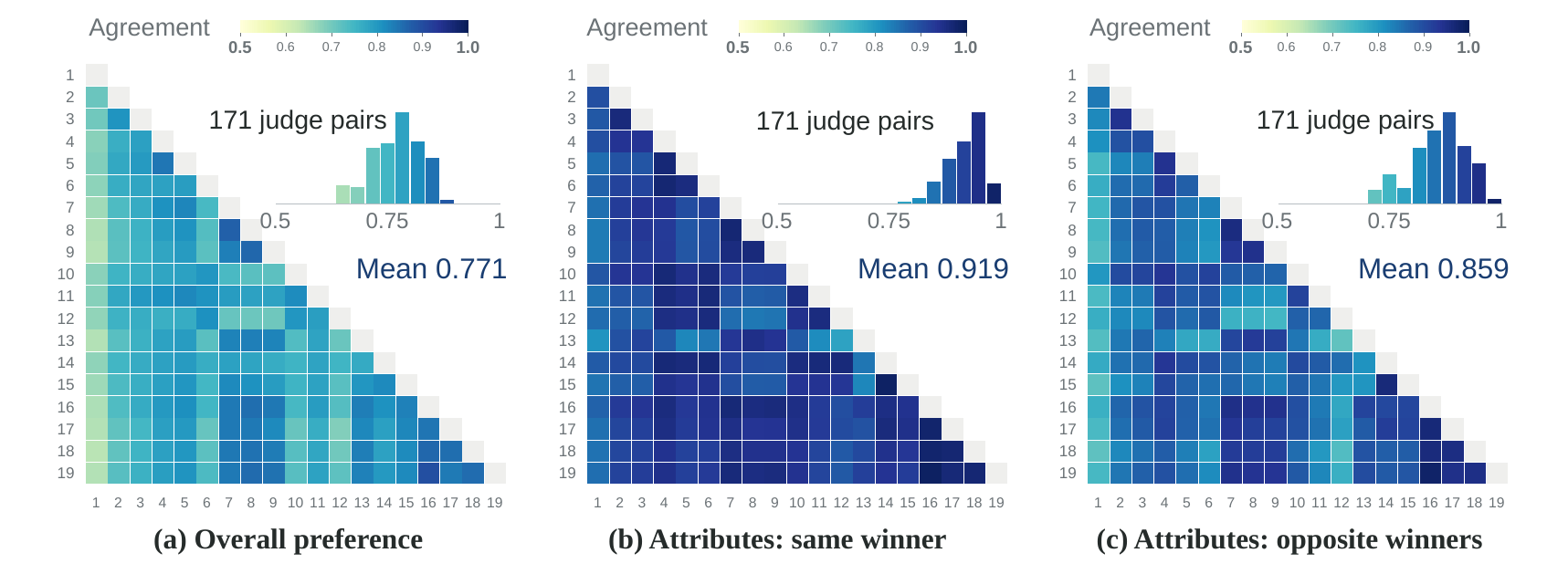}
\caption{
\textbf{Judges agree on attributes despite different overall choices.} Agreement for 171 pairs of 19 judges: overall choices (a), and attributes when winners are the same (b) or different (c). We include an attribute comparison only if each judge makes a definite choice of the same response in both presentation orders. Heatmaps share a color scale; inset distributions have separate vertical scales.
}
\label{fig:hc-strata}
\end{figure}

\begin{figure}[t]
\centering
\includegraphics[width=1\linewidth]{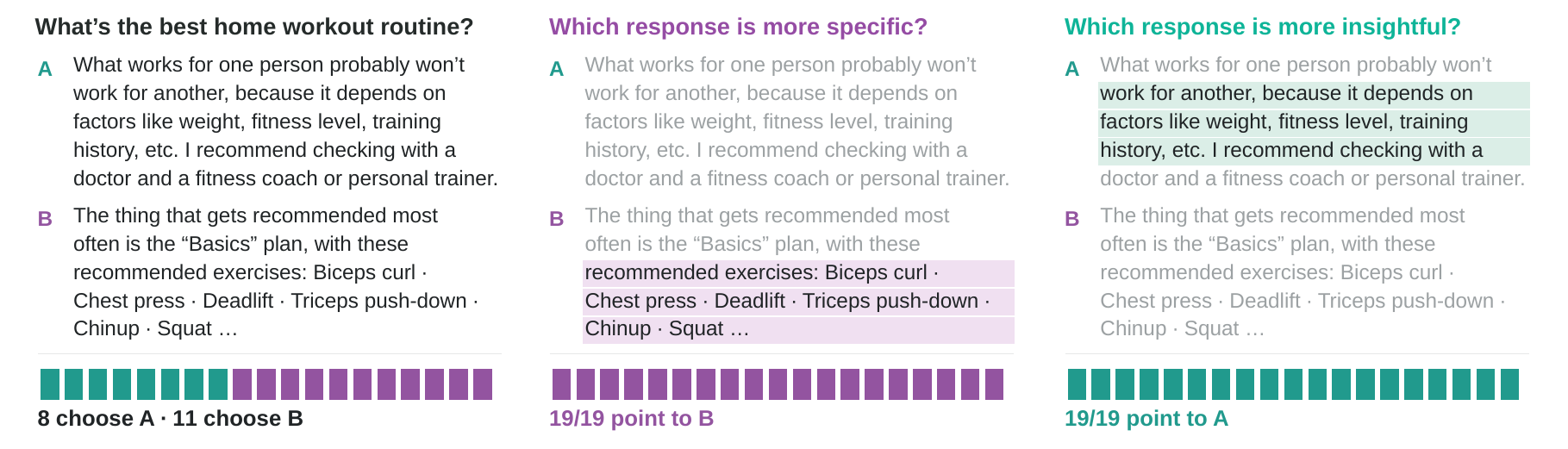}
\caption{
\textbf{Shared attribute judgments, different winners.} Judges agree that workout response B is more specific and A is more insightful, but disagree on which is better overall.
}
\label{fig:hc-example}
\end{figure}

We start with perception: when LLM judges choose different winners, do their attribute judgments also differ?

\begin{tcolorbox}[colback=gray!5,colframe=gray!55,boxrule=0.5pt,arc=2pt,left=6pt,right=6pt,top=5pt,bottom=5pt]
\textbf{Finding 1: Attribute judgments often agree even when overall judgments differ.}
We call this pattern \emph{hidden consensus}. We observe it among LLM judges and in a small human study.
\end{tcolorbox}

\subsection{Evidence 1: Attribute judgments agree even when judges choose different winners.}
Figure~\ref{fig:hc-strata} compares overall agreement with attribute agreement when two judges choose the same or different winners. For judges $j$ and $k$, let $d_j,d_k$ denote their attribute judgments and $y_j,y_k$ their overall judgments:
\begin{equation}
\begin{aligned}
A_{\mathrm{same}} &= \Pr(d_j=d_k\mid y_j=y_k),\\
A_{\mathrm{opposite}} &= \Pr(d_j=d_k\mid y_j\ne y_k).
\end{aligned}
\label{eq:conditional-attribute-agreement}
\end{equation}
After excluding two judges with severe position bias in their attribute judgments (Appendix~\ref{app:judges}), mean attribute agreement among the remaining 19 judges is 91.9\% when they choose the same winner and 85.9\% when they choose different winners (Figure~\ref{fig:hc-strata}). Even with different winners, all 171 pairs exceed the 50\% balanced-random reference. Including all 21 judges gives 83.6\% agreement (Appendix~\ref{app:hc-protocol}). Their attribute judgments therefore often agree despite different overall judgments.

For example, the 19 judges split 8--11 on which home-workout response is better (Figure~\ref{fig:hc-example}). Yet all 19 find B more specific, and all 18 whose insightfulness judgments are consistent across both response orders find A more insightful. Their attribute judgments agree, but their overall judgments differ.

\subsection{Evidence 2: Model and human disagreement cases show the same pattern.}
\begin{figure}[t]
\centering
\setbox0=\hbox{\includegraphics[width=0.49\linewidth]{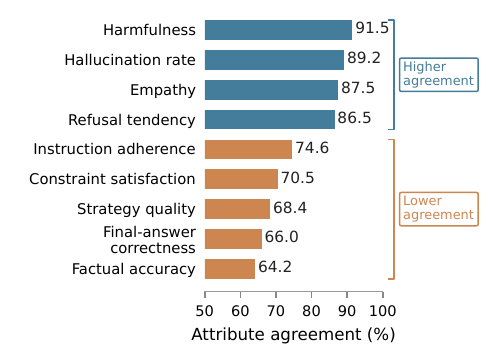}}
\dimen0=\ht0\advance\dimen0 by\dp0
\begin{minipage}[t]{0.48\linewidth}
\vspace{0pt}\centering
\begin{minipage}[t][\dimen0][t]{\linewidth}
\vspace{0pt}\centering
\small
\setlength{\tabcolsep}{3pt}
\renewcommand{\arraystretch}{1.65}
\begin{tabular*}{\linewidth}{@{\extracolsep{\fill}}lrrr@{}}
\toprule
Response pairs & Overall & Attr. & Gap\\
\midrule
\multicolumn{4}{@{}l}{\textit{21 LLM judges}}\\
Full collection & 74.0 & 88.2 & +14.2\\
\textbf{SubjectiveSet--LLM} & \textbf{49.8} & \textbf{84.3} & \textbf{+34.5}\\
\midrule
\multicolumn{4}{@{}l}{\textit{7 human raters}}\\
All cases (23) & 62.1 & 92.3 & +30.2\\
\textbf{SubjectiveSet--Human} & \textbf{57.6} & \textbf{92.8} & \textbf{+35.2}\\
\bottomrule
\end{tabular*}
\end{minipage}
\captionof{table}{\textbf{Attribute agreement on disagreement cases.} Pairwise agreement (\%); Gap = attribute minus overall agreement (pp). Bold rows are disagreement subsets. Details: Appendix~\ref{app:jc-subsets}.}
\label{tab:subjectivity-summary}
\end{minipage}\hfill
\begin{minipage}[t]{0.49\linewidth}
\vspace{0pt}\centering
\begin{minipage}[t][\dimen0][t]{\linewidth}
\vspace{0pt}\centering
\includegraphics[width=\linewidth]{attribute_agreement_selected9.pdf}
\end{minipage}
\captionof{figure}{\textbf{Agreement between attribute judgments.} Nine attributes selected from the highest quartile of fitted weight magnitudes. Details: Appendix~\ref{app:attribute-consensus-detail}.}
\label{fig:attribute-agreement-selected}
\end{minipage}
\end{figure}

SubjectiveSet--LLM contains response pairs on which the judges disagree overall. Across all 21 judges, overall agreement on these pairs is 49.8\%, while attribute agreement is 84.3\% (Table~\ref{tab:subjectivity-summary}). Low overall agreement is expected because of how the pairs were selected; attribute agreement nevertheless remains high.

The selected human cases show a similar pattern. On the 21 disagreement cases in SubjectiveSet--Human, seven raters have 57.6\% overall agreement and 92.8\% attribute agreement. These cases show that hidden consensus also occurs in human judgments.

\subsection{Discussion: Attribute judgments can still be subjective.}
Agreement between attribute judgments varies by attribute (Figure~\ref{fig:attribute-agreement-selected}). We fit logistic regression models to predict each judge's overall judgments from attribute judgments, learning one weight per attribute (Section~\ref{sec:fingerprint}). We measure attribute importance using the absolute values of these weights. The figure shows nine attributes selected from the highest importance quartile. When judges choose different winners, agreement is 66.0\% for final-answer correctness, 68.4\% for strategy quality, and 64.2\% for factual accuracy. By comparison, agreement is 91.5\% for harmfulness and 89.2\% for hallucination rate. The high average therefore hides substantial disagreement on some important attributes.

\section{Prioritization and Subjectivity}
\label{sec:fingerprint}\label{sec:different-weights}
The hidden consensus in Section~\ref{sec:hidden-consensus} leaves an important question: why do judges still disagree overall when their attribute judgments agree? We examine prioritization as another source of subjectivity, then test whether changing attribute weights can steer subjectivity.

\begin{tcolorbox}[colback=gray!5,colframe=gray!55,boxrule=0.5pt,arc=2pt,left=6pt,right=6pt,top=5pt,bottom=5pt]
\textbf{Finding 2: Prioritization is another source of subjectivity.}
Judges can agree on attribute judgments yet choose different winners. Their fitted priorities also differ, even when they use the same attribute representation.
\end{tcolorbox}

\subsection{Evidence 1: Shared attribute judgments do not guarantee the same winner.}
Across all 87 attributes, we select pairs of overall judgments with no conflict in the observed attribute judgments: both judges give definite overall judgments, each has a definite attribute judgment consistent across both response orders for at least 20 shared attributes, and the two judges agree on all of those attributes. Winners still differ on 14.5\% of these cases (Figure~\ref{fig:consensus-limits}). Coverage and sensitivity analyses are in Appendix~\ref{app:no-conflict-detail}.

\subsection{Evidence 2: Fitted priorities differ across judges and are reproducible within judges.}
\begin{figure}[t]
\centering
\begin{minipage}[t]{0.388\linewidth}
\vspace{0pt}\centering
\begin{minipage}[c][4.2cm][c]{\linewidth}
\centering
\resizebox{\linewidth}{!}{%
\begin{tikzpicture}[x=1cm,y=1cm,font=\sffamily\small] \useasboundingbox (0,0) rectangle (8,5.3); \node[anchor=west,text=black!55,font=\sffamily\footnotesize] at (0,5.1) {Illustration}; \node[font=\sffamily\small] at (3.3,4.7) {Attribute judgments}; \node[font=\sffamily\small] at (6.9,4.7) {Winner}; \node[anchor=west] at (0,4.12) {Judge 1}; \node[anchor=west] at (0,3.42) {Judge 2}; \foreach \yy in {4.12,3.42} { \node[fill=blue!12,draw=blue!40,minimum width=0.46cm,minimum height=0.43cm,inner sep=0pt] at (2.05,\yy) {A}; \node[fill=orange!18,draw=orange!60,minimum width=0.46cm,minimum height=0.43cm,inner sep=0pt] at (2.75,\yy) {B}; \node[fill=blue!12,draw=blue!40,minimum width=0.46cm,minimum height=0.43cm,inner sep=0pt] at (3.45,\yy) {A}; \node at (4.12,\yy) {\ldots}; \node[fill=orange!18,draw=orange!60,minimum width=0.46cm,minimum height=0.43cm,inner sep=0pt] at (4.8,\yy) {B}; \draw[->,thick,black!45] (5.25,\yy)--(6.3,\yy); } \node[fill=blue!12,draw=blue!50,minimum width=0.64cm,minimum height=0.46cm,inner sep=0pt,font=\sffamily\bfseries] at (6.9,4.12) {A}; \node[fill=orange!18,draw=orange!60,minimum width=0.64cm,minimum height=0.46cm,inner sep=0pt,font=\sffamily\bfseries] at (6.9,3.42) {B}; \draw[black!15] (0,2.95)--(8,2.95); \node[anchor=west,font=\sffamily\small\bfseries] at (0,2.58) {Different winners (\%)}; \node[anchor=west] at (0,1.95) {No conflict}; \node[anchor=west] at (0,1.12) {Any conflict}; \fill[orange!80!black] (2.65,1.73) rectangle (5.434,2.17); \fill[black!30] (2.65,0.90) rectangle (6.8356,1.34); \node[anchor=west,font=\sffamily\bfseries] at (5.55,1.95) {14.5}; \node[anchor=west] at (6.95,1.12) {21.8}; \draw[black!45] (2.65,0.63)--(7.45,0.63); \foreach \xx/\lab in {2.65/0,4.57/10,6.49/20} { \draw[black!45] (\xx,0.63)--(\xx,0.54); \node[anchor=north,font=\sffamily\footnotesize,text=black!60] at (\xx,0.48) {\lab}; } \end{tikzpicture}}
\end{minipage}
\captionof{figure}{\textbf{Shared attribute judgments, different winners.} Identical attribute judgments can accompany different overall judgments. Appendix~\ref{app:no-conflict-detail}.}
\label{fig:consensus-limits}
\end{minipage}\hfill
\begin{minipage}[t]{0.582\linewidth}
\vspace{0pt}\centering
\begin{minipage}[c][4.2cm][c]{\linewidth}
\centering
\includegraphics[width=\linewidth]{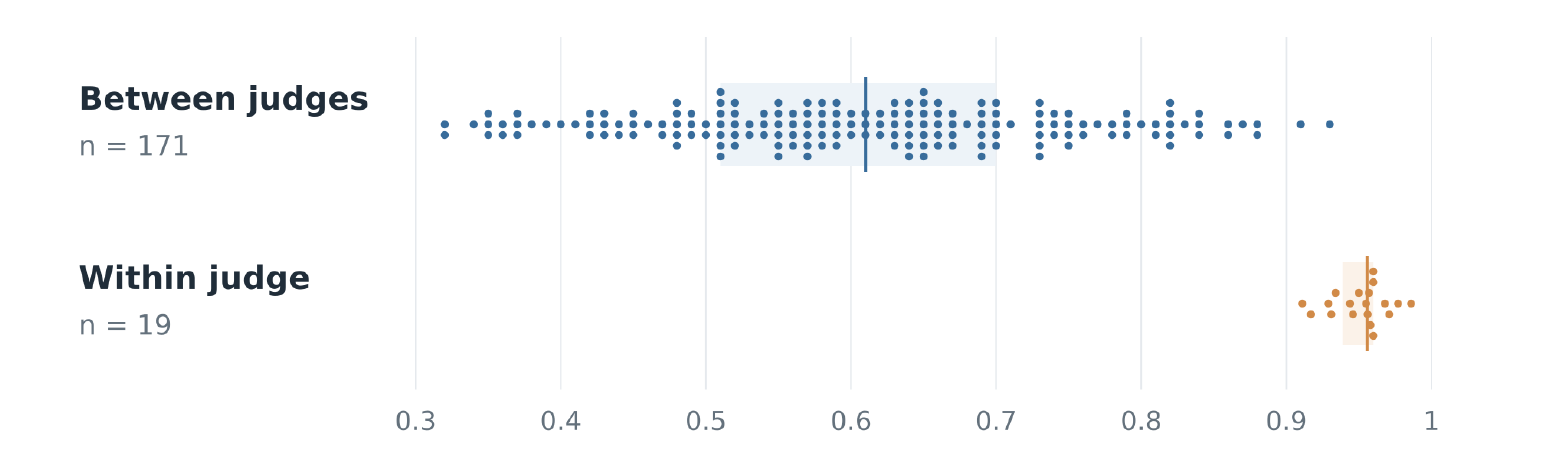}
\end{minipage}
\captionof{figure}{\textbf{Different, reproducible priorities.} Between-judge similarity and within-judge reliability. Perception profiles measure attribute resolution rates. Appendix~\ref{app:hc-priorities}.}
\label{fig:vectors}
\end{minipage}
\end{figure}

We fit each judge's overall judgments from its attribute judgments:
\begin{equation}
\Pr\!\left(y_j=A\mid\mathbf d_j\right)
\approx \sigma\!\left(b_j+\mathbf w_j^{\mathsf T}\mathbf d_j\right),
\label{eq:priority-model}
\end{equation}
where $\sigma$ is the logistic function, $b_j$ an intercept, and $\mathbf w_j$ the fitted priorities. These weights describe the judge's behavior, not its internal computation. They are reproducible within judges (corrected split-half reliability 0.911--0.986) but differ across judges (mean Pearson correlation 0.609; Figure~\ref{fig:vectors}). When all judges use the same attribute representation, their fitted weights still differ (correlation 0.575), with similar prediction accuracy (0.820 versus 0.822). Differences in fitted priorities therefore remain after fixing the attribute representation.

\begin{tcolorbox}[colback=gray!5,colframe=gray!55,boxrule=0.5pt,arc=2pt,left=6pt,right=6pt,top=5pt,bottom=5pt]
\textbf{Finding 3: Priorities can be adjusted explicitly.}
Reweighting directly changes how attribute judgments contribute to an overall judgment. Fine-tuning and rubric prompting can change overall judgments, but do not directly set these weights.
\end{tcolorbox}

For all three experiments, we split the full SubjectiveSet into 75\% training and 25\% test sets on every source. Reweighting, fine-tuning, and rubric construction use only the training set. All results are reported on the held-out test set.

\subsection{Experiment 1: Reweighting fixed attribute judgments.}
\label{sec:weight-only-alignment}
The preceding fits predict each judge's own overall judgments. For adaptation, we instead fit 87 weights to external reference labels for each distribution, keeping the model and its attribute judgments fixed.

\begin{figure}[t]
\centering
\includegraphics[width=\linewidth]{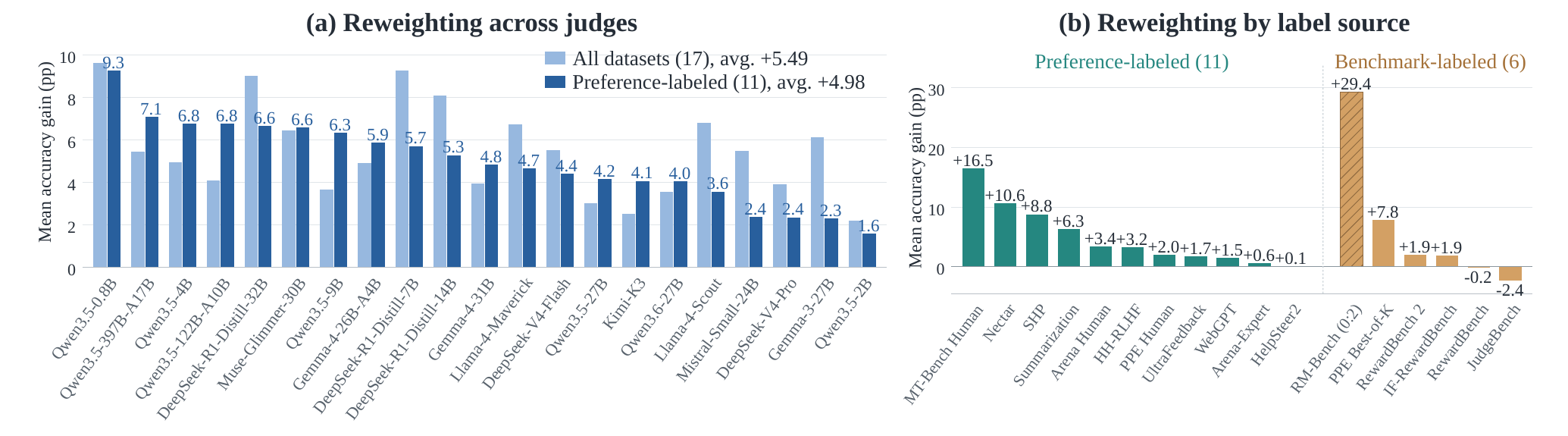}
\caption{\textbf{Reweighting improves alignment across judges and distributions.}
(a) Mean accuracy gains over all 17 datasets and the 11
preference-labeled datasets; judges are ordered by the latter gain.
(b) Mean gains across 21 judges for each dataset.}
\label{fig:reweighting}
\end{figure}

Mean accuracy rises from 66.48\% to 71.97\% and all 21 judges improve (Figure~\ref{fig:reweighting}). Changing the weights alone can therefore improve alignment with a specified reference. On our RM-Bench pairs, the clear style differences make reference labels highly predictable from attribute judgments, consistent with the near-perfect agreement achieved by reweighting for several judges. Excluding RM-Bench, the mean gain remains 3.99 percentage points across the other 16 datasets (Appendix~\ref{app:alignment}).

\subsection{Experiment 2: Reweighting versus fine-tuning.}
\label{sec:sft-comparison}
\begin{figure}[t]
\centering
\includegraphics[width=\linewidth,trim=0 36bp 0 0,clip]{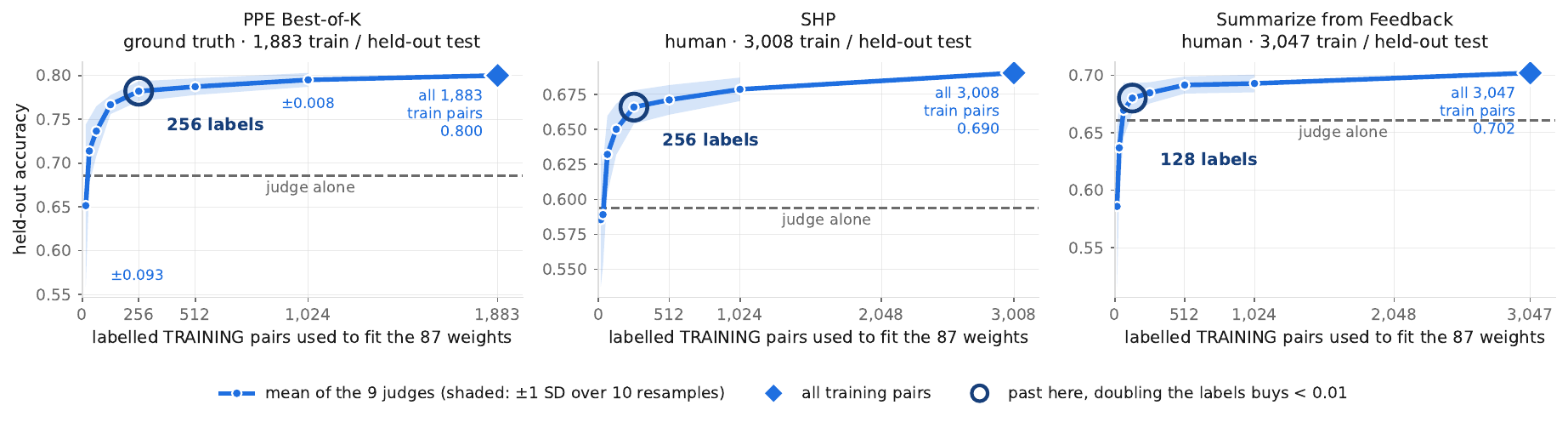}
\caption{\textbf{Useful priorities can be learned from few labels.} Held-out accuracy on three datasets. Blue curves average judges; shading shows $\pm$ standard deviation over ten training-subset resamples. Dashed lines show unadapted judges; diamonds use all training pairs. Open circles mark the first budget after which every later tested doubling gains less than 0.01 accuracy.}
\label{fig:sample-efficiency-main}
\end{figure}

We compare reweighting with LoRA SFT (Table~\ref{tab:sft-vs-reweighting}). At 256 labels per source, reweighting has higher accuracy on all four judges. For three judges, 32 labels already exceed SFT-256. The separate nine-judge experiment shows that gains flatten after 128--256 labels on the three tested distributions (Figure~\ref{fig:sample-efficiency-main}).

One possible explanation is that overall labels specify which response to prefer, but not which attribute judgments or priorities should change. With limited labels, SFT has little evidence to guide these changes. Reweighting narrows the learning problem by fixing attribute judgments and fitting only priorities. With more labels, SFT may benefit from changes beyond what fixed attribute judgments and linear weights can express, consistent with its stronger full-data performance on three of four judges.

\begin{table}[t]
\centering
\caption{\textbf{Reweighting versus fine-tuning.} Accuracy is in percent; $\Delta$ is the gain in percentage points. Run details and parsing rules: Appendix~\ref{app:alignment}.\newline}

\label{tab:sft-vs-reweighting}
\small
\setlength{\tabcolsep}{3.5pt}
\renewcommand{\arraystretch}{1.1}
\begin{tabular}{lrrrrrrrr}
\toprule
& \multicolumn{2}{c}{\textbf{Qwen3.5-4B}} & \multicolumn{2}{c}{\textbf{Qwen3.5-9B}} & \multicolumn{2}{c}{\textbf{Qwen3.5-27B}} & \multicolumn{2}{c}{\textbf{Llama-4-Scout}} \\
\cmidrule(lr){2-3}\cmidrule(lr){4-5}\cmidrule(lr){6-7}\cmidrule(lr){8-9}
\textbf{Method} & \textbf{Acc.} & $\Delta$ & \textbf{Acc.} & $\Delta$ & \textbf{Acc.} & $\Delta$ & \textbf{Acc.} & $\Delta$ \\
\midrule
Unadapted & 63.79 & -- & 66.74 & -- & 70.79 & -- & 66.91 & -- \\
\midrule
SFT-256 & 65.70 & +1.91 & 66.43 & -0.31 & 70.96 & +0.17 & 64.81 & -2.10 \\
Reweighting-32 & 66.02 & +2.23 & 67.21 & +0.47 & 69.96 & -0.83 & 66.67 & -0.24 \\
Reweighting-64 & 65.98 & +2.18 & 67.83 & +1.09 & 70.55 & -0.25 & 68.35 & +1.44 \\
Reweighting-128 & 66.72 & +2.93 & 68.24 & +1.50 & 71.87 & +1.08 & 68.98 & +2.07 \\
Reweighting-256 & \textbf{67.78} & \textbf{+3.99} & \textbf{70.80} & \textbf{+4.06} & \textbf{72.46} & \textbf{+1.67} & \textbf{70.16} & \textbf{+3.25} \\
\midrule
SFT-Full & \textbf{72.94} & \textbf{+9.15} & \textbf{73.50} & \textbf{+6.76} & \textbf{75.65} & \textbf{+4.86} & 68.48 & +1.57 \\
Reweighting-Full & 69.49 & +5.69 & 71.58 & +4.84 & 74.29 & +3.49 & \textbf{72.04} & \textbf{+5.13} \\
\bottomrule
\end{tabular}
\end{table}

\begin{table}[t]
\centering
\caption{\textbf{Rubric prompting versus reweighting.} For each judge and source, reweighting uses the $k$ attributes with the largest absolute weights from the full 87-attribute training fit; $k=6$ uses the same attributes as the rubric. Bold and underlined values mark the best and second-best accuracy and gains for each judge. Weight alignment is cosine similarity to the full 87-attribute target, expressed as a percentage (100 for the full fit by definition). Linear explainability measures how well fixed attribute judgments predict overall judgments. Protocol: Appendix~\ref{app:alignment}.\newline}
\label{tab:rubric-comparison}
\small
\setlength{\tabcolsep}{4pt}
\renewcommand{\arraystretch}{1.1}
\begin{tabular}{llrrrr}
\toprule
\textbf{Judge} & \textbf{Method} & \textbf{Accuracy} & $\Delta$ & \shortstack{\textbf{Weight}\\\textbf{alignment}} & \shortstack{\textbf{Linear}\\\textbf{explainability}} \\
\midrule
Qwen3.5-4B & Unadapted & 63.79 & -- & 58.89 & 77.10 \\
           & Rubric prompting (6) & 66.83 & +3.03 & 54.20 & 74.20 \\
           & Reweighting (6) & 68.58 & +4.79 & 57.30 & -- \\
           & Reweighting (12) & \underline{69.15} & \underline{+5.36} & 67.12 & -- \\
           & Reweighting (87) & \textbf{69.49} & \textbf{+5.69} & 100.00 & -- \\
\midrule
Qwen3.5-9B & Unadapted & 66.74 & -- & 60.46 & 80.18 \\
           & Rubric prompting (6) & 68.14 & +1.40 & 74.26 & 76.73 \\
           & Reweighting (6) & 70.94 & +4.20 & 58.58 & -- \\
           & Reweighting (12) & \underline{71.25} & \underline{+4.50} & 70.47 & -- \\
           & Reweighting (87) & \textbf{71.58} & \textbf{+4.84} & 100.00 & -- \\
\midrule
Qwen3.5-27B & Unadapted & 70.79 & -- & 51.21 & 84.58 \\
            & Rubric prompting (6) & 73.33 & +2.53 & 71.53 & 81.99 \\
            & Reweighting (6) & 73.26 & +2.46 & 57.12 & -- \\
            & Reweighting (12) & \underline{74.21} & \underline{+3.42} & 69.57 & -- \\
            & Reweighting (87) & \textbf{74.29} & \textbf{+3.49} & 100.00 & -- \\
\midrule
Llama-4-Scout & Unadapted & 66.91 & -- & 69.89 & 79.38 \\
              & Rubric prompting (6) & 68.07 & +1.16 & 71.87 & 76.17 \\
              & Reweighting (6) & 70.69 & +3.78 & 59.85 & -- \\
              & Reweighting (12) & \underline{71.59} & \underline{+4.67} & 72.58 & -- \\
              & Reweighting (87) & \textbf{72.04} & \textbf{+5.13} & 100.00 & -- \\
\bottomrule
\end{tabular}
\end{table}

\subsection{Experiment 3: Reweighting versus rubric prompting.}
\label{sec:rubric-comparison}
Rubrics specify evaluation criteria but leave the judge to apply them~\citep{roy2026premise,chen2026calibratedrubric}. For each judge and source, we select the six attributes with the largest absolute weights from the 87-attribute fit to reference labels on the full training set. We use these same six attributes as rubric criteria and refit their weights on the full training set for six-attribute reweighting. The attribute selection uses no test labels, and both methods are evaluated on the same held-out test set. With six attributes in both methods, reweighting has higher accuracy on three judges and is lower by only 0.07 percentage points on Qwen3.5-27B (Table~\ref{tab:rubric-comparison}). The table also reports 12- and 87-attribute fits to show the effect of adding attributes.

Matching the number of attributes does not make the two methods equivalent. A rubric names criteria, but leaves the judge to determine their relative importance and resolve conflicts between them. Reweighting instead fixes attribute judgments and directly fits their contribution to the overall judgment.

Weight alignment and linear explainability help characterize this difference. Weight alignment measures whether fitted weights move toward the target weights. On Qwen3.5-4B, rubric prompting improves accuracy while reducing weight alignment, showing that better predictions need not imply closer weight alignment. Linear explainability measures how well the original attribute judgments predict the adapted judge's overall judgments. It decreases after rubric prompting on all four judges, suggesting that the original attribute judgments become less sufficient to explain the new choices through a linear model. Together, these results show that rubrics do not provide direct control over attribute weights.

\subsection{Discussion: Which attributes matter varies by the judge and dataset.}
\begin{figure}[t]
\centering
\includegraphics[width=\linewidth]{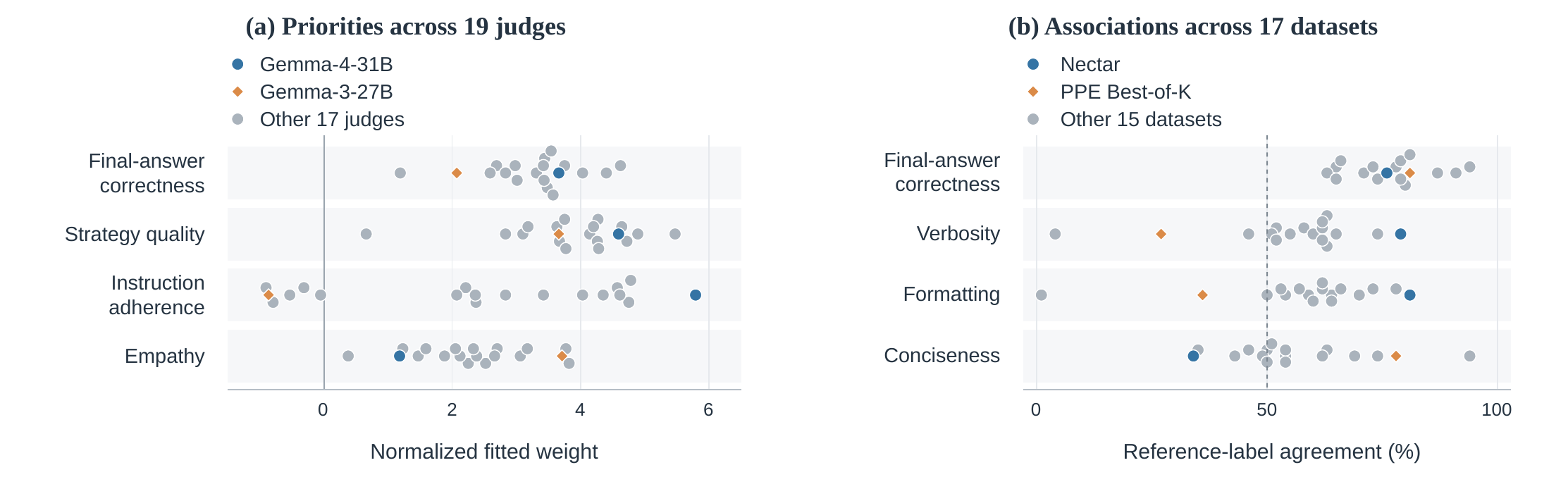}
\caption{\textbf{Different judge priorities and dataset associations.} (a) Normalized fitted weights under a shared attribute representation; each dot is one judge. (b) Median agreement with reference labels across eligible readers when selecting the response with more of an attribute; each dot is one dataset. Vertical offsets only separate overlapping points. Four selected attributes are shown per panel. Full heatmaps and protocols: Appendix~\ref{app:priority-display}.}
\label{fig:s5-compact}
\end{figure}

Fitted weights also reveal which attributes each judge favors. For example, Gemma-4-31B assigns more weight to instruction adherence than correctness, whereas Gemma-3-27B assigns more weight to empathy than correctness (Figure~\ref{fig:s5-compact}a). Reference datasets also differ: verbosity is positively associated with preferred responses in Nectar but negatively associated in PPE Best-of-K (Figure~\ref{fig:s5-compact}b). Further analysis of these differences appears in Appendix~\ref{sec:what-judges-favor}.

\section{Related Work}

\subsection{LLM-as-a-Judge and subjectivity.}
MT-Bench and G-Eval establish LLMs as useful evaluators~\citep{zheng2023judging,liu2023geval}. Subsequent work examines length and position biases~\citep{dubois2024lengthcontrolled,shi2025positionbias}, while PoLL combines diverse judges to reduce dependence on a single model~\citep{verga2024juries}. Separately, OpinionQA finds that LLMs represent human viewpoints unevenly~\citep{santurkar2023opinions}, and pluralistic alignment argues for supporting diverse values~\citep{sorensen2024pluralistic}. These concerns motivate examining disagreement without treating every difference as an error. We ask whether judges disagree in perception or prioritization.

\subsection{Attribute judgments and interpretable weights.}
FLASK evaluates instruction-specific skills, and HelpSteer annotates response attributes alongside helpfulness~\citep{ye2023flask,wang2023helpsteer}. VibeCheck discovers qualitative differences between models and relates them to user preferences~\citep{dunlap2024vibecheck}. WIMHF learns interpretable preference features and annotator-specific weights~\citep{movva2025wimhf}. ArmoRM combines objective scores through context-dependent weights~\citep{wang2024armorm}. These studies show how attributes and weights can explain preferences. We use a shared attribute inventory to study agreement across LLM judges, showing that shared attribute judgments can coexist with different overall judgments and fitted weights.

\subsection{Adapting LLM judges.}
JudgeLM fine-tunes models on generated judgments~\citep{zhu2025judgelm}, while Prometheus and Prometheus~2 train evaluators to follow custom criteria~\citep{kim2024prometheus,kim2024prometheus2}. LoRA reduces the number of parameters updated during fine-tuning~\citep{hu2021lora}. Rubric-based methods offer another form of control: PReMISE audits rubric reliability, preference fit, and robustness~\citep{roy2026premise}, and CalibratedRubric selects task-adaptive rubric banks~\citep{chen2026calibratedrubric}. We instead test how much changing attribute weights alone can steer subjectivity while keeping recorded attribute judgments fixed. Comparisons with fine-tuning under matched label budgets and rubric prompting with matched attribute counts assess the value of this direct control over prioritization.

\section{Conclusion}
LLM judges can agree on response qualities while disagreeing on what makes a response better. \rc{} makes this distinction measurable by separating perception, how a judge compares responses on attributes, from prioritization, how much those attributes matter to its overall judgment. Across 21 LLM judges, we find substantial attribute consensus even when judges choose different winners, a pattern also seen in human judgments. Yet judges' fitted weights differ even when estimated from the same attribute judgments. Reweighting shows that changing how existing assessments contribute to a decision can improve agreement with target reference labels without updating the language model, especially when labels are limited.

\subsection{Limitations and future work.}
For simplicity, we model prioritization with a logistic model that combines attribute judgments linearly. This makes prioritization measurable and interpretable, but does not imply that judges combine attributes linearly; they may weigh attributes differently depending on the query, or rely on interactions among them. A single weight vector per source cannot capture such trade-offs, which may explain why full-data fine-tuning surpasses reweighting on three of four judges. Richer models of prioritization, such as context-dependent or nonlinear weighting, are a natural next step.

\subsection{Broader implications.}
Better perception alone does not specify whose standards a judge should follow. Judge training and evaluation should therefore specify not only which responses are preferred, but also which attributes to prioritize and how to resolve trade-offs between them. Explicit priorities make judge behavior easier to audit and adapt.

\clearpage
\label{main-text-end}

\bibliography{paper,insight_references}

@article{sorensen2024pluralistic,
 title={A Roadmap to Pluralistic Alignment},
 author={Sorensen, Taylor and Moore, Jared and Fisher, Jillian and Gordon, Mitchell and Mireshghallah, Niloofar and Rytting, Christopher Michael and Ye, Andre and Jiang, Liwei and Lu, Ximing and Dziri, Nouha and Althoff, Tim and Choi, Yejin},
 journal={arXiv preprint arXiv:2402.05070},
 year={2024},
 url={https://arxiv.org/abs/2402.05070}
}

@article{wang2024armorm,
 title={Interpretable Preferences via Multi-Objective Reward Modeling and Mixture-of-Experts},
 author={Wang, Haoxiang and Xiong, Wei and Xie, Tengyang and Zhao, Han and Zhang, Tong},
 journal={arXiv preprint arXiv:2406.12845},
 year={2024},
 url={https://arxiv.org/abs/2406.12845}
}

@inproceedings{santurkar2023opinions,
  title={Whose Opinions Do Language Models Reflect?},
  author={Santurkar, Shibani and Durmus, Esin and Ladhak, Faisal and Lee, Cinoo and Liang, Percy and Hashimoto, Tatsunori},
  booktitle={Proceedings of the 40th International Conference on Machine Learning},
  volume={202}, pages={29971--30004}, year={2023},
  publisher={PMLR}, url={https://proceedings.mlr.press/v202/santurkar23a.html}
}

@inproceedings{shi2025positionbias,
  title={Judging the Judges: A Systematic Study of Position Bias in {LLM}-as-a-Judge},
  author={Shi, Lin and Ma, Chiyu and Liang, Wenhua and Diao, Xingjian and Ma, Weicheng and Vosoughi, Soroush},
  booktitle={Proceedings of the 14th International Joint Conference on Natural Language Processing and the 4th Conference of the Asia-Pacific Chapter of the Association for Computational Linguistics},
  pages={292--314}, year={2025}, doi={10.18653/v1/2025.ijcnlp-long.18},
  url={https://aclanthology.org/2025.ijcnlp-long.18/}
}

@article{verga2024juries,
  title={Replacing Judges with Juries: Evaluating {LLM} Generations with a Panel of Diverse Models},
  author={Verga, Pat and Hofstatter, Sebastian and Althammer, Sophia and Su, Yixuan and Piktus, Aleksandra and Arkhangorodsky, Arkady and Xu, Minjie and White, Naomi and Lewis, Patrick},
  journal={arXiv preprint arXiv:2404.18796}, year={2024},
  url={https://arxiv.org/abs/2404.18796}
}

@inproceedings{zhu2025judgelm,
  title={{JudgeLM}: Fine-tuned Large Language Models are Scalable Judges},
  author={Zhu, Lianghui and Wang, Xinggang and Wang, Xinlong},
  booktitle={International Conference on Learning Representations}, year={2025},
  url={https://arxiv.org/abs/2310.17631}
}

@inproceedings{kim2024prometheus2,
  title={Prometheus 2: An Open Source Language Model Specialized in Evaluating Other Language Models},
  author={Kim, Seungone and Suk, Juyoung and Longpre, Shayne and Lin, Bill Yuchen and Shin, Jamin and Welleck, Sean and Neubig, Graham and Lee, Moontae and Lee, Kyungjae and Seo, Minjoon},
  booktitle={Proceedings of the 2024 Conference on Empirical Methods in Natural Language Processing},
  pages={4334--4353}, year={2024}, doi={10.18653/v1/2024.emnlp-main.248},
  url={https://aclanthology.org/2024.emnlp-main.248/}
}

@article{hu2021lora,
  title={{LoRA}: Low-Rank Adaptation of Large Language Models},
  author={Hu, Edward J. and Shen, Yelong and Wallis, Phillip and Allen-Zhu, Zeyuan and Li, Yuanzhi and Wang, Shean and Wang, Lu and Chen, Weizhu},
  journal={arXiv preprint arXiv:2106.09685}, year={2021},
  url={https://arxiv.org/abs/2106.09685}
}

@article{tan2024judgebench,
  title={{JudgeBench}: A Benchmark for Evaluating {LLM}-Based Judges},
  author={Tan, Sijun and Zhuang, Siyuan and Montgomery, Kyle and Tang, Willian Yuan and Cuadron, Alejandro and Wang, Chenguang and Popa, Raluca Ada and Stoica, Ion},
  journal={arXiv preprint arXiv:2410.12784}, year={2024},
  url={https://arxiv.org/abs/2410.12784}
}

@article{wen2026ifrewardbench,
  title={{IF-RewardBench}: Benchmarking Judge Models for Instruction-Following Evaluation},
  author={Wen, Bosi and Niu, Yilin and Wang, Cunxiang and Ling, Xiaoying and Zhang, Ying and Ke, Pei and Wang, Hongning and Huang, Minlie},
  journal={arXiv preprint arXiv:2603.04738}, year={2026},
  url={https://arxiv.org/abs/2603.04738}
}

@article{liu2024rmbench,
  title={{RM-Bench}: Benchmarking Reward Models of Language Models with Subtlety and Style},
  author={Liu, Yantao and Yao, Zijun and Min, Rui and Cao, Yixin and Hou, Lei and Li, Juanzi},
  journal={arXiv preprint arXiv:2410.16184}, year={2024},
  url={https://arxiv.org/abs/2410.16184}
}

@inproceedings{malik2026rewardbench2,
  title={{RewardBench 2}: Advancing Reward Model Evaluation},
  author={Malik, Saumya and Pyatkin, Valentina and Land, Sander and Morrison, Jacob and Smith, Noah A. and Hajishirzi, Hannaneh and Lambert, Nathan},
  booktitle={International Conference on Learning Representations}, year={2026},
  url={https://arxiv.org/abs/2506.01937}
}

@misc{arenaexpert5k,
  title={{arena-expert-5k}}, author={{LMArena}},
  year={n.d.}, howpublished={Hugging Face dataset},
  url={https://huggingface.co/datasets/lmarena-ai/arena-expert-5k},
  note={Accessed September 24, 2026}
}

@misc{arenahuman140k,
  title={{arena-human-preference-140k}}, author={{LMArena}},
  year={n.d.}, howpublished={Hugging Face dataset},
  url={https://huggingface.co/datasets/lmarena-ai/arena-human-preference-140k},
  note={Accessed September 24, 2026}
}

@misc{zhu2023starling,
  title={{Starling-7B}: Improving {LLM} Helpfulness \& Harmlessness with {RLAIF}},
  author={Zhu, Banghua and Frick, Evan and Wu, Tianhao and Zhu, Hanlin and Jiao, Jiantao},
  year={2023},
  url={https://huggingface.co/datasets/berkeley-nest/Nectar}
}

@article{frick2024ppe,
  title={How to Evaluate Reward Models for {RLHF}},
  author={Frick, Evan and Li, Tianle and Chen, Connor and Chiang, Wei-Lin and Angelopoulos, Anastasios N. and Jiao, Jiantao and Zhu, Banghua and Gonzalez, Joseph E. and Stoica, Ion},
  journal={arXiv preprint arXiv:2410.14872}, year={2024},
  url={https://arxiv.org/abs/2410.14872}
}

@article{wang2024helpsteer2,
  title={{HelpSteer2}: Open-source dataset for training top-performing reward models},
  author={Wang, Zhilin and Dong, Yi and Delalleau, Olivier and Zeng, Jiaqi and Shen, Gerald and Egert, Daniel and Zhang, Jimmy J. and Sreedhar, Makesh Narsimhan and Kuchaiev, Oleksii},
  journal={arXiv preprint arXiv:2406.08673}, year={2024},
  url={https://arxiv.org/abs/2406.08673}
}

@misc{qwen2026qwen35,
  title={{Qwen3.5}: Towards Native Multimodal Agents},
  author={{Qwen Team}}, year={2026}, month={February},
  url={https://qwen.ai/blog?id=qwen3.5}
}

@misc{qwen2026qwen36,
  title={{Qwen3.6-27B}: Flagship-Level Coding in a {27B} Dense Model},
  author={{Qwen Team}}, year={2026}, month={April},
  url={https://qwen.ai/blog?id=qwen3.6-27b}
}

@article{deepseek2025r1,
  title={{DeepSeek-R1}: Incentivizing Reasoning Capability in {LLMs} via Reinforcement Learning},
  author={{DeepSeek-AI}}, journal={arXiv preprint arXiv:2501.12948}, year={2025},
  url={https://arxiv.org/abs/2501.12948}
}

@misc{deepseek2026v4,
  title={{DeepSeek-V4}: Towards Highly Efficient Million-Token Context Intelligence},
  author={{DeepSeek-AI}}, year={2026},
  url={https://huggingface.co/deepseek-ai/DeepSeek-V4-Pro}
}

@article{gemma2025gemma3,
  title={Gemma 3 Technical Report}, author={{Gemma Team}},
  journal={arXiv preprint arXiv:2503.19786}, year={2025},
  url={https://arxiv.org/abs/2503.19786}
}

@article{gemma2026gemma4,
  title={Gemma 4 Technical Report}, author={{Gemma Team}},
  journal={arXiv preprint arXiv:2607.02770}, year={2026},
  url={https://arxiv.org/abs/2607.02770}
}

@misc{meta2025llama4,
  title={Llama 4 Model Card}, author={{Meta}}, year={2025},
  url={https://huggingface.co/meta-llama/Llama-4-Scout-17B-16E-Instruct}
}

@misc{mistral2025small3,
  title={Mistral Small 3}, author={{Mistral AI Team}}, year={2025},
  url={https://mistral.ai/news/mistral-small-3/}
}

@misc{meta2026muse,
  title={Muse Glimmer Model Card}, author={{Meta Superintelligence Lab}}, year={2026},
  url={https://huggingface.co/meta-models/Muse-Glimmer-30B}
}

@article{kimi2026k3,
  title={{Kimi K3}: Open Frontier Intelligence}, author={{Kimi Team}},
  journal={arXiv preprint arXiv:2607.24653}, year={2026},
  url={https://arxiv.org/abs/2607.24653}
}

@inproceedings{zheng2023judging,
  title     = {Judging {LLM}-as-a-Judge with {MT}-Bench and Chatbot Arena},
  author    = {Zheng, Lianmin and Chiang, Wei-Lin and Sheng, Ying and Zhuang, Siyuan and Wu, Zhanghao and Zhuang, Yonghao and Lin, Zi and Li, Zhuohan and Li, Dacheng and Xing, Eric P. and Zhang, Hao and Gonzalez, Joseph E. and Stoica, Ion},
  booktitle = {Advances in Neural Information Processing Systems, Datasets and Benchmarks Track},
  year      = {2023}
}

@article{chiang2024arena,
  title   = {Chatbot Arena: An Open Platform for Evaluating {LLMs} by Human Preference},
  author  = {Chiang, Wei-Lin and Zheng, Lianmin and Sheng, Ying and Angelopoulos, Anastasios Nikolas and Li, Tianle and Li, Dacheng and Zhang, Hao and Zhu, Banghua and Jordan, Michael and Gonzalez, Joseph E. and Stoica, Ion},
  journal = {arXiv preprint arXiv:2403.04132},
  year    = {2024}
}

@article{ye2023flask,
  title   = {{FLASK}: Fine-grained Language Model Evaluation based on Alignment Skill Sets},
  author  = {Ye, Seonghyeon and Kim, Doyoung and Kim, Sungdong and Hwang, Hyeonbin and Kim, Seungone and Jo, Yongrae and Thorne, James and Kim, Juho and Seo, Minjoon},
  journal = {arXiv preprint arXiv:2307.10928},
  year    = {2023}
}

@article{wang2023helpsteer,
  title   = {HelpSteer: Multi-attribute Helpfulness Dataset for {SteerLM}},
  author  = {Wang, Zhilin and Dong, Yi and Zeng, Jiaqi and Adams, Virginia and Sreedhar, Makesh Narsimhan and Egert, Daniel and Delalleau, Olivier and Scowcroft, Jane Polak and Kant, Neel and Swope, Aidan and Kuchaiev, Oleksii},
  journal = {arXiv preprint arXiv:2311.09528},
  year    = {2023}
}

@article{dunlap2024vibecheck,
  title   = {VibeCheck: Discover and Quantify Qualitative Differences in Large Language Models},
  author  = {Dunlap, Lisa and Mandal, Krishna and Darrell, Trevor and Steinhardt, Jacob and Gonzalez, Joseph E.},
  journal = {arXiv preprint arXiv:2410.12851},
  year    = {2024}
}

@article{movva2025wimhf,
  title   = {What's In My Human Feedback? Learning Interpretable Descriptions of Preference Data},
  author  = {Movva, Rajiv and Milli, Smitha and Min, Sewon and Pierson, Emma},
  journal = {arXiv preprint arXiv:2510.26202},
  year    = {2025}
}

@inproceedings{liu2023geval,
  title     = {{G-Eval}: {NLG} Evaluation using {GPT-4} with Better Human Alignment},
  author    = {Liu, Yang and Iter, Dan and Xu, Yichong and Wang, Shuohang and Xu, Ruochen and Zhu, Chenguang},
  booktitle = {Proceedings of the 2023 Conference on Empirical Methods in Natural Language Processing},
  year      = {2023}
}

@inproceedings{kim2024prometheus,
  title     = {Prometheus: Inducing Fine-Grained Evaluation Capability in Language Models},
  author    = {Kim, Seungone and others},
  booktitle = {International Conference on Learning Representations},
  year      = {2024}
}

@article{dubois2024lengthcontrolled,
  title   = {Length-Controlled {AlpacaEval}: A Simple Way to Debias Automatic Evaluators},
  author  = {Dubois, Yann and Galambosi, Bal{\'a}zs and Liang, Percy and Hashimoto, Tatsunori B.},
  journal = {arXiv preprint arXiv:2404.04475},
  year    = {2024}
}

@inproceedings{stiennon2020learning,
  title     = {Learning to Summarize with Human Feedback},
  author    = {Stiennon, Nisan and Ouyang, Long and Wu, Jeffrey and Ziegler, Daniel M. and Lowe, Ryan and Voss, Chelsea and Radford, Alec and Amodei, Dario and Christiano, Paul F.},
  booktitle = {Advances in Neural Information Processing Systems},
  year      = {2020}
}

@article{bai2022training,
  title   = {Training a Helpful and Harmless Assistant with Reinforcement Learning from Human Feedback},
  author  = {Bai, Yuntao and others},
  journal = {arXiv preprint arXiv:2204.05862},
  year    = {2022}
}

@article{nakano2021webgpt,
  title   = {{WebGPT}: Browser-Assisted Question-Answering with Human Feedback},
  author  = {Nakano, Reiichiro and Hilton, Jacob and Balaji, Suchir and Wu, Jeff and Ouyang, Long and Kim, Christina and Hesse, Christopher and Jain, Shantanu and Kosaraju, Vineet and Saunders, William and Jiang, Xu and Cobbe, Karl and Eloundou, Tyna and Krueger, Gretchen and Button, Kevin and Knight, Matthew and Chess, Benjamin and Schulman, John},
  journal = {arXiv preprint arXiv:2112.09332},
  year    = {2021}
}

@inproceedings{ethayarajh2022understanding,
  title     = {Understanding Dataset Difficulty with $\mathcal{V}$-Usable Information},
  author    = {Ethayarajh, Kawin and Choi, Yejin and Swayamdipta, Swabha},
  booktitle = {International Conference on Machine Learning},
  year      = {2022}
}

@article{cui2023ultrafeedback,
  title   = {{UltraFeedback}: Boosting Language Models with Scaled {AI} Feedback},
  author  = {Cui, Ganqu and Yuan, Lifan and Ding, Ning and Yao, Guanming and Zhu, Wei and Ni, Yuan and Xie, Guotong and Liu, Zhiyuan and Sun, Maosong},
  journal = {arXiv preprint arXiv:2310.01377},
  year    = {2023}
}

@article{lambert2024rewardbench,
  title   = {{RewardBench}: Evaluating Reward Models for Language Modeling},
  author  = {Lambert, Nathan and Pyatkin, Valentina and Morrison, Jacob and Miranda, LJ and Lin, Bill Yuchen and Chandu, Khyathi and Dziri, Nouha and Kumar, Sachin and Zick, Tom and Choi, Yejin and Smith, Noah A. and Hajishirzi, Hannaneh},
  journal = {arXiv preprint arXiv:2403.13787},
  year    = {2024}
}

@article{roy2026premise,
  title   = {{PReMISE}: Policy Rubrics as Measurement Specifications for {LLM} Judges},
  author  = {Roy, Swastik and Pujari, Rajkumar and Kumarage, Tharindu and Peris, Charith and Gupta, Rahul and Rumshisky, Anna and Natarajan, Pradeep and Saligrama, Venkatesh},
  journal = {arXiv preprint arXiv:2605.30803},
  year    = {2026}
}

@article{chen2026calibratedrubric,
  title   = {CalibratedRubric: Task-Adaptive Rubric Banks for Open-Ended {LLM} Evaluation},
  author  = {Chen, Mengting and Sun, Yanshu and Liang, Wanting and Luan, Beidi and Sun, Rui and Chen, Dezhi and Li, Jing and Bai, Zuo},
  journal = {arXiv preprint arXiv:2607.29252},
  year    = {2026}
}
\bibliographystyle{assets/plainnat}
\clearpage
\appendix
\raggedbottom
\makeatletter
\setlength{\@fptop}{0pt}
\setlength{\@fpsep}{12pt}
\setlength{\@fpbot}{0pt plus 1fil}
\setlength{\@dblfptop}{0pt}
\setlength{\@dblfpsep}{12pt}
\setlength{\@dblfpbot}{0pt plus 1fil}
\makeatother
\setcounter{topnumber}{4}
\renewcommand{\topfraction}{0.95}
\renewcommand{\textfraction}{0.05}
\setlength{\parskip}{3pt}
\setlength{\floatsep}{8pt plus 2pt minus 2pt}
\setlength{\textfloatsep}{9pt plus 2pt minus 2pt}
\setlength{\intextsep}{8pt plus 2pt minus 2pt}
\setlength{\abovecaptionskip}{4pt}
\setlength{\belowcaptionskip}{2pt}

\section*{Appendix Contents}
\noindent The appendices provide protocols, complete results, robustness checks, and the human-study cases supporting Sections~\ref{sec:hidden-consensus} and~\ref{sec:fingerprint}. The analysis map in Table~\ref{tab:analysis-map} distinguishes judge populations and prediction targets.
\begin{center}
\small\renewcommand{\arraystretch}{1.35}
\begin{tabularx}{\linewidth}{@{}>{\raggedright\arraybackslash}Xr@{}}
\toprule
\textbf{Appendix and purpose} & \textbf{Page}\\
\midrule
\hyperref[app:methods]{\textbf{\ref*{app:methods}. Evaluation Setup and Measurement}}\newline Sources, models, attributes, order consistency, agreement metrics, and data splits. & \pageref{app:methods}\\
\hyperref[app:hc-judge]{\textbf{\ref*{app:hc-judge}. Perception: Agreement and Robustness}}\newline Conditional agreement, coverage, single-read controls, and human agreement. & \pageref{app:hc-judge}\\
\hyperref[app:priority-controls]{\textbf{\ref*{app:priority-controls}. Prioritization: Fitted Weights and Controls}}\newline Reproducibility, shared-attribute fits, component swaps, and judge profiles. & \pageref{app:priority-controls}\\
\hyperref[app:attribute-consensus-detail]{\textbf{\ref*{app:attribute-consensus-detail}. Attribute-level Consensus and Importance}}\newline The 21-judge statistics and selection rule behind the nine-attribute figure. & \pageref{app:attribute-consensus-detail}\\
\hyperref[app:no-conflict-detail]{\textbf{\ref*{app:no-conflict-detail}. Overall Disagreement Without Attribute Conflict}}\newline The 14.5\% result, coverage thresholds, and judge/source sensitivity. & \pageref{app:no-conflict-detail}\\
\hyperref[app:alignment]{\textbf{\ref*{app:alignment}. Reweighting and Adaptation}}\newline All judge--source results, RM-Bench sampling, and matched adaptation controls. & \pageref{app:alignment}\\
\hyperref[app:human-cases]{\textbf{\ref*{app:human-cases}. SubjectiveSet--Human: Complete Case Studies}}\newline All 23 cases, displayed attributes, and individual overall judgments. & \pageref{app:human-cases}\\
\bottomrule
\end{tabularx}
\end{center}
\clearpage

\section{Evaluation Setup and Measurement}
\label{app:methods}
This appendix defines the evaluation populations and measurement rules used throughout the paper. Table~\ref{tab:analysis-map} identifies the population and target of each analysis.

\begin{table}[t]
\centering\small\setlength{\tabcolsep}{4pt}
\caption{\textbf{Each analysis addresses a different part of judgment.} Judge counts and evaluation targets distinguish the study populations. \newline}

\label{tab:analysis-map}
\begin{tabularx}{\linewidth}{@{}>{\raggedright\arraybackslash}p{0.30\linewidth}l >{\raggedright\arraybackslash}X@{}}
\toprule
Analysis & Population & Comparison or prediction target \\
\midrule
Heatmap consensus & 19 LLMs & Full SubjectiveSet; 171 judge pairs in Figure~\ref{fig:hc-strata} \\
Collection and subset consensus & 21 LLMs & Full collection, SubjectiveSet--LLM, and its complement; 210 judge pairs \\
Human cases & 7 raters & SubjectiveSet--Human: 21 disagreement cases and 2 controls \\
Priority profiles & 19 LLMs & Each judge's own verdicts, using own or shared attributes \\
Attribute importance and consensus & 21 LLMs & Own-verdict weights and attribute agreement; Figure~\ref{fig:attribute-agreement-selected} \\
No-conflict comparison & 21 LLMs & Overall disagreement after requiring at least 20 shared attribute directions; Figure~\ref{fig:consensus-limits} \\
Broad reweighting & 21 LLMs & Source reference labels; 357 judge--distribution combinations \\
Sample-efficiency curves & 9 LLMs & Reference labels on three sources; ten training-subset resamples \\
SFT and rubric controls & 4 LLMs & Reference labels on the same 11,765 non-tie test pairs \\
\bottomrule
\end{tabularx}
\end{table}
\FloatBarrier

\subsection{Response pairs, reference labels, and subsets}
\label{app:datasets}
\begin{figure}[t]
\centering
\begingroup
\tcbset{
  label-example/.style={
    colback=white,
    colframe=blue!35!black,
    boxrule=0.5pt,
    arc=2pt,
    left=6pt,
    right=6pt,
    top=6pt,
    bottom=6pt,
    equal height group=labeltypeexamples,
    space to upper,
    fontupper=\small,
    fontlower=\small,
    before upper={\raggedright},
    before lower={\raggedright}
  }
}

\begin{minipage}[t]{0.485\linewidth}
\vspace{0pt}
\begin{tcolorbox}[label-example]
\textbf{(a) Preference-labeled data}\par\smallskip
Arena human preferences~\citep{chiang2024arena}\par
\textbf{Label basis:} recorded human preference.\par\medskip

\textbf{Prompt:} Whats 1 + 1.. 11 right?\par\medskip

\textcolor[HTML]{4B8FA8}{\textbf{Response A (summary)}}\par
Says that $1+1=2$, but treats 11 as a playful joke.\par\smallskip

\textcolor[HTML]{C98455}{\textbf{Response B (summary)}}\par
Corrects the answer to 2 and explains how addition
differs from concatenation.\par\medskip

\colorbox{black!6}{\strut\textbf{Source reference label: B}}

\tcblower

\textbf{LLM attribute judgments}\par\smallskip
Humor: \textbf{A higher} (19/21).\par
Rigor: \textbf{B higher} (18/21).\par\smallskip
The preferred response is not higher on every attribute.
\end{tcolorbox}
\end{minipage}\hfill
\begin{minipage}[t]{0.485\linewidth}
\vspace{0pt}
\begin{tcolorbox}[label-example]
\textbf{(b) Benchmark-labeled data}\par\smallskip
RewardBench~\citep{lambert2024rewardbench}\par
\textbf{Label basis:} reference choice supplied by the benchmark.\par\medskip

\textbf{Prompt:} Why does the second batch of pancakes
brown much faster?\par\medskip

\textcolor[HTML]{4B8FA8}{\textbf{Response A (summary)}}\par
Gives a short explanation: the pan is already hot.\par\smallskip

\textcolor[HTML]{C98455}{\textbf{Response B (summary)}}\par
Gives a longer explanation covering residual heat,
leftover batter or oil, and pan type.\par\medskip

\colorbox{black!6}{\strut\textbf{Source reference label: B}}

\tcblower

\textbf{LLM attribute judgments}\par\smallskip
Conciseness: \textbf{A higher} (18/21).\par
Explanatory depth: \textbf{B higher} (21/21).\par\smallskip
The benchmark label does not rank each attribute separately.
\end{tcolorbox}
\end{minipage}

\endgroup
\caption{\textbf{Two types of reference labels in SubjectiveSet.}
Real examples from Arena and RewardBench; responses are summarized
for space. Counts show how many of the 21 LLM judges identify the
indicated response as higher on an attribute. Full responses,
source identifiers, and judgments appear in
Cases~\hyperref[jc:case:8]{8} and~\hyperref[jc:case:21]{21}
in Appendix~\ref{app:human-cases}.}
\label{fig:label-type-examples}
\end{figure}

Our evaluation collection contains 50,013 response pairs from 17 public
source distributions. Each item consists of a query and two candidate
responses, denoted $A$ and $B$. Table~\ref{tab:app-datasets} lists the source
identifiers and the numbers of collected, training, and test pairs. These
counts describe the subsets used in our study, rather than the full sizes
of the original datasets. The collection spans conversational assistance,
summarization, instruction following, and preference evaluation.

Reference labels retain their source-specific meaning. They include human
preferences, AI-generated preferences, and benchmark-provided labels. We
therefore use \emph{reference-label accuracy} when reporting agreement with
these targets.

For the three adaptation experiments, each source is split into 75\% training and 25\% test data. The counts below retain only pairs with a non-tie target in $\{A,B\}$ for supervised adaptation and reference-label evaluation. This gives 35,156 training pairs and
11,765 held-out test pairs. The remaining 3,092 tied pairs are excluded
from both labeled splits. They remain part of the 50,013-pair collection
used to characterize judges' response comparisons.

\begin{table*}[t]
\centering
\caption{The 17 source distributions used in our study. ``Benchmark''
denotes the benchmark-provided label category, distinct from human and AI preferences. Training and test
counts include only non-tie pairs with a usable $A/B$ reference label.\newline}

\label{tab:app-datasets}
\small
\setlength{\tabcolsep}{4pt}
\renewcommand{\arraystretch}{1.08}
\begin{tabularx}{\textwidth}{@{}>{\raggedright\arraybackslash}Xlrrr@{}}
\toprule
\textbf{Source distribution} & \textbf{Label type} & \textbf{Pairs} & \textbf{Train} & \textbf{Test} \\
\midrule
\path{ScalerLab/JudgeBench}~\citep{tan2024judgebench} & Benchmark & 617 & 463 & 154 \\
\path{IF-RewardBench}~\citep{wen2026ifrewardbench} & Benchmark & 807 & 605 & 202 \\
\path{THU-KEG/RM-Bench}~\citep{liu2024rmbench} & Benchmark & 1,146 & 860 & 286 \\
\path{allenai/reward-bench-2}~\citep{malik2026rewardbench2} & Benchmark & 1,657 & 1,243 & 414 \\
\path{lmarena/arena-expert-5k}~\citep{arenaexpert5k} & Human & 1,797 & 1,348 & 449 \\
\path{berkeley-nest/Nectar}~\citep{zhu2023starling} & AI & 2,500 & 1,875 & 625 \\
\path{openbmb/UltraFeedback}~\citep{cui2023ultrafeedback} & AI & 2,500 & 1,875 & 625 \\
\path{lmarena-ai/PPE-*-Best-of-K}~\citep{frick2024ppe} & Benchmark & 2,508 & 1,881 & 627 \\
\path{nvidia/HelpSteer2}~\citep{wang2024helpsteer2} & Human & 4,000 & 2,131 & 724 \\
\path{openai/webgpt_comparisons}~\citep{nakano2021webgpt} & Human & 4,000 & 2,197 & 756 \\
\path{lmsys/mt_bench_human_judgments}~\citep{zheng2023judging} & Human & 4,000 & 2,317 & 783 \\
\path{Anthropic/hh-rlhf}~\citep{bai2022training} & Human & 4,000 & 3,000 & 1,000 \\
\path{lmarena/arena-human-preference-140k}~\citep{arenahuman140k} & Human & 4,000 & 3,000 & 1,000 \\
\path{lmarena-ai/PPE-Human-Preference-V1}~\citep{frick2024ppe} & Human & 4,000 & 3,000 & 1,000 \\
\path{openai/summarize_from_feedback}~\citep{stiennon2020learning} & Human & 4,000 & 3,000 & 1,000 \\
\path{stanfordnlp/SHP}~\citep{ethayarajh2022understanding} & Human & 4,000 & 3,000 & 1,000 \\
\path{allenai/reward-bench}~\citep{lambert2024rewardbench} & Benchmark & 4,481 & 3,361 & 1,120 \\
\midrule
\textbf{Total} & & \textbf{50,013} & \textbf{35,156} & \textbf{11,765} \\
\bottomrule
\end{tabularx}
\end{table*}

\subsection{SubjectiveSet: subset definitions and intended use.}
\label{app:jc-subsets}
The SubjectiveSet dataset is the complete 50,013-pair collection. Its two curated research subsets overlap and are distinct from the original training/test partition. They retain source identifiers and reference labels so that preference disagreement can be distinguished from errors on verifiable tasks.

For item $i$, let $n_A(i)$ and $n_B(i)$ count definite overall verdicts among all 21 LLM judges, excluding ties and unparsed outputs. Define the minority share
\begin{equation}
m(i)=\frac{\min\{n_A(i),n_B(i)\}}{n_A(i)+n_B(i)},\qquad
\mathrm{SubjectiveSet\text{--}LLM}=\{i:m(i)\geq 1/3\}.
\end{equation}
This selects 11,524 pairs across all 17 sources; 38,489 pairs remain outside the subset. The source-specific selection rate ranges from 7.7\% on RewardBench to 43.1\% on IF-RewardBench. Excluding Qwen3.5-0.8B and Qwen3.5-2B gives a 19-judge sensitivity set of 9,624 pairs: 9,222 overlap the primary set, 2,302 occur only under the 21-judge definition, and 402 only under the 19-judge definition. All main SubjectiveSet--LLM results use the 21-judge definition.

SubjectiveSet--Human contains all 23 curated cases, including 21 model-disagreement cases and two controls, with seven raters and 55 selected item--attribute cells (two or three attributes per item). Nineteen of the 23 satisfy the SubjectiveSet--LLM threshold; the other two disagreement cases fall below it, and neither control qualifies. The set was curated to make response differences inspectable.

\begin{figure}[t]
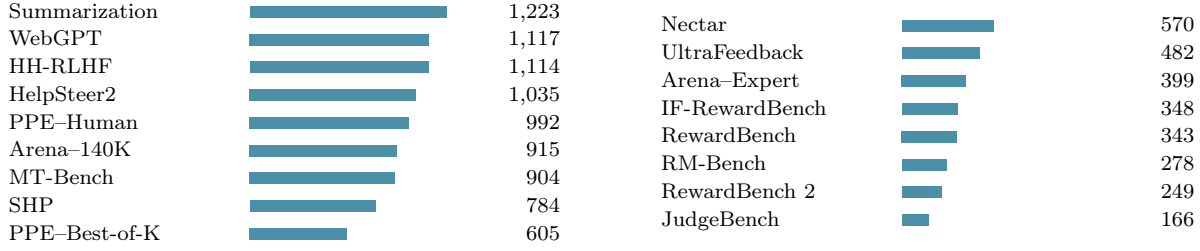

\centering
\begin{minipage}[t]{0.48\linewidth}
\footnotesize
\setlength{\tabcolsep}{0pt}
\renewcommand{\arraystretch}{1.10}
\begin{tabular}{@{}p{0.40\linewidth}p{0.43\linewidth}r@{}}
Summarization & \textcolor[HTML]{4B8FA8}{\rule{26.00mm}{1.5mm}} & 1,223 \\
WebGPT & \textcolor[HTML]{4B8FA8}{\rule{23.74mm}{1.5mm}} & 1,117 \\
HH-RLHF & \textcolor[HTML]{4B8FA8}{\rule{23.68mm}{1.5mm}} & 1,114 \\
HelpSteer2 & \textcolor[HTML]{4B8FA8}{\rule{22.00mm}{1.5mm}} & 1,035 \\
PPE--Human & \textcolor[HTML]{4B8FA8}{\rule{21.09mm}{1.5mm}} & 992 \\
Arena--140K & \textcolor[HTML]{4B8FA8}{\rule{19.45mm}{1.5mm}} & 915 \\
MT-Bench & \textcolor[HTML]{4B8FA8}{\rule{19.22mm}{1.5mm}} & 904 \\
SHP & \textcolor[HTML]{4B8FA8}{\rule{16.66mm}{1.5mm}} & 784 \\
PPE--Best-of-K & \textcolor[HTML]{4B8FA8}{\rule{12.87mm}{1.5mm}} & 605 \\
\end{tabular}
\end{minipage}
\hfill
\begin{minipage}[t]{0.48\linewidth}
\footnotesize
\setlength{\tabcolsep}{0pt}
\renewcommand{\arraystretch}{1.10}
\begin{tabular}{@{}p{0.40\linewidth}p{0.43\linewidth}r@{}}
Nectar & \textcolor[HTML]{4B8FA8}{\rule{12.11mm}{1.5mm}} & 570 \\
UltraFeedback & \textcolor[HTML]{4B8FA8}{\rule{10.24mm}{1.5mm}} & 482 \\
Arena--Expert & \textcolor[HTML]{4B8FA8}{\rule{8.48mm}{1.5mm}} & 399 \\
IF-RewardBench & \textcolor[HTML]{4B8FA8}{\rule{7.40mm}{1.5mm}} & 348 \\
RewardBench & \textcolor[HTML]{4B8FA8}{\rule{7.29mm}{1.5mm}} & 343 \\
RM-Bench & \textcolor[HTML]{4B8FA8}{\rule{5.91mm}{1.5mm}} & 278 \\
RewardBench 2 & \textcolor[HTML]{4B8FA8}{\rule{5.29mm}{1.5mm}} & 249 \\
JudgeBench & \textcolor[HTML]{4B8FA8}{\rule{3.53mm}{1.5mm}} & 166 \\
\end{tabular}
\end{minipage}
\caption{\textbf{SubjectiveSet--LLM by source.} The 11,524 selected response pairs across 17 sources. Bar lengths use a shared scale; numbers give pair counts.}
\label{fig:jc-composition}
\end{figure}

\subsection{Which data each analysis uses.}
Section~\ref{sec:hidden-consensus}, Evidence~1, and the main priority-profile analyses use eligible comparisons from the full collection with 19 judges. Section~\ref{sec:hidden-consensus}, Evidence~2, reports the full collection, SubjectiveSet--LLM, and its complement with all 21 judges (210 judge pairs), and separately reports SubjectiveSet--Human and its two case strata. The attribute-importance analysis and the no-conflict comparison in Section~\ref{sec:fingerprint}, Evidence~1, also use all 21 judges. Adaptation uses source-specific training labels and non-tie test pairs from the full collection: 35,156 training and 11,765 test pairs. No human-study labels are used for training.

LLM agreement rates average eligible pairwise rates equally over judge pairs. Attribute rates condition on both judges resolving a stable direction; the selected subset has 18.7\% joint attribute coverage, versus 26.1\% in the remaining pairs. For opposite-winner comparisons, the corresponding coverages are 18.3\% and 19.7\%. Human aggregate agreement pools eligible rater-pair comparisons: 246/396 overall and 1,003/1,087 attribute comparisons across all 23 cases; the 21 disagreement cases contribute 204/354 and 942/1,015, respectively. Thus, the LLM and human blocks illustrate the same qualitative contrast, but differ in populations, attribute selection, and aggregation.

\begin{table}[t]
\centering
\caption{\textbf{Full agreement breakdown by collection and winner stratum.} Values are percentages; $\Delta$ is all-attribute minus overall agreement. LLM rates average judge pairs; human aggregate rates pool eligible comparisons. The human disagreement and control rows partition the 23-case SubjectiveSet--Human set.\newline}

\label{tab:jc-detailed-agreement}
\footnotesize\setlength{\tabcolsep}{4.5pt}
\begin{tabular}{llrrrrrr}
\toprule
& & & & \multicolumn{3}{c}{Attribute} & \\
\cmidrule(lr){5-7}
Judges & Items & $n$ & Overall & All & Same winner & Opp.\ winner & $\Delta$ \\
\midrule
LLMs (21)  & All pairs                          & 50{,}013 & 74.0 & 88.2 & 89.5 & 83.6 & $+14.2$ \\
           & \quad\textsc{SubjectiveSet--LLM}   & 11{,}524 & 49.8 & 84.3 & 85.8 & 82.9 & $+34.5$ \\
           & \quad Remaining pairs              & 38{,}489 & 81.1 & 89.0 & 89.9 & 84.0 & $+7.9$ \\
\midrule
Humans (7) & \textsc{SubjectiveSet--Human}                          & 23       & 62.1 & 92.3 & 92.2 & 89.9 & $+30.2$ \\
           & \quad Disagreement cases & 21       & 57.6 & 92.8 & 93.3 & 89.9 & $+35.2$ \\
           & \quad Control cases                & 2        & 100.0 & 84.7 & 84.7 & --- & $-15.3$ \\
\bottomrule
\end{tabular}
\end{table}

\subsection{Judges and analysis population}
\label{app:judges}

We collect judgments from 21 models across seven families: Qwen (8),
DeepSeek (5), Gemma (3), Llama (2), Mistral (1), Muse (1), and Kimi (1).
Table~\ref{tab:app-judges} gives the complete model list. Total parameter
counts range from 0.8B to 2.8T. For mixture-of-experts models, size refers
to total parameters, not the number of active parameters per token. The
four size groups in the overview figure contain 5 models below 10B,
9 models from 10B to below 100B, 5 models from 100B to below 1T, and
2 models at or above 1T.

The collection includes all 21 models. The heatmaps and main priority profiles use 19 judges, excluding Qwen3.5-0.8B and Qwen3.5-2B because of severe position bias in their attribute judgments: both choose position B about 80\% of the time on order-balanced data, including after swapping the responses. This behavior limits their reliability as attribute judges. Their fractions of order-consistent attribute directions are correspondingly low: 0.165 and 0.106 (Appendix~\ref{app:coverage}). Both models remain in the collection-scale counts, 21-judge agreement analyses, and broad adaptation evaluation.

\begin{table*}[t]
\centering
\caption{All 21 judges in the evaluation collection. Parameter counts
are total parameters. Accuracy is the unadapted judge's reference-label
accuracy (\%) on the common 11,765-pair test set, with unparsed verdicts
counted as errors. Models marked $\dagger$ are excluded from the main
19-judge agreement analyses. The Mistral run is marked $\ddagger$ because
its metadata contains an \texttt{extra\_system\_prompt} flag.\newline}

\label{tab:app-judges}
\small
\setlength{\tabcolsep}{7pt}
\renewcommand{\arraystretch}{1.08}
\begin{tabularx}{\textwidth}{@{}>{\raggedright\arraybackslash}Xlrr@{}}
\toprule
\textbf{Judge} & \textbf{Family} & \textbf{Parameters} & \textbf{Accuracy (\%)} \\
\midrule
Qwen3.5-0.8B$^{\dagger}$~\citep{qwen2026qwen35} & Qwen & 0.8B & 47.90 \\
Qwen3.5-2B$^{\dagger}$~\citep{qwen2026qwen35} & Qwen & 2B & 59.30 \\
Qwen3.5-4B~\citep{qwen2026qwen35} & Qwen & 4B & 63.79 \\
DeepSeek-R1-Distill-7B~\citep{deepseek2025r1} & DeepSeek & 7B & 58.74 \\
Qwen3.5-9B~\citep{qwen2026qwen35} & Qwen & 9B & 66.74 \\
DeepSeek-R1-Distill-14B~\citep{deepseek2025r1} & DeepSeek & 14B & 64.54 \\
Mistral-Small-24B$^{\ddagger}$~\citep{mistral2025small3} & Mistral & 24B & 65.21 \\
Gemma-4-26B-A4B~\citep{gemma2026gemma4} & Gemma & 26B & 69.22 \\
Qwen3.5-27B~\citep{qwen2026qwen35} & Qwen & 27B & 70.79 \\
Qwen3.6-27B~\citep{qwen2026qwen36} & Qwen & 27B & 71.59 \\
Gemma-3-27B~\citep{gemma2025gemma3} & Gemma & 27B & 67.20 \\
Muse-Glimmer-30B~\citep{meta2026muse} & Muse & 30B & 69.22 \\
Gemma-4-31B~\citep{gemma2026gemma4} & Gemma & 31B & 71.64 \\
DeepSeek-R1-Distill-32B~\citep{deepseek2025r1} & DeepSeek & 32B & 65.26 \\
DeepSeek-V4-Flash~\citep{deepseek2026v4} & DeepSeek & 291B & 67.22 \\
Llama-4-Scout~\citep{meta2025llama4} & Llama & 109B & 66.91 \\
Qwen3.5-122B-A10B~\citep{qwen2026qwen35} & Qwen & 122B & 69.04 \\
Llama-4-Maverick~\citep{meta2025llama4} & Llama & 400B & 67.51 \\
Qwen3.5-397B-A17B~\citep{qwen2026qwen35} & Qwen & 397B & 69.14 \\
DeepSeek-V4-Pro~\citep{deepseek2026v4} & DeepSeek & 1.7T & 69.80 \\
Kimi-K3~\citep{kimi2026k3} & Kimi & 2.8T & 72.70 \\
\bottomrule
\end{tabularx}
\end{table*}

\subsection{Attribute inventory and directional comparisons}
\label{app:attributes}

The attribute task separates a response characteristic from an overall
preference. For each query--response pair, a judge compares which response
exhibits more of one specified characteristic. Thus, a direction toward
$A$ means that $A$ has more of the named characteristic; it does not
necessarily mean that $A$ is the better response.

We use 87 attributes grouped into 14 families. Table~\ref{tab:app-attributes}
lists the complete inventory. The overview figure
combines four singleton families---content, format, interpersonal, and
stance---into \emph{Other} for display. All four remain separate attributes
in the evaluation.

\begin{table*}[t]
\centering
\caption{Complete inventory of the 87 response attributes. Spaces replace
underscores in the attribute identifiers for readability. ``Other'' in the
overview figure comprises the four families containing one attribute each.\newline}

\label{tab:app-attributes}
\small
\setlength{\tabcolsep}{5pt}
\renewcommand{\arraystretch}{1.12}
\begin{tabularx}{\textwidth}{@{}lr>{\raggedright\arraybackslash}X@{}}
\toprule
\textbf{Family} & \textbf{Count} & \textbf{Attributes} \\
\midrule
Alignment & 6 & harmfulness, moralizing tendency, neutrality, refusal tendency, safety conservatism, sycophancy \\
Communication & 8 & accessibility, clarity, coherence, fluency, notation clarity, organization, readability, terminological precision \\
Content & 1 & elaboration \\
Correctness & 11 & assumption validity, constraint satisfaction, factual accuracy, final answer correctness, groundedness, hallucination rate, instruction adherence, internal consistency, justification quality, logical validity, step correctness \\
Creative & 9 & analogy use, creativity, humor, interestingness, novelty, originality, persuasiveness, storytelling quality, vividness \\
Epistemic & 7 & acknowledgment of limitations, assertiveness, calibration, confidence, hedging, qualification, uncertainty expression \\
Format & 1 & formatting \\
Information & 10 & breadth, completeness, depth, detail level, evidence use, example use, explanatory depth, information density, informativeness, specificity \\
Interpersonal & 1 & forcefulness \\
Reasoning & 9 & decomposition quality, insightfulness, reasoning completeness, reasoning efficiency, rigor, robustness, self correction, strategy quality, verification \\
Relevance & 5 & digression, directness, focus, goal directedness, relevance \\
Social & 8 & audience adaptation, empathy, enthusiasm, friendliness, personalization, politeness, respectfulness, warmth \\
Stance & 1 & caution framing \\
Style & 10 & abstraction level, conceptual complexity, conciseness, conversationality, difficulty, formality, linguistic complexity, phrasing originality, technicality, verbosity \\
\midrule
\textbf{Total} & \textbf{87} & \\
\bottomrule
\end{tabularx}
\end{table*}

\subsection{Dataset screening and analysis population.}
The response-pair collection was screened using order-stable judgments from Qwen3.5-27B before the cross-judge analysis. This dataset-level selection is separate from the judge-specific stability mask applied to individual item--attribute cells below. Reported agreement describes this screened collection and the comparisons eligible for each judge pair.

\subsection{Auxiliary confidence and magnitude outputs.}
Attribute elicitation also requests confidence and magnitude in addition to direction. Confidence and magnitude are not analyzed in this paper.

\subsection{Presentation order, inference scale, and coverage}
\label{app:inference}

Each attribute comparison is elicited in its own inference call. We repeat
it under both presentation orders, $(A,B)$ and $(B,A)$, and map the output
back to the original response identities before comparing the two reads.
This distinguishes a preference for a response from a preference for its
position in the prompt. The completed collection contains all 87 attributes
for each of the 21 judges.

The number of attribute-level inference calls is therefore
\begin{equation}
\begin{aligned}
N_{\mathrm{attr}}
  &= N_{\mathrm{pairs}}\,N_{\mathrm{judges}}\,
     N_{\mathrm{attributes}}\,N_{\mathrm{orders}} \\
  &= 50{,}013 \times 21 \times 87 \times 2 \\
  &= 182{,}747{,}502.
\end{aligned}
\end{equation}
We report this as \textbf{182.7 million attribute-level inference calls}.
Holistic preference judgments are additional and are not included in this
number. This count describes the evaluation workload, not the number of
stable or usable attribute directions: an inference call can yield an
unresolved direction or a direction that changes after swapping the
response order.

\subsection{Stable directions and coverage.}
\label{app:coverage}

Let $r_{ijk}^{(o)}\in\{-1,0,+1\}$ denote the mapped attribute direction
for item $i$, judge $j$, attribute $k$, and presentation order $o$.
Here $+1$ points to the original response $A$, $-1$ points to the original
response $B$, and $0$ represents a read without a resolved binary direction.

A stable direction is defined by
\begin{equation}
s_{ijk}=
\begin{cases}
r_{ijk}^{(1)}, & r_{ijk}^{(1)}=r_{ijk}^{(2)}\in\{-1,+1\},\\
0, & \text{otherwise}.
\end{cases}
\label{eq:app-stable-direction}
\end{equation}
Thus, an order reversal or an unresolved read contributes no stable
direction. Two unresolved reads do not count as agreement about an
attribute. We write $m_{ijk}=\mathbb{1}[s_{ijk}\ne0]$ for the stable-read
indicator. With $N=50{,}013$ items and $K=87$ attributes, a judge's stable
coverage is
\begin{equation}
C_j=\frac{1}{NK}\sum_{i=1}^{N}\sum_{k=1}^{K}m_{ijk}.
\end{equation}
Coverage measures the fraction of item--attribute cells with a usable,
order-consistent direction.

\subsection{Agreement metrics}
\label{app:agreement}

\subsection{Overall verdict agreement.}
Let $v_{ij}\in\{A,B\}$ denote a resolved holistic verdict. For two judges
$j$ and $\ell$, verdict agreement is the fraction of items on which their
resolved verdicts match. To study opposite choices, we partition the
items with two resolved verdicts into
\begin{align}
\mathcal{I}^{\mathrm{same}}_{j\ell}
  &=\{i:v_{ij}=v_{i\ell}\},\\
\mathcal{I}^{\mathrm{opp}}_{j\ell}
  &=\{i:v_{ij}\ne v_{i\ell}\}.
\end{align}
Verdict agreement is exactly $1$ in the first group and $0$ in the second
by construction. An unresolved verdict cannot define either stratum.

\subsection{Attribute agreement for a judge pair.}
For a specified item set $\mathcal{I}$, agreement between judges $j$ and
$\ell$ over their shared stable attribute directions is
\begin{equation}
A_{j\ell}(\mathcal{I})=
\frac{\displaystyle\sum_{i\in\mathcal{I}}\sum_{k=1}^{K}
 m_{ijk}m_{i\ell k}\,\mathbb{1}[s_{ijk}=s_{i\ell k}]}{
 \displaystyle\sum_{i\in\mathcal{I}}\sum_{k=1}^{K}m_{ijk}m_{i\ell k}}.
\label{eq:app-pair-agreement}
\end{equation}
The same definition applies separately to the same-verdict and
opposite-verdict item sets. Only cells resolved stably by both judges
enter the denominator. The rate is undefined when no such cells exist.
A mean over judge pairs gives each defined judge-pair rate equal weight;
pooling the underlying comparisons instead weights each judge pair by
its number of eligible cells.

\subsection{Per-attribute agreement within an item.}
For item $i$ and attribute $k$, let $a_{ik}$ and $b_{ik}$ count stable
judges pointing to $A$ and $B$, respectively, and let
$n_{ik}=a_{ik}+b_{ik}$. Agreement among unordered judge pairs is
\begin{equation}
G_{ik}=\frac{\binom{a_{ik}}{2}+\binom{b_{ik}}{2}}{\binom{n_{ik}}{2}},
\qquad n_{ik}\ge2.
\label{eq:app-axis-agreement}
\end{equation}
An attribute is comparable only when at least two judges give stable
directions. It is unanimous when $G_{ik}=1$.

\subsection{Pooled and unweighted attribute agreement.}
Let $\mathcal{K}_i=\{k:n_{ik}\ge2\}$ be the comparable attributes of an item.
The pooled agreement reported in case studies is
\begin{equation}
G_i^{\mathrm{pool}}=
\frac{\displaystyle\sum_{k\in\mathcal{K}_i}
 [\binom{a_{ik}}{2}+\binom{b_{ik}}{2}]}{
 \displaystyle\sum_{k\in\mathcal{K}_i}\binom{n_{ik}}{2}}.
\end{equation}
Equivalently, it averages $G_{ik}$ with weight $\binom{n_{ik}}{2}$.
The unweighted mean is
$G_i^{\mathrm{macro}}=|\mathcal{K}_i|^{-1}\sum_{k\in\mathcal{K}_i}G_{ik}$.
The latter gives an attribute with two stable judges the same weight as
one with nineteen, whereas the pooled statistic weights individual
judge-pair comparisons equally.

\subsection{Within- and across-verdict camps.}
For a single item, judges can be grouped by their overall choice. We
compute attribute agreement within each group or across the two groups
by restricting the eligible judge pairs in the pooled calculation.
Overall verdict agreement is $1$ within a camp and $0$ across camps.
For example, the workout case has 8 judges choosing $A$ and 11 choosing
$B$, producing $\binom{8}{2}=28$ within-$A$ pairs,
$\binom{11}{2}=55$ within-$B$ pairs, and $8\times11=88$ across-camp pairs.
The all-pair verdict agreement is $83/171=0.485$; this is distinct from
the across-camp verdict agreement of $0$.

\subsection{Chance reference and interpretation.}
Independent, equally likely binary directions agree with probability
$0.5$, which provides the balanced-random reference in the agreement
plots.

\subsection{Reference-label accuracy and adaptation budgets}
\label{app:adaptation-evaluation}

Let $\mathcal{T}$ be the common set of 11,765 held-out non-tie pairs,
$y_i\in\{A,B\}$ their reference labels, and $\hat y_i$ a method's predicted
verdict. We compute
\begin{equation}
\operatorname{Acc}=\frac{1}{|\mathcal{T}|}
\sum_{i\in\mathcal{T}}\mathbb{1}[\hat y_i=y_i].
\end{equation}
An unparsed verdict contributes zero to the numerator and remains in the
denominator. All methods in the controlled SFT and rubric comparisons use the same
test items. The improvement
in percentage points over an unadapted judge is
$\Delta=100(\operatorname{Acc}_{\mathrm{adapted}}-
\operatorname{Acc}_{\mathrm{base}})$.

Adaptation budgets are specified \emph{per source}, not for the entire
collection. Budgets of 32, 64, 128, and 256 labeled pairs per source
correspond to 544, 1,088, 2,176, and 4,352 training pairs in total,
respectively. The Full setting uses all 35,156 eligible training pairs.
Reweighting is fitted separately for each source, matching the per-source
SFT setup. Training examples and test examples remain in their respective
splits.

The main SFT comparison uses the frozen, SFT-matched training subsets for
all four judges. Every reweighting fit uses only the 87 attribute directions.
Tables~\ref{tab:sft-vs-reweighting} and~\ref{tab:rubric-comparison} report matched-subset results. The separate sample-efficiency experiment in Figure~\ref{fig:sample-efficiency-main} uses nine judges on three sources and ten training-subset resamples. Its curves average judges and its shading reports standard deviation across resamples, not confidence intervals. Those curves and the matched SFT table use different training-subset protocols. SFT was run once per reported model--budget condition;
these tables do not provide repeated-run uncertainty.
Appendix~\ref{app:adaptation-controls} gives the rubric construction and the
definitions of its diagnostic metrics.

The broad 21-judge evaluation reports macro averages over judge--distribution cells. Its rounded accuracy and gain matrices
are retained in Appendix~\ref{app:alignment}.

\clearpage
\section{Perception: Agreement and Robustness}
\label{app:hc-judge}

This appendix supplements Sections~\ref{sec:hidden-consensus} and~\ref{sec:different-weights}. Unless stated otherwise, results use the 19-judge analysis population, 50,013 response pairs, 87 attributes, and all 171 unordered judge pairs.

\subsection{Order consistency and conditional agreement}
\label{app:hc-protocol}
For item $i$, attribute $a$, and judge $j$, let $s_{jia}^{(1)},s_{jia}^{(2)}\in\{-1,0,+1\}$ be direction readings after mapping both presentation orders back to the original responses. Zero denotes no resolved direction. Define
\begin{equation}
d_{jia}=s_{jia}^{(1)}\,\mathbb{1}\!\left[s_{jia}^{(1)}=s_{jia}^{(2)}\ne0\right].
\end{equation}
For judges $j,k$, let $I_{jk}^{+}$ and $I_{jk}^{-}$ contain items on which both give a definite overall verdict and choose the same or opposite winner. For either stratum $I$, define $R_{jkia}=\mathbb{1}[d_{jia}\ne0,d_{kia}\ne0]$. Conditional agreement and joint coverage are
\begin{equation}
 A_{jk}(I)=\frac{\sum_{i\in I}\sum_a R_{jkia}\mathbb{1}[d_{jia}=d_{kia}]}{\sum_{i\in I}\sum_a R_{jkia}},\qquad
 C_{jk}(I)=\frac{\sum_{i\in I}\sum_a R_{jkia}}{87|I|}.
\end{equation}
We average these rates equally over judge pairs. Within a judge pair, attribute agreement pools resolved item--attribute comparisons, rather than averaging attributes equally. Mutual abstentions do not contribute agreement. Per-judge summaries average over the other 18 judges; per-distribution summaries retain the same pairwise aggregation. Counts summed over judge pairs are repeated item--pair observations, not distinct response pairs.

The single-read control replaces $d$ with the first-order reading. Holistic preferences have one presentation order and no within-judge retest, so the stable-attribute comparison is not a matched-noise comparison between tasks. The balanced-random reference is 0.5; empirical marginal imbalance is instead handled by Cohen's $\kappa=(p_o-p_e)/(1-p_e)$, with $p_e$ computed from the relevant direction marginals. Mean pairwise $\kappa$ need not equal $2\bar A-1$.

\begin{table}[t]
\centering
\caption{Judge-pair strata under stable and single-read protocols. Coverage is the fraction of jointly resolved item--attribute cells. The 21-judge sensitivity analysis includes the two smallest models.\newline}

\label{tab:hc-protocols}
\small
\setlength{\tabcolsep}{4pt}
\begin{tabular}{llrrrrr}
\toprule
Judges & Read & Pairs & Same agr. & Same cov. & Opp. agr. & Opp. cov. \\
\midrule
19 & stable & 171 & 0.919 & 0.304 & 0.859 & 0.220 \\
19 & single & 171 & 0.785 & 0.529 & 0.695 & 0.480 \\
21 & stable & 210 & 0.895 & 0.260 & 0.836 & 0.190 \\
21 & single & 210 & 0.732 & 0.552 & 0.685 & 0.511 \\
\bottomrule
\end{tabular}
\end{table}
The 19-judge stable opposite-winner agreements span 0.718--0.983, with all 171 pairs above 0.5. For 21 judges, the range is 0.626--0.983 and all 210 pairs remain above 0.5. Corresponding single-read counts are 169/171 and 207/210. The additional models have stable coverage of 0.165 and 0.106, compared with 0.274--0.640 for the retained judges. Mistral's holistic task also uses an additional default system prompt, which may contribute to its differences from other judges.

When abstention is retained as a third category instead of conditioning on resolved directions, the pattern reverses: mean three-category $\kappa$ is 0.491 for preference, versus 0.425 for stable attributes and 0.431 for single-read attributes. Thus, substantial disagreement remains about whether an attribute receives a direction at all.

\subsection{Agreement by judge and source}
\label{app:agreement-by-source}
\begin{table}[t]
\centering
\caption{Stable attribute agreement and joint coverage for each judge, averaged over its 18 partners. R1 and V4 denote DeepSeek models.\newline}

\label{tab:hc-judge-strata}
\small
\setlength{\tabcolsep}{4pt}
\begin{tabular}{lrrrr}
\toprule
Judge & Same agr. & Same cov. & Opp. agr. & Opp. cov. \\
\midrule
Qwen3.5-27B & 0.945 & 0.298 & 0.897 & 0.216 \\
V4-Flash & 0.945 & 0.264 & 0.897 & 0.177 \\
Qwen3.5-122B & 0.935 & 0.322 & 0.889 & 0.245 \\
Llama-4-Scout & 0.937 & 0.295 & 0.885 & 0.185 \\
Qwen3.6-27B & 0.935 & 0.342 & 0.882 & 0.257 \\
Gemma-4-26B & 0.926 & 0.306 & 0.879 & 0.236 \\
Qwen3.5-4B & 0.935 & 0.223 & 0.876 & 0.125 \\
Gemma-4-31B & 0.917 & 0.297 & 0.872 & 0.236 \\
R1-32B & 0.915 & 0.326 & 0.868 & 0.247 \\
Qwen3.5-397B & 0.930 & 0.310 & 0.865 & 0.232 \\
Kimi-K3 & 0.916 & 0.328 & 0.865 & 0.258 \\
Qwen3.5-9B & 0.914 & 0.280 & 0.857 & 0.188 \\
R1-14B & 0.911 & 0.312 & 0.854 & 0.235 \\
V4-Pro & 0.917 & 0.311 & 0.853 & 0.206 \\
Gemma-3-27B & 0.929 & 0.360 & 0.846 & 0.248 \\
Llama-4-Maverick & 0.915 & 0.368 & 0.840 & 0.258 \\
Muse-Glimmer-30B & 0.877 & 0.342 & 0.839 & 0.285 \\
Mistral-24B & 0.898 & 0.272 & 0.804 & 0.169 \\
R1-7B & 0.861 & 0.220 & 0.762 & 0.171 \\
\bottomrule
\end{tabular}
\end{table}
\begin{table}[t]
\centering
\caption{Stable conditional agreement by distribution. Counts are distinct response pairs before verdict filtering; coverage conditions on the corresponding winner stratum.\newline}

\label{tab:hc-distributions}
\small
\setlength{\tabcolsep}{4pt}
\begin{tabular}{lrrrrr}
\toprule
Distribution & Pairs & Same agr. & Same cov. & Opp. agr. & Opp. cov. \\
\midrule
MT-Bench & 4,000 & 0.937 & 0.311 & 0.897 & 0.229 \\
SHP & 4,000 & 0.928 & 0.411 & 0.874 & 0.293 \\
RewardBench & 4,481 & 0.930 & 0.351 & 0.873 & 0.279 \\
Arena-Human & 4,000 & 0.918 & 0.325 & 0.869 & 0.248 \\
Nectar & 2,500 & 0.923 & 0.307 & 0.867 & 0.234 \\
HH-RLHF & 4,000 & 0.907 & 0.329 & 0.867 & 0.276 \\
RM-Bench & 1,146 & 0.895 & 0.308 & 0.864 & 0.263 \\
HelpSteer2 & 4,000 & 0.917 & 0.284 & 0.855 & 0.213 \\
UltraFeedback & 2,500 & 0.918 & 0.349 & 0.853 & 0.263 \\
WebGPT & 4,000 & 0.916 & 0.273 & 0.850 & 0.205 \\
PPE-Human & 4,000 & 0.918 & 0.289 & 0.849 & 0.227 \\
Arena-Expert & 1,797 & 0.913 & 0.345 & 0.849 & 0.252 \\
RewardBench 2 & 1,657 & 0.901 & 0.347 & 0.847 & 0.272 \\
PPE-Best-of-K & 2,508 & 0.908 & 0.191 & 0.846 & 0.135 \\
Summarization & 4,000 & 0.903 & 0.151 & 0.822 & 0.093 \\
IF-RewardBench & 807 & 0.876 & 0.159 & 0.821 & 0.142 \\
JudgeBench & 617 & 0.919 & 0.157 & 0.817 & 0.090 \\
\bottomrule
\end{tabular}
\end{table}
Opposite-winner attribute agreement ranges from 0.817 to 0.897 across distributions; all 171 judge pairs exceed 0.5 in each distribution.

\subsection{Matched items and single-read controls}
\label{app:matched-core}
For each judge pair, matched items require two definite overall preferences and at least one jointly resolved attribute. On this fixed item set, overall preference agreement contributes one decision per item, whereas attribute agreement pools multiple resolved cells. Averaging each item's attribute agreement before averaging items gives 0.892 under the stable protocol, compared with 0.907 from cell pooling.

The core set is selected without using cross-judge agreement. Both within-judge order consistency and detection rate must be at least their across-attribute medians, 0.778 and 0.482. The resulting 20 attributes are breadth, completeness, conceptual complexity, decomposition quality, depth, detail level, example use, explanatory depth, informativeness, justification quality, linguistic complexity, organization, qualification, reasoning completeness, relevance, rigor, robustness, specificity, strategy quality, and terminological precision.

\begin{table}[t]
\centering
\caption{Per-judge agreement on matched items under the stable protocol. Attribute agreement remains conditional on jointly resolved directions. The final row is the mean over all 171 judge pairs.\newline}

\label{tab:matched-core}
\small
\setlength{\tabcolsep}{4pt}
\begin{tabular}{lrrrrr}
\toprule
Judge & Items & Preference & All 87 & Core 20 & Cov. 87 \\
\midrule
R1-7B & 44,590 & 0.665 & 0.833 & 0.872 & 0.211 \\
R1-14B & 45,796 & 0.744 & 0.899 & 0.928 & 0.299 \\
R1-32B & 45,220 & 0.767 & 0.906 & 0.934 & 0.314 \\
V4-Flash & 44,358 & 0.784 & 0.936 & 0.957 & 0.260 \\
V4-Pro & 46,048 & 0.795 & 0.907 & 0.945 & 0.302 \\
Gemma-3-27B & 46,970 & 0.755 & 0.913 & 0.943 & 0.341 \\
Gemma-4-26B & 46,029 & 0.794 & 0.918 & 0.945 & 0.297 \\
Gemma-4-31B & 46,943 & 0.791 & 0.908 & 0.938 & 0.290 \\
Kimi-K3 & 46,739 & 0.782 & 0.905 & 0.939 & 0.319 \\
Llama-4-Scout & 45,793 & 0.751 & 0.928 & 0.950 & 0.282 \\
Llama-4-Maverick & 45,127 & 0.786 & 0.902 & 0.940 & 0.354 \\
Mistral-24B & 42,732 & 0.737 & 0.880 & 0.930 & 0.273 \\
Muse-Glimmer-30B & 46,808 & 0.780 & 0.869 & 0.916 & 0.334 \\
Qwen3.5-4B & 40,881 & 0.774 & 0.926 & 0.946 & 0.223 \\
Qwen3.5-9B & 43,850 & 0.784 & 0.904 & 0.927 & 0.276 \\
Qwen3.5-27B & 46,208 & 0.801 & 0.936 & 0.961 & 0.291 \\
Qwen3.5-122B & 46,182 & 0.781 & 0.926 & 0.950 & 0.313 \\
Qwen3.5-397B & 45,897 & 0.788 & 0.917 & 0.942 & 0.301 \\
Qwen3.6-27B & 46,811 & 0.794 & 0.924 & 0.950 & 0.332 \\
Mean & 45,420 & 0.771 & 0.907 & 0.937 & 0.295 \\
\bottomrule
\end{tabular}
\end{table}
The stable gap is positive for all 171 judge pairs and all 19 per-judge averages. Under a single read, overall agreement is 0.767, all-87 agreement is 0.765, and core-20 agreement is 0.810. Only 95/171 pairs and 12/19 per-judge means have a positive all-87 gap. The attribute advantage therefore depends on stable filtering.

\subsection{Single-read controls on matched items.}
\label{app:agreement-protocol-control}
The main text reports order-consistent attribute comparisons. Table~\ref{tab:agreement-protocol-control} retains the corresponding first-order control: the all-attribute advantage is absent under this protocol. This sensitivity does not invalidate the main conditional result, but limits its scope to the reliable directions that both judges resolve.

\begin{table}[t]
\centering
\caption{\textbf{Stable attributes and selected human attributes show higher agreement than overall preferences.} All values are percentages; $\Delta$ is attribute minus preference agreement, in percentage points. LLM comparisons use matched items within each protocol. Human agreement pools seven raters on 23 selected cases, using two or three attributes per item. All attribute rates condition on jointly resolved directions. The single-read all-attribute control does not show an advantage. Core attributes are selected for within-judge consistency and coverage.\newline}

\label{tab:agreement-protocol-control}
\small\setlength{\tabcolsep}{5pt}
\begin{tabular}{llrrr}
\toprule
Population / protocol & Attributes & Preference & Attribute & $\Delta$ \\
\midrule
LLMs / stable & All 87 & 77.1 & 90.7 & +13.6 \\
 & Core 20 & 77.1 & 93.7 & +16.6 \\
LLMs / single read & All 87 & 76.7 & 76.5 & -0.2 \\
 & Core 20 & 76.7 & 81.0 & +4.3 \\
Humans / selected cases & 2--3 per item & 62.1 & 92.3 & +30.2 \\
\bottomrule
\end{tabular}
\end{table}

\subsection{Reference-label difficulty control.}
After excluding 3,092 response pairs with tie reference labels, we separate items where both judges match the reference, both depart from it, or the judges disagree. These strata use reference-label concordance as a proxy for difficulty.
\begin{table}[t]
\centering
\caption{Difficulty-related strata. Counts sum item observations over judge pairs.\newline}

\label{tab:hc-difficulty}
\small
\setlength{\tabcolsep}{4pt}
\begin{tabular}{lrrrr}
\toprule
Stratum & Stable agr. & Stable cov. & Single agr. & Observations \\
\midrule
Both match reference & 0.922 & 0.322 & 0.793 & 4,436,700 \\
Both depart from reference & 0.909 & 0.266 & 0.762 & 1,436,019 \\
Opposite preferences & 0.859 & 0.220 & 0.695 & 1,888,438 \\
\bottomrule
\end{tabular}
\end{table}
Agreement remains high when both judges depart from the reference.

\subsection{Human agreement and study scope}
\label{app:hc-human}
Seven raters evaluated 23 selected response pairs: 21 witnesses of model preference disagreement and two controls. Each item included two or three attribute questions, yielding 55 item--attribute cells. The set contains five items each from HH-RLHF, SHP, RewardBench, and PPE-Human, and one each from Arena, HelpSteer2, and WebGPT. Raters answered overall preference before attributes. Comparisons below pool jointly resolved rater pairs; denominators differ because of abstentions and the number of attribute questions.
\begin{table}[t]
\centering
\caption{Human pairwise agreement. Parentheses give agreeing/eligible comparisons; these comparisons are not independent samples.\newline}

\label{tab:hc-human}
\small
\setlength{\tabcolsep}{4pt}
\begin{tabular}{lrrr}
\toprule
Subset & Items & Preference agreement & Attribute agreement \\
\midrule
Disagreement witnesses & 21 & 0.576 (204/354) & 0.928 (942/1,015) \\
Controls & 2 & 1.000 (42/42) & 0.847 (61/72) \\
Pooled & 23 & 0.621 (246/396) & 0.923 (1,003/1,087) \\
\bottomrule
\end{tabular}
\end{table}
Pooled preference and attribute response rates are 0.907 and 0.969. Item-level preference agreement ranges from 0.400 to 1.000 and attribute agreement from 0.583 to 1.000. These selected witnesses establish that the qualitative pattern also occurs in human judgments; they do not estimate its prevalence in a random population. The small control set and possible anchoring from question order further limit inference.

\subsection{Complete attribute summaries for 19 judges}
\label{app:hc-all-attributes}
These tables support the 19-judge analysis in Figure~\ref{fig:hc-strata}. They include all 87 attributes. The 21-judge attribute results used in Figure~\ref{fig:attribute-agreement-selected} are reported separately in Appendix~\ref{app:attribute-consensus-detail}; the two populations yield different rates. ``Resolve'' and ``Self'' denote mean detection rate and within-judge order consistency. Agreement and coverage are means over judge pairs. A star marks membership in the reliability-selected core set.
\begin{table}[t]
\centering
\caption{Complete attribute summaries, part 1 of 2. All and opposite-winner (Opp.) rates use stable directions.\newline}

\label{tab:hc-attributes-1}
\small
\setlength{\tabcolsep}{4pt}
\begin{tabular}{lrrrrrr}
\toprule
Attribute & Resolve & Self & All agr. & All cov. & Opp. agr. & Opp. cov. \\
\midrule
verbosity & 0.701 & 0.762 & 0.965 & 0.558 & 0.962 & 0.517 \\
moralizing tendency & 0.225 & 0.816 & 0.959 & 0.128 & 0.959 & 0.121 \\
technicality & 0.466 & 0.827 & 0.965 & 0.323 & 0.956 & 0.266 \\
evidence use & 0.234 & 0.825 & 0.960 & 0.127 & 0.954 & 0.109 \\
self correction & 0.089 & 0.828 & 0.928 & 0.021 & 0.950 & 0.018 \\
conceptual complexity$^{*}$ & 0.541 & 0.819 & 0.956 & 0.381 & 0.948 & 0.314 \\
linguistic complexity$^{*}$ & 0.592 & 0.788 & 0.953 & 0.424 & 0.948 & 0.370 \\
acknowledgment of limitations & 0.278 & 0.826 & 0.950 & 0.165 & 0.947 & 0.146 \\
humor & 0.113 & 0.819 & 0.951 & 0.058 & 0.947 & 0.057 \\
analogy use & 0.172 & 0.816 & 0.952 & 0.080 & 0.946 & 0.073 \\
organization$^{*}$ & 0.537 & 0.791 & 0.960 & 0.387 & 0.944 & 0.307 \\
detail level$^{*}$ & 0.704 & 0.799 & 0.955 & 0.563 & 0.942 & 0.497 \\
harmfulness & 0.129 & 0.834 & 0.946 & 0.064 & 0.940 & 0.059 \\
example use$^{*}$ & 0.500 & 0.820 & 0.951 & 0.355 & 0.940 & 0.297 \\
vividness & 0.328 & 0.802 & 0.951 & 0.191 & 0.938 & 0.163 \\
breadth$^{*}$ & 0.612 & 0.809 & 0.946 & 0.449 & 0.935 & 0.388 \\
digression & 0.493 & 0.770 & 0.926 & 0.318 & 0.933 & 0.304 \\
enthusiasm & 0.423 & 0.780 & 0.933 & 0.264 & 0.927 & 0.229 \\
hedging & 0.371 & 0.731 & 0.925 & 0.226 & 0.927 & 0.210 \\
hallucination rate & 0.320 & 0.766 & 0.923 & 0.166 & 0.924 & 0.154 \\
specificity$^{*}$ & 0.673 & 0.798 & 0.943 & 0.518 & 0.922 & 0.443 \\
creativity & 0.398 & 0.785 & 0.935 & 0.236 & 0.921 & 0.193 \\
storytelling quality & 0.317 & 0.799 & 0.943 & 0.159 & 0.920 & 0.129 \\
depth$^{*}$ & 0.640 & 0.814 & 0.940 & 0.484 & 0.918 & 0.400 \\
explanatory depth$^{*}$ & 0.580 & 0.814 & 0.946 & 0.431 & 0.918 & 0.353 \\
notation clarity & 0.141 & 0.810 & 0.948 & 0.088 & 0.915 & 0.061 \\
decomposition quality$^{*}$ & 0.512 & 0.798 & 0.949 & 0.338 & 0.913 & 0.250 \\
uncertainty expression & 0.333 & 0.772 & 0.910 & 0.189 & 0.910 & 0.172 \\
verification & 0.213 & 0.808 & 0.931 & 0.079 & 0.906 & 0.059 \\
refusal tendency & 0.198 & 0.759 & 0.901 & 0.095 & 0.906 & 0.082 \\
justification quality$^{*}$ & 0.589 & 0.805 & 0.941 & 0.433 & 0.904 & 0.342 \\
interestingness & 0.626 & 0.770 & 0.924 & 0.455 & 0.899 & 0.381 \\
novelty & 0.281 & 0.725 & 0.907 & 0.134 & 0.897 & 0.108 \\
empathy & 0.154 & 0.780 & 0.934 & 0.066 & 0.896 & 0.053 \\
informativeness$^{*}$ & 0.689 & 0.799 & 0.939 & 0.534 & 0.895 & 0.439 \\
warmth & 0.363 & 0.788 & 0.929 & 0.210 & 0.895 & 0.163 \\
safety conservatism & 0.261 & 0.794 & 0.923 & 0.138 & 0.894 & 0.122 \\
qualification$^{*}$ & 0.517 & 0.793 & 0.928 & 0.350 & 0.894 & 0.292 \\
politeness & 0.318 & 0.793 & 0.917 & 0.179 & 0.889 & 0.142 \\
reasoning completeness$^{*}$ & 0.506 & 0.806 & 0.936 & 0.335 & 0.880 & 0.247 \\
assertiveness & 0.521 & 0.709 & 0.888 & 0.331 & 0.877 & 0.281 \\
friendliness & 0.434 & 0.740 & 0.897 & 0.260 & 0.875 & 0.213 \\
conversationality & 0.507 & 0.721 & 0.877 & 0.325 & 0.870 & 0.290 \\
formality & 0.469 & 0.718 & 0.898 & 0.284 & 0.868 & 0.221 \\
\bottomrule
\end{tabular}
\end{table}
\begin{table}[t]
\centering
\caption{Complete attribute summaries, part 2 of 2. All and opposite-winner (Opp.) rates use stable directions.\newline}

\label{tab:hc-attributes-2}
\small
\setlength{\tabcolsep}{4pt}
\begin{tabular}{lrrrrrr}
\toprule
Attribute & Resolve & Self & All agr. & All cov. & Opp. agr. & Opp. cov. \\
\midrule
sycophancy & 0.146 & 0.689 & 0.857 & 0.052 & 0.866 & 0.050 \\
confidence & 0.504 & 0.728 & 0.886 & 0.320 & 0.864 & 0.257 \\
personalization & 0.228 & 0.767 & 0.918 & 0.106 & 0.864 & 0.084 \\
insightfulness & 0.476 & 0.777 & 0.924 & 0.301 & 0.861 & 0.225 \\
persuasiveness & 0.518 & 0.776 & 0.922 & 0.337 & 0.861 & 0.253 \\
difficulty & 0.483 & 0.735 & 0.861 & 0.307 & 0.860 & 0.261 \\
formatting & 0.369 & 0.774 & 0.864 & 0.255 & 0.857 & 0.214 \\
rigor$^{*}$ & 0.590 & 0.797 & 0.923 & 0.424 & 0.843 & 0.314 \\
caution framing & 0.373 & 0.729 & 0.841 & 0.219 & 0.838 & 0.200 \\
completeness$^{*}$ & 0.524 & 0.805 & 0.935 & 0.349 & 0.838 & 0.236 \\
directness & 0.531 & 0.706 & 0.850 & 0.345 & 0.837 & 0.296 \\
elaboration & 0.659 & 0.751 & 0.851 & 0.492 & 0.835 & 0.428 \\
conciseness & 0.641 & 0.691 & 0.825 & 0.445 & 0.827 & 0.408 \\
originality & 0.488 & 0.682 & 0.854 & 0.293 & 0.819 & 0.238 \\
groundedness & 0.324 & 0.746 & 0.906 & 0.164 & 0.819 & 0.126 \\
relevance$^{*}$ & 0.482 & 0.794 & 0.929 & 0.312 & 0.808 & 0.201 \\
terminological precision$^{*}$ & 0.501 & 0.800 & 0.911 & 0.331 & 0.804 & 0.228 \\
abstraction level & 0.541 & 0.673 & 0.779 & 0.333 & 0.794 & 0.297 \\
coherence & 0.590 & 0.772 & 0.907 & 0.414 & 0.793 & 0.291 \\
neutrality & 0.186 & 0.744 & 0.876 & 0.070 & 0.777 & 0.052 \\
accessibility & 0.538 & 0.687 & 0.831 & 0.333 & 0.776 & 0.259 \\
goal directedness & 0.557 & 0.760 & 0.893 & 0.376 & 0.769 & 0.267 \\
forcefulness & 0.503 & 0.631 & 0.774 & 0.301 & 0.765 & 0.258 \\
instruction adherence & 0.507 & 0.758 & 0.913 & 0.328 & 0.761 & 0.204 \\
information density & 0.665 & 0.747 & 0.838 & 0.485 & 0.758 & 0.400 \\
robustness$^{*}$ & 0.528 & 0.778 & 0.897 & 0.339 & 0.752 & 0.221 \\
audience adaptation & 0.324 & 0.773 & 0.900 & 0.149 & 0.750 & 0.085 \\
fluency & 0.402 & 0.736 & 0.861 & 0.213 & 0.737 & 0.139 \\
respectfulness & 0.365 & 0.778 & 0.884 & 0.197 & 0.724 & 0.123 \\
clarity & 0.603 & 0.733 & 0.850 & 0.406 & 0.719 & 0.296 \\
constraint satisfaction & 0.143 & 0.778 & 0.903 & 0.049 & 0.717 & 0.028 \\
readability & 0.574 & 0.685 & 0.760 & 0.367 & 0.712 & 0.298 \\
reasoning efficiency & 0.479 & 0.748 & 0.804 & 0.285 & 0.707 & 0.223 \\
step correctness & 0.264 & 0.815 & 0.907 & 0.134 & 0.705 & 0.072 \\
phrasing originality & 0.483 & 0.626 & 0.738 & 0.277 & 0.691 & 0.222 \\
strategy quality$^{*}$ & 0.581 & 0.796 & 0.901 & 0.409 & 0.687 & 0.255 \\
focus & 0.523 & 0.713 & 0.789 & 0.324 & 0.686 & 0.246 \\
logical validity & 0.430 & 0.787 & 0.889 & 0.251 & 0.684 & 0.152 \\
final answer correctness & 0.370 & 0.811 & 0.914 & 0.219 & 0.664 & 0.114 \\
factual accuracy & 0.416 & 0.786 & 0.852 & 0.248 & 0.643 & 0.161 \\
assumption validity & 0.495 & 0.718 & 0.836 & 0.297 & 0.634 & 0.199 \\
calibration & 0.402 & 0.752 & 0.868 & 0.218 & 0.626 & 0.129 \\
internal consistency & 0.251 & 0.704 & 0.785 & 0.102 & 0.603 & 0.057 \\
\bottomrule
\end{tabular}
\end{table}

\section{Prioritization: Fitted Weights and Controls}
\label{app:priority-controls}
This appendix supports Section~\ref{sec:fingerprint}: it defines the fitted weights, checks their reproducibility, and separates shared-representation refitting from component-swap controls. Unless stated otherwise, these analyses use 19 judges and predict their own overall judgments, not source reference labels.

\subsection{Weight estimation and reproducibility}
\label{app:hc-priorities}
For each judge, we fit an L2-regularized logistic regression ($C=1$) from its 87 stable directions to its own holistic verdict, excluding ties and unparsed verdicts. Perception profiles are the 87-dimensional stable resolution rates. Weight profiles summarize the fitted logistic coefficients. Hence, a higher perception-profile correlation specifically indicates similar patterns of attribute resolution.

We calculate Pearson and Spearman correlations over attributes for each unordered judge pair. The identical-preference null assigns all 19 judges a common true weight vector, the mean observed weights, generates verdicts using each judge's own perception matrix, and refits the models over 20 repetitions. We report the observed mean correlation divided by the mean null correlation.
\begin{table}[t]
\centering
\caption{Profile similarity with distinct estimation-noise references. Null-corrected values use unrounded source estimates.\newline}

\label{tab:hc-profile-correction}
\small
\setlength{\tabcolsep}{4pt}
\begin{tabular}{lrrr}
\toprule
Profile / metric & Observed & Null mean & Corrected \\
\midrule
Perception / Pearson & 0.765 & 1.000 & 0.765 \\
Prioritization / Pearson & 0.609 & 0.944 & 0.645 \\
Perception / Spearman & 0.757 & 0.999 & 0.757 \\
Prioritization / Spearman & 0.462 & 0.847 & 0.545 \\
\bottomrule
\end{tabular}
\end{table}
For half-sample correlation $r_h$, Spearman--Brown reliability is $2r_h/(1+r_h)$. Raw half-sample correlations and corrected reliability are reported separately below.

\begin{table}[t]
\centering
\caption{Priority-model reproducibility and five-fold prediction accuracy. Raw half correlations and Spearman--Brown (SB) reliability are distinct quantities. CV columns predict each judge's own preferences.\newline}

\label{tab:hc-reliability}
\small
\setlength{\tabcolsep}{4pt}
\begin{tabular}{lrrrrr}
\toprule
Judge & Raw half r & SB & Shared SB & Own CV & Shared CV \\
\midrule
R1-7B & 0.868 & 0.929 & 0.923 & 0.718 & 0.706 \\
R1-14B & 0.905 & 0.950 & 0.871 & 0.805 & 0.798 \\
R1-32B & 0.920 & 0.958 & 0.900 & 0.824 & 0.815 \\
V4-Flash & 0.923 & 0.960 & 0.897 & 0.796 & 0.823 \\
V4-Pro & 0.924 & 0.960 & 0.872 & 0.836 & 0.845 \\
Gemma-3-27B & 0.837 & 0.911 & 0.930 & 0.846 & 0.836 \\
Gemma-4-26B & 0.944 & 0.971 & 0.933 & 0.843 & 0.825 \\
Gemma-4-31B & 0.956 & 0.977 & 0.943 & 0.870 & 0.835 \\
Kimi-K3 & 0.938 & 0.968 & 0.949 & 0.858 & 0.830 \\
Llama-4-Scout & 0.918 & 0.957 & 0.913 & 0.795 & 0.794 \\
Llama-4-Maverick & 0.916 & 0.956 & 0.937 & 0.850 & 0.845 \\
Mistral-24B & 0.915 & 0.955 & 0.914 & 0.794 & 0.806 \\
Muse-Glimmer-30B & 0.973 & 0.986 & 0.939 & 0.857 & 0.826 \\
Qwen3.5-4B & 0.876 & 0.934 & 0.878 & 0.775 & 0.800 \\
Qwen3.5-9B & 0.871 & 0.931 & 0.891 & 0.787 & 0.803 \\
Qwen3.5-27B & 0.894 & 0.944 & 0.947 & 0.845 & 0.844 \\
Qwen3.5-122B & 0.847 & 0.917 & 0.915 & 0.805 & 0.805 \\
Qwen3.5-397B & 0.898 & 0.946 & 0.948 & 0.843 & 0.823 \\
Qwen3.6-27B & 0.924 & 0.960 & 0.945 & 0.852 & 0.839 \\
\bottomrule
\end{tabular}
\end{table}
The raw priority half correlations range from 0.837 to 0.973; SB reliability ranges from 0.911 to 0.986. Perception-profile half correlations are at least 0.9998 in the source estimates.

\subsection{Refitting with shared attribute judgments}
\label{app:shared-refitting}
The shared feature is the majority direction $S_{ia}=\operatorname{sign}(\sum_j d_{jia})$, with zero for a tied or absent majority. Refitting each judge's own verdicts on this common representation yields mean cross-judge weight correlations of 0.575 (Pearson) and 0.495 (Spearman). Mean five-fold prediction accuracy is 0.820, compared with 0.822 using own perception. The remaining weight diversity is therefore not solely a consequence of fitting in different feature representations.

\subsection{Component swaps and their interpretation}
\label{app:swap-control}
This control is distinct from shared-perception refitting. For the weight swap, Qwen3.5-27B is the fixed donor: the target judge's 87 attribute coefficients are replaced by the donor's fitted coefficients while the target's attribute readings are retained. The perception swap instead replaces item-level attribute readings while holding the target's fitted weights fixed.

The reported in-sample prediction accuracies are 0.822 with own perception and weights, 0.739 after the perception swap, and 0.799 after the weight swap. Thus, perception replacement is more disruptive in this particular control, reducing accuracy by 8.3 percentage points versus 2.3 for weight replacement.

The swaps act on different objects and are not matched in perturbation size. Perception replacement changes an item-by-attribute matrix, whereas weight replacement changes a single 87-dimensional aggregation vector. Their effects also depend on donor--target differences and compatibility between the substituted component and the retained component. Consequently, the accuracy drops measure sensitivity to these replacements, not the relative causal contributions of perception and prioritization.

\subsection{Estimation noise and source-specific fits}
\label{app:weight-noise}
Before measuring coefficient dispersion, each judge's weights are divided by their mean absolute magnitude across all 87 attributes. Observed cross-judge SD exceeds the null 95th percentile on 85 of 87 attributes; depth and friendliness are the exceptions. The median observed/null-mean SD ratio is 2.16. The table reports the 15 attributes with the largest observed/null-mean ratios.
\begin{table}[t]
\centering
\caption{Detailed null comparison for the 15 selected attributes. Ratio divides observed SD by mean null SD, not by the null 95th percentile.\newline}

\label{tab:hc-null}
\small
\setlength{\tabcolsep}{4pt}
\begin{tabular}{lrrrr}
\toprule
Attribute & Observed SD & Null mean SD & Null q95 & Ratio \\
\midrule
strategy quality & 1.874 & 0.316 & 0.400 & 5.93 \\
harmfulness & 2.466 & 0.479 & 0.628 & 5.15 \\
calibration & 1.769 & 0.353 & 0.451 & 5.01 \\
relevance & 1.501 & 0.339 & 0.419 & 4.42 \\
final answer correctness & 1.600 & 0.373 & 0.490 & 4.28 \\
assumption validity & 1.243 & 0.299 & 0.414 & 4.15 \\
informativeness & 1.010 & 0.269 & 0.345 & 3.75 \\
directness & 0.891 & 0.244 & 0.288 & 3.66 \\
instruction adherence & 1.040 & 0.286 & 0.364 & 3.63 \\
verbosity & 0.889 & 0.249 & 0.304 & 3.57 \\
groundedness & 0.960 & 0.271 & 0.348 & 3.55 \\
focus & 1.002 & 0.283 & 0.352 & 3.54 \\
clarity & 0.937 & 0.274 & 0.337 & 3.42 \\
persuasiveness & 0.949 & 0.281 & 0.359 & 3.38 \\
digression & 0.857 & 0.259 & 0.316 & 3.31 \\
\bottomrule
\end{tabular}
\end{table}
Within each distribution, we refit priorities and estimate separate split-half ceilings. These ratios use a different correction from the global identical-preference null. Perception ratios exceed priority ratios in all 17 distributions, but small ceilings make per-distribution ratios noisy.
\begin{table}[t]
\centering
\caption{Distribution-specific profile correlations and split-half ceilings. Ratios are computed before rounding.\newline}

\label{tab:hc-source-profiles}
\small
\setlength{\tabcolsep}{4pt}
\begin{tabular}{lrrrrrr}
\toprule
Distribution & P r & P ceiling & P ratio & W r & W ceiling & W ratio \\
\midrule
Nectar & 0.772 & 0.998 & 0.773 & 0.305 & 0.525 & 0.581 \\
HelpSteer2 & 0.770 & 0.999 & 0.771 & 0.320 & 0.599 & 0.534 \\
RewardBench & 0.766 & 0.999 & 0.767 & 0.393 & 0.702 & 0.560 \\
MT-Bench & 0.763 & 0.999 & 0.764 & 0.175 & 0.780 & 0.224 \\
UltraFeedback & 0.762 & 0.998 & 0.763 & 0.238 & 0.410 & 0.582 \\
WebGPT & 0.762 & 0.999 & 0.763 & 0.170 & 0.384 & 0.443 \\
PPE-Human & 0.757 & 0.999 & 0.758 & 0.306 & 0.552 & 0.555 \\
Arena-Human & 0.754 & 0.999 & 0.755 & 0.299 & 0.576 & 0.518 \\
IF-RewardBench & 0.746 & 0.991 & 0.753 & 0.108 & 0.193 & 0.556 \\
RM-Bench & 0.736 & 0.997 & 0.738 & 0.245 & 0.526 & 0.466 \\
RewardBench 2 & 0.734 & 0.997 & 0.736 & 0.279 & 0.409 & 0.682 \\
PPE-Best-of-K & 0.727 & 0.998 & 0.728 & 0.206 & 0.454 & 0.452 \\
Arena-Expert & 0.723 & 0.997 & 0.725 & 0.195 & 0.365 & 0.534 \\
SHP & 0.718 & 0.999 & 0.719 & 0.210 & 0.490 & 0.429 \\
JudgeBench & 0.700 & 0.992 & 0.705 & 0.107 & 0.225 & 0.474 \\
HH-RLHF & 0.676 & 0.999 & 0.677 & 0.409 & 0.640 & 0.638 \\
Summarization & 0.598 & 0.999 & 0.598 & 0.127 & 0.395 & 0.323 \\
\bottomrule
\end{tabular}
\end{table}

\begin{figure}[t]
\centering
\includegraphics[width=\linewidth]{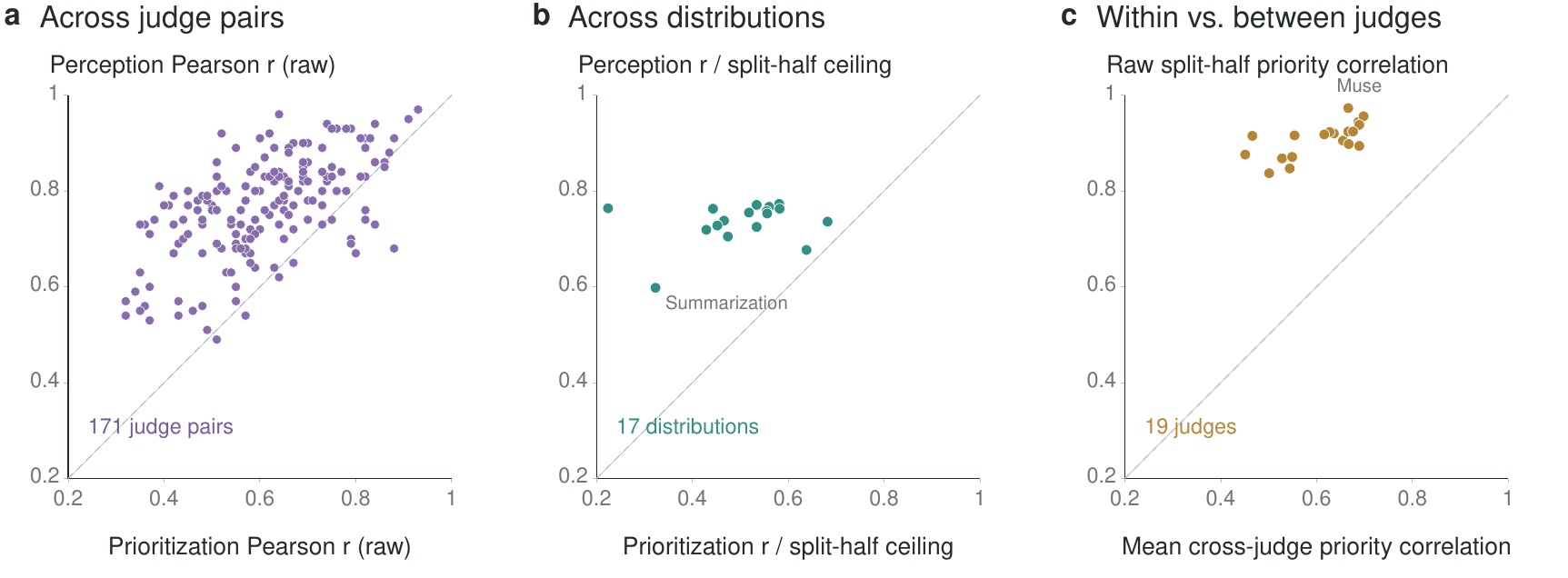}
\caption{\textbf{Shared resolution patterns and reproducible differences in priorities.} (a) Raw cross-judge profile correlations for all 171 judge pairs; perception profiles contain stable attribute resolution rates. Global null-corrected Pearson correlations are 0.765 for perception and 0.645 for priorities. (b) Correlations divided by their separately estimated split-half ceilings within each of 17 distributions. This is a different correction from the global null. (c) For each judge, raw split-half priority correlation versus its mean cross-judge priority correlation. Dashed lines indicate equality. Each point represents an observed pair, distribution, or judge, respectively.}
\label{fig:profile-diagnostics}
\end{figure}

\subsection{What Do Judges and Reference Labels Favor?}
\label{sec:what-judges-favor}
\subsection{Judges assign different priorities to shared attributes.}
With the same attribute representation, all 19 judges assign positive weights to correctness and strategy quality (Figure~\ref{fig:s5-compact}a). Their relative priorities differ: Gemma-4-31B weights instruction adherence above correctness, whereas Gemma-3-27B weights empathy above correctness.

\subsection{Reference-label associations depend on the dataset.}
Correctness predicts the reference winner above 50\% in all 17 datasets (Figure~\ref{fig:s5-compact}b). Style associations can reverse: choosing the more verbose response matches Nectar labels 79\% of the time but PPE Best-of-K labels only 27\%; conciseness shows the opposite pattern (34\% versus 78\%). Formatting also reverses. These differences reflect both reference targets and response-pair construction, so alignment must specify which target and distribution it serves.

\subsection{Attribute-priority display details.}\label{app:priority-display}

\begin{figure}[t]
\centering
\includegraphics[width=\linewidth]{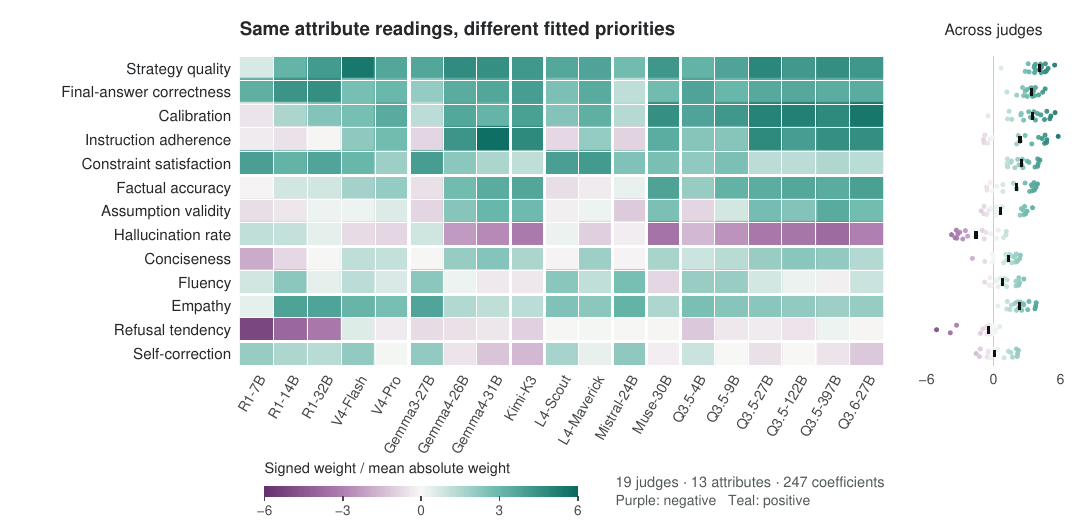}
\caption{\textbf{Shared measurements reveal different priorities.} Normalized logistic coefficients for 19 judges on the same attribute representation; 13 attributes selected by combined importance and variation. Purple/teal indicate negative/positive conditional associations. Right: individual judges and median ticks. Selection, normalization, and the 21-judge check are in Appendix~\ref{app:priority-display}.}
\label{fig:attribute-priorities}
\end{figure}

\begin{figure}[t]
\centering
\includegraphics[width=\linewidth]{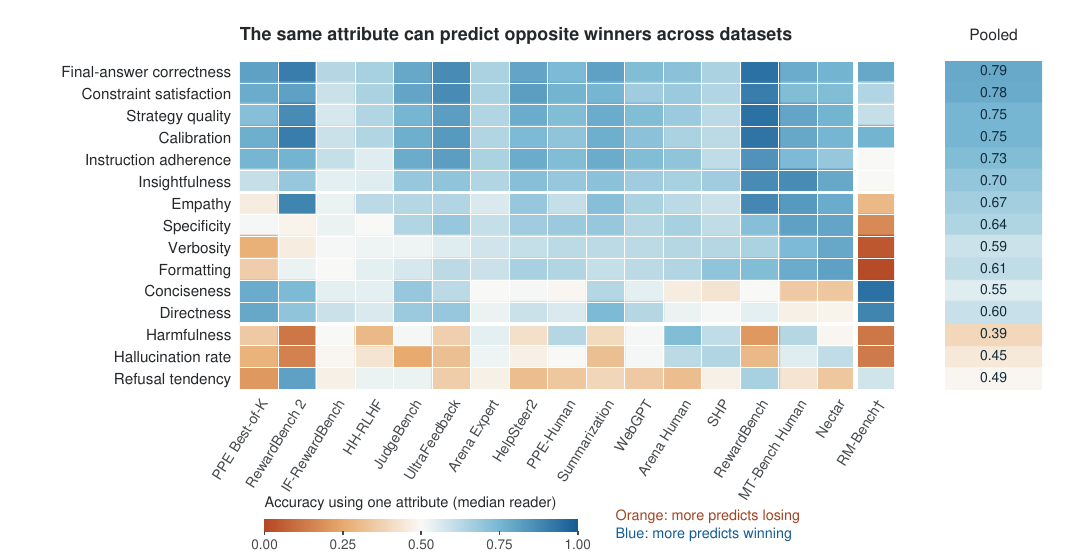}
\caption{\textbf{Attribute--label associations depend on the distribution.} Each cell is median accuracy across eligible readers when choosing the response with more of one attribute (21 readers; 15 attributes, 17 distributions). Orange/blue indicate below/above 50\%. The pooled column hides source-specific reversals. Sources are ordered by verbosity association, with the style-confounded RM-Bench subset ($\dagger$) separated at right. Coverage and full pooled statistics: Appendix~\ref{app:priority-display}.}
\label{fig:reference-by-source}
\end{figure}

For each response pair and attribute, we take the sign of the summed stable reader directions to obtain a shared representation. Every judge is fitted on this same representation using L2-regularized logistic regression ($C=1$), with that judge's own overall choices as labels; ties and unparsed verdicts are excluded. The full collection supplies 50,013 pairs before these exclusions. We normalize each fitted coefficient by the judge's mean absolute coefficient across all 87 attributes:
\[
\widetilde w_{ja}=w_{ja}\Big/\left(\frac{1}{87}\sum_{b=1}^{87}|w_{jb}|\right).
\]

For Figure~\ref{fig:attribute-priorities}, attributes are ranked by mean absolute normalized weight and by cross-judge standard deviation. We retain the 13 with the largest sum of ranks, with larger ranks denoting larger statistics. Rows are arranged for readability, not by effect magnitude; columns follow model groups. The symmetric color scale spans $[-6,6]$ without clipping. Each right-hand dot is one judge; a black tick marks the median, not a confidence interval. Displayed coefficients are rounded to two decimal places. R1/V4 denote DeepSeek, L4 denotes Llama-4, and Q denotes Qwen; model sizes in mixture-of-experts names are total parameters.

\subsection{Sensitivity to the judge set.}
\begin{figure}[t]
\centering
\includegraphics[width=\linewidth]{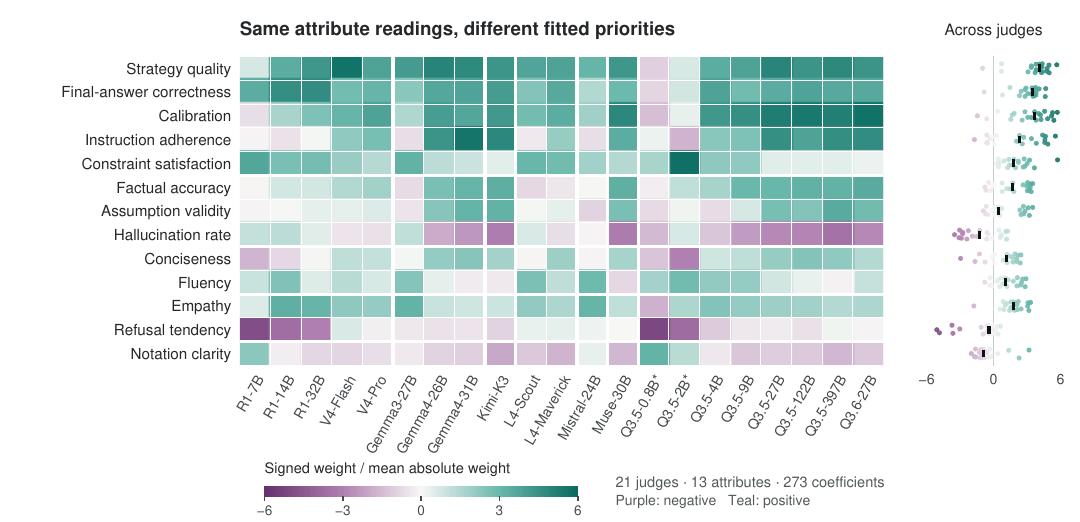}
\caption{\textbf{Shared-representation priorities with all 21 judges.} The selection and normalization procedure matches Figure~\ref{fig:attribute-priorities}. Asterisks mark the two judges with low stable coverage. Dots show individual normalized weights; black ticks show medians.}
\label{fig:priorities-21-check}
\end{figure}

The main priority display uses the 19-judge set from Section~\ref{sec:hidden-consensus}. Figure~\ref{fig:priorities-21-check} repeats the procedure with all 21 judges. The selected attribute sets share 12 axes; notation clarity replaces self-correction in the 21-judge display. The Gemma instruction-adherence and empathy contrasts in the main text hold in both versions. Qwen3.5-0.8B and Qwen3.5-2B have low stable coverage (16.5\% and 10.6\%); their fitted priorities warrant greater caution. Shared-representation cross-validation accuracy is 55.9\% for Qwen3.5-0.8B, compared with 69.4--84.5\% for the other judges.

\subsection{Source-specific associations are not fitted priorities.}
For each attribute and reader, we predict the reference winner by choosing the response with more of that attribute, using only stable directional readings and non-tie reference labels. In each source, a reader enters the median only with more than 50 usable comparisons. Figure~\ref{fig:reference-by-source} shows 15 attributes spanning task quality, presentation, interpersonal qualities, and potentially undesirable response characteristics. Sources are sorted by their verbosity association, except RM-Bench, which is placed last because its sampled chosen-style-0/rejected-style-2 construction strongly opposes length and formatting to the target label. The reversals are not exclusive to RM-Bench: verbosity is 0.79 on Nectar versus 0.27 on PPE Best-of-K, and conciseness is 0.34 versus 0.78.

The pooled column is recomputed over the pooled collection, not a mean of the source medians. For pooled statistics, readers require more than 200 usable comparisons. Tables~\ref{tab:reference-signals-main}--\ref{tab:reference-signals-rest} report all 87 attributes in decreasing association strength. Strength is the median reader-level $|\mathrm{accuracy}-0.5|$; sign is determined by whether median accuracy is at least 0.5. Coverage is mean stable coverage across the 21 readers. Comparisons are conditional on attribute-specific coverage and need not use the same response pairs. For example, constraint satisfaction has a strong pooled association but only 16.0\% coverage.

\begin{table}[t]
\centering\footnotesize\setlength{\tabcolsep}{6pt}\renewcommand{\arraystretch}{1.02}
\caption{\textbf{Pooled single-attribute associations.} Coverage and accuracy are percentages; strength is median $|\mathrm{accuracy}-0.5|$. All values condition on a stable direction.\newline}

\label{tab:reference-signals-main}
\begin{tabular}{@{}lrrrr@{}}\toprule Attribute & Coverage & Accuracy & Strength & Sign \\ \midrule
Final answer correctness & 34.9 & 78.8 & 0.288 & $+$ \\
Constraint satisfaction & 16.0 & 78.4 & 0.284 & $+$ \\
Step correctness & 27.0 & 76.0 & 0.260 & $+$ \\
Strategy quality & 54.1 & 75.3 & 0.253 & $+$ \\
Calibration & 37.7 & 75.2 & 0.252 & $+$ \\
Logical validity & 41.9 & 74.9 & 0.249 & $+$ \\
Internal consistency & 25.2 & 73.8 & 0.238 & $+$ \\
Factual accuracy & 40.1 & 73.7 & 0.237 & $+$ \\
Respectfulness & 34.5 & 73.4 & 0.234 & $+$ \\
Assumption validity & 46.8 & 73.1 & 0.231 & $+$ \\
Robustness & 48.7 & 72.8 & 0.228 & $+$ \\
Instruction adherence & 49.6 & 72.7 & 0.227 & $+$ \\
Fluency & 37.8 & 72.5 & 0.225 & $+$ \\
Audience adaptation & 30.1 & 72.4 & 0.224 & $+$ \\
Relevance & 46.9 & 72.1 & 0.221 & $+$ \\
Completeness & 49.5 & 71.3 & 0.213 & $+$ \\
Coherence & 55.8 & 71.2 & 0.212 & $+$ \\
Terminological precision & 46.6 & 70.8 & 0.208 & $+$ \\
Clarity & 57.3 & 70.6 & 0.206 & $+$ \\
Persuasiveness & 48.1 & 70.3 & 0.203 & $+$ \\
Groundedness & 31.3 & 70.0 & 0.200 & $+$ \\
Rigor & 54.8 & 69.9 & 0.199 & $+$ \\
Goal directedness & 52.3 & 69.9 & 0.199 & $+$ \\
Insightfulness & 44.2 & 69.7 & 0.197 & $+$ \\
Reasoning completeness & 48.0 & 69.3 & 0.193 & $+$ \\
Justification quality & 55.1 & 68.5 & 0.185 & $+$ \\
Informativeness & 64.8 & 67.8 & 0.178 & $+$ \\
Notation clarity & 14.9 & 67.5 & 0.175 & $+$ \\
Decomposition quality & 47.7 & 67.3 & 0.173 & $+$ \\
Information density & 61.5 & 67.0 & 0.170 & $+$ \\
Empathy & 14.9 & 66.7 & 0.167 & $+$ \\
Personalization & 21.9 & 66.3 & 0.163 & $+$ \\
Accessibility & 49.5 & 66.1 & 0.161 & $+$ \\
Warmth & 34.0 & 66.1 & 0.161 & $+$ \\
Explanatory depth & 54.1 & 66.1 & 0.161 & $+$ \\
Formality & 43.9 & 66.0 & 0.160 & $+$ \\
Interestingness & 58.2 & 65.9 & 0.159 & $+$ \\
Storytelling quality & 30.3 & 65.8 & 0.158 & $+$ \\
Reasoning efficiency & 44.8 & 65.7 & 0.157 & $+$ \\
Neutrality & 18.0 & 65.6 & 0.156 & $+$ \\
Confidence & 47.2 & 65.2 & 0.152 & $+$ \\
Organization & 50.6 & 65.2 & 0.152 & $+$ \\
Evidence use & 22.8 & 65.1 & 0.151 & $+$ \\
Conceptual complexity & 50.4 & 64.9 & 0.149 & $+$ \\
\bottomrule\end{tabular}\end{table}

\begin{table}[t]
\centering\footnotesize\setlength{\tabcolsep}{6pt}\renewcommand{\arraystretch}{1.02}
\caption{\textbf{Pooled single-attribute associations (continued).} Coverage and accuracy are percentages; strength is median $|\mathrm{accuracy}-0.5|$. All values condition on a stable direction.\newline}

\label{tab:reference-signals-rest}
\begin{tabular}{@{}lrrrr@{}}\toprule Attribute & Coverage & Accuracy & Strength & Sign \\ \midrule
Assertiveness & 48.7 & 64.9 & 0.149 & $+$ \\
Creativity & 38.2 & 64.8 & 0.148 & $+$ \\
Depth & 59.2 & 64.7 & 0.147 & $+$ \\
Breadth & 56.2 & 64.7 & 0.147 & $+$ \\
Elaboration & 64.9 & 64.7 & 0.147 & $+$ \\
Qualification & 48.2 & 64.6 & 0.146 & $+$ \\
Friendliness & 40.6 & 64.3 & 0.143 & $+$ \\
Novelty & 27.4 & 64.2 & 0.142 & $+$ \\
Specificity & 64.6 & 64.1 & 0.141 & $+$ \\
Originality & 45.6 & 64.0 & 0.140 & $+$ \\
Focus & 49.9 & 64.0 & 0.140 & $+$ \\
Technicality & 43.8 & 63.7 & 0.137 & $+$ \\
Vividness & 30.8 & 63.6 & 0.136 & $+$ \\
Phrasing originality & 49.1 & 63.6 & 0.136 & $+$ \\
Example use & 47.5 & 63.6 & 0.136 & $+$ \\
Detail level & 66.6 & 63.5 & 0.135 & $+$ \\
Politeness & 30.5 & 63.3 & 0.133 & $+$ \\
Verification & 21.8 & 63.3 & 0.133 & $+$ \\
Linguistic complexity & 55.9 & 63.1 & 0.131 & $+$ \\
Enthusiasm & 39.6 & 63.0 & 0.130 & $+$ \\
Safety conservatism & 25.2 & 62.9 & 0.129 & $+$ \\
Readability & 53.1 & 62.0 & 0.120 & $+$ \\
Analogy use & 16.9 & 61.6 & 0.116 & $+$ \\
Formatting & 37.7 & 61.2 & 0.112 & $+$ \\
Harmfulness & 12.6 & 39.3 & 0.107 & $-$ \\
Forcefulness & 51.0 & 60.3 & 0.103 & $+$ \\
Conversationality & 47.3 & 60.3 & 0.103 & $+$ \\
Difficulty & 48.3 & 60.2 & 0.102 & $+$ \\
Directness & 50.2 & 59.6 & 0.096 & $+$ \\
Verbosity & 65.4 & 58.7 & 0.087 & $+$ \\
Acknowledgment of limitations & 27.0 & 58.5 & 0.085 & $+$ \\
Moralizing tendency & 21.2 & 57.8 & 0.078 & $+$ \\
Caution framing & 37.8 & 57.5 & 0.075 & $+$ \\
Abstraction level & 49.8 & 56.6 & 0.066 & $+$ \\
Uncertainty expression & 32.2 & 55.7 & 0.057 & $+$ \\
Humor & 11.3 & 55.3 & 0.053 & $+$ \\
Hallucination rate & 30.7 & 44.9 & 0.051 & $-$ \\
Conciseness & 59.9 & 54.7 & 0.047 & $+$ \\
Hedging & 35.1 & 54.3 & 0.043 & $+$ \\
Self correction & 9.5 & 51.8 & 0.018 & $+$ \\
Sycophancy & 14.0 & 51.8 & 0.018 & $+$ \\
Refusal tendency & 19.0 & 48.9 & 0.011 & $-$ \\
Digression & 45.4 & 50.9 & 0.009 & $+$ \\
\bottomrule\end{tabular}\end{table}

\FloatBarrier

\clearpage
\subsection{Selected coefficients and per-judge rankings}
\label{app:attribute-preferences}
The own-perception fit predicts each judge's verdicts from its own stable
attribute directions. The shared-perception fit uses the cell-wise
majority direction of the 19 readers for all judges.
For cross-judge coefficient summaries, each vector is divided by its mean
absolute coefficient across all 87 attributes; normalization does not
change within-vector rankings. Correlated predictors, differences in
resolution, and regularization affect coefficients.

\begin{table}[t]
\centering\small\setlength{\tabcolsep}{6pt}
\caption{Selected coefficient summaries across 19 judges. Values are mean normalized coefficients with each judge's own readings or a common majority-based reading. Attributes are selected to illustrate both shared positive associations and penalties.\newline}

\label{tab:priority-consensus}
\begin{tabular}{lrr}
\toprule
Attribute & Own perception & Shared perception \\
\midrule
Final answer correctness & 4.29 & 3.26 \\
Strategy quality & 3.20 & 3.88 \\
Instruction adherence & 2.45 & 2.35 \\
Constraint satisfaction & 2.32 & 2.53 \\
Calibration & 1.92 & 3.26 \\
Insightfulness & 1.39 & 1.82 \\
Harmfulness & -5.71 & -2.54 \\
Hallucination rate & -2.24 & -1.49 \\
\bottomrule
\end{tabular}
\end{table}

\begin{table}[t]
\centering\footnotesize\setlength{\tabcolsep}{4pt}
\caption{Three largest positive coefficients for each judge in the shared-perception fit. Coefficients are raw, rounded to two decimal places; tied entries are retained in source order. Correctness abbreviates final-answer correctness, constraints abbreviates constraint satisfaction, and instruction abbreviates instruction adherence. All 87 attributes were eligible for this ranking. R1 and V4 denote DeepSeek models.\newline}

\label{tab:judge-top-priorities}
\begin{tabular}{lrrr}
\toprule
Judge & First & Second & Third \\
\midrule
R1-7B & Constraints (0.21) & Correctness (0.18) & step correctness (0.14) \\
R1-14B & Correctness (0.26) & Empathy (0.23) & Strategy (0.19) \\
R1-32B & Correctness (0.29) & Strategy (0.26) & Empathy (0.24) \\
V4-Flash & Strategy (0.36) & Constraints (0.20) & Empathy (0.20) \\
V4-Pro & Strategy (0.28) & Calibration (0.26) & robustness (0.23) \\
Gemma-3-27B & Constraints (0.31) & Strategy (0.29) & Empathy (0.29) \\
Gemma-4-26B & Strategy (0.39) & Instruction (0.36) & Calibration (0.30) \\
Gemma-4-31B & Instruction (0.52) & Strategy (0.41) & Correctness (0.33) \\
Kimi-K3 & Instruction (0.41) & Strategy (0.36) & Correctness (0.34) \\
Llama-4-Scout & Constraints (0.26) & Strategy (0.23) & Consistency (0.18) \\
Llama-4-Maverick & Constraints (0.31) & Strategy (0.28) & Correctness (0.26) \\
Mistral-24B & formatting (0.23) & Empathy (0.21) & Strategy (0.19) \\
Muse-Glimmer-30B & Calibration (0.41) & Strategy (0.39) & Factual accuracy (0.35) \\
Qwen3.5-4B & Correctness (0.23) & Calibration (0.23) & Consistency (0.20) \\
Qwen3.5-9B & Calibration (0.29) & Strategy (0.25) & Factual accuracy (0.22) \\
Qwen3.5-27B & Calibration (0.41) & Strategy (0.40) & Instruction (0.38) \\
Qwen3.5-122B & Calibration (0.39) & Strategy (0.32) & Instruction (0.31) \\
Qwen3.5-397B & Calibration (0.39) & Strategy (0.38) & Instruction (0.37) \\
Qwen3.6-27B & Calibration (0.45) & Instruction (0.37) & Strategy (0.35) \\
\bottomrule
\end{tabular}
\end{table}

\FloatBarrier
\clearpage
\section{Attribute-level Consensus and Importance}
\label{app:attribute-consensus-detail}
\subsection{Fitted attribute importance.}
To relate perception to what predicts each judge's choices, we fit a logistic model with 87 attribute coefficients to that judge's own overall verdicts (Section~\ref{sec:different-weights}). For attribute $a$, fitted importance is
\begin{equation}
I_a=\frac{1}{J}\sum_{j=1}^{J}\frac{|w_{ja}|}{87^{-1}\sum_{a'=1}^{87}|w_{ja'}|},\qquad J=21.
\label{eq:fitted-importance}
\end{equation}
This normalizes the coefficient scale within each judge before averaging magnitudes.

\begin{table}[t]
\centering\small
\caption{\textbf{Hidden consensus is uneven across attributes.} Opposite-winner agreement and joint stable coverage (\%) for all 21 judges. Quartiles are defined by fitted importance; each summary averages attributes within its group after averaging judge pairs equally. The selected attributes illustrate both lower-consensus qualities and high-consensus exceptions.}
\label{tab:importance-consensus}
\setlength{\tabcolsep}{7pt}
\begin{tabular}{@{}lrrr@{}}
\toprule
Attribute group & Attributes & Agreement & Coverage\\
\midrule
Q1 (lowest fitted importance) & 22 & 87.1 & 23.6\\
Q2 & 22 & 82.4 & 20.8\\
Q3 & 21 & 82.4 & 16.9\\
Q4 (highest fitted importance) & 22 & 76.8 & 14.7\\
\midrule
Final-answer correctness & 1 & 66.0 & 9.9\\
Strategy quality & 1 & 68.4 & 21.6\\
Factual accuracy & 1 & 64.2 & 14.8\\
Harmfulness & 1 & 91.5 & 5.0\\
Hallucination rate & 1 & 89.2 & 13.6\\
\bottomrule
\end{tabular}
\end{table}

All results in this appendix use 21 judges and 210 judge pairs under the order-consistent-direction protocol. These are the statistics used in Figure~\ref{fig:attribute-agreement-selected}, whereas the heatmaps and Tables~\ref{tab:hc-attributes-1}--\ref{tab:hc-attributes-2} use 19 judges. The figure selects nine attributes from the highest importance quartile to show variation within that group; they are not the nine highest-ranked attributes overall. Fitted importance is defined in Equation~\ref{eq:fitted-importance}; it uses each judge's own attribute representation and verdicts. Agreement conditions on joint stable resolution. Coverage reports how often both judges resolve an attribute in the corresponding winner stratum. Attribute summaries first average judge pairs equally; family summaries then average their constituent attributes.

Across the 87 attributes, fitted importance correlates negatively with opposite-winner agreement (Pearson $-0.253$, Spearman $-0.379$). For example, harmfulness and hallucination rate retain high consensus. Table~\ref{tab:consensus-families} reports all families; Tables~\ref{tab:consensus-attributes-1}--\ref{tab:consensus-attributes-3} report the full inventory. Agreement and coverage are proportions.
\begin{table}[t]
\centering\small
\caption{Attribute consensus by family under opposite overall winners. Importance is mean normalized coefficient magnitude.}
\label{tab:consensus-families}
\setlength{\tabcolsep}{4pt}\renewcommand{\arraystretch}{1.08}
\begin{tabular}{@{}lrrrr@{}}
\toprule
Family & Attributes & Importance & Agreement & Coverage\\
\midrule
correctness & 11 & 1.611 & 0.722 & 0.133\\
interpersonal & 1 & 0.419 & 0.748 & 0.225\\
communication & 8 & 0.882 & 0.777 & 0.204\\
relevance & 5 & 0.979 & 0.779 & 0.231\\
reasoning & 9 & 1.214 & 0.810 & 0.173\\
stance & 1 & 0.428 & 0.819 & 0.170\\
content & 1 & 0.494 & 0.823 & 0.365\\
social & 8 & 0.888 & 0.831 & 0.114\\
style & 10 & 0.703 & 0.842 & 0.272\\
epistemic & 7 & 0.888 & 0.844 & 0.182\\
format & 1 & 0.413 & 0.845 & 0.186\\
alignment & 6 & 1.894 & 0.861 & 0.069\\
information & 10 & 0.631 & 0.882 & 0.310\\
creative & 9 & 0.735 & 0.884 & 0.152\\
\bottomrule
\end{tabular}
\end{table}
\begin{table}[t]
\centering\small
\caption{All 87 attributes, ordered by fitted importance (part 1 of 3). Agreement: all items, same winner, opposite winners; coverage: opposite-winner items.}
\label{tab:consensus-attributes-1}
\setlength{\tabcolsep}{4pt}\renewcommand{\arraystretch}{1.08}
\begin{tabular}{@{}lrrrrr@{}}
\toprule
Attribute & Importance & All & Same & Opposite & Coverage\\
\midrule
harmfulness & 5.373 & 0.921 & 0.922 & 0.915 & 0.050\\
final answer correctness & 3.944 & 0.881 & 0.919 & 0.660 & 0.099\\
strategy quality & 2.958 & 0.873 & 0.910 & 0.684 & 0.216\\
constraint satisfaction & 2.490 & 0.873 & 0.905 & 0.705 & 0.029\\
hallucination rate & 2.304 & 0.890 & 0.888 & 0.892 & 0.136\\
instruction adherence & 2.281 & 0.883 & 0.909 & 0.746 & 0.194\\
calibration & 1.939 & 0.838 & 0.879 & 0.624 & 0.110\\
robustness & 1.741 & 0.870 & 0.898 & 0.738 & 0.184\\
sycophancy & 1.723 & 0.833 & 0.831 & 0.840 & 0.042\\
refusal tendency & 1.692 & 0.857 & 0.853 & 0.865 & 0.069\\
empathy & 1.619 & 0.916 & 0.929 & 0.875 & 0.045\\
clarity & 1.559 & 0.821 & 0.850 & 0.701 & 0.267\\
warmth & 1.408 & 0.897 & 0.908 & 0.860 & 0.135\\
step correctness & 1.400 & 0.877 & 0.908 & 0.700 & 0.072\\
insightfulness & 1.356 & 0.900 & 0.915 & 0.840 & 0.189\\
relevance & 1.318 & 0.904 & 0.928 & 0.792 & 0.188\\
completeness & 1.287 & 0.911 & 0.931 & 0.819 & 0.207\\
conciseness & 1.285 & 0.768 & 0.769 & 0.766 & 0.353\\
factual accuracy & 1.154 & 0.830 & 0.869 & 0.642 & 0.148\\
phrasing originality & 1.105 & 0.727 & 0.739 & 0.684 & 0.197\\
safety conservatism & 1.069 & 0.899 & 0.907 & 0.872 & 0.105\\
reasoning efficiency & 1.060 & 0.766 & 0.789 & 0.674 & 0.193\\
acknowledgment of limitations & 1.038 & 0.932 & 0.933 & 0.929 & 0.127\\
goal directedness & 1.036 & 0.867 & 0.893 & 0.755 & 0.229\\
assumption validity & 1.030 & 0.815 & 0.853 & 0.636 & 0.175\\
notation clarity & 1.027 & 0.926 & 0.935 & 0.895 & 0.056\\
informativeness & 1.002 & 0.918 & 0.931 & 0.874 & 0.385\\
self correction & 0.983 & 0.902 & 0.900 & 0.915 & 0.017\\
digression & 0.969 & 0.885 & 0.883 & 0.891 & 0.254\\
\bottomrule
\end{tabular}
\end{table}
\begin{table}[t]
\centering\small
\caption{All 87 attributes, ordered by fitted importance (part 2 of 3). Agreement: all items, same winner, opposite winners; coverage: opposite-winner items.}
\label{tab:consensus-attributes-2}
\setlength{\tabcolsep}{4pt}\renewcommand{\arraystretch}{1.08}
\begin{tabular}{@{}lrrrrr@{}}
\toprule
Attribute & Importance & All & Same & Opposite & Coverage\\
\midrule
internal consistency & 0.967 & 0.760 & 0.791 & 0.596 & 0.057\\
respectfulness & 0.949 & 0.865 & 0.893 & 0.722 & 0.104\\
persuasiveness & 0.943 & 0.903 & 0.918 & 0.845 & 0.213\\
interestingness & 0.905 & 0.901 & 0.909 & 0.875 & 0.323\\
neutrality & 0.834 & 0.843 & 0.863 & 0.748 & 0.045\\
confidence & 0.830 & 0.870 & 0.878 & 0.844 & 0.218\\
readability & 0.827 & 0.727 & 0.740 & 0.678 & 0.251\\
analogy use & 0.820 & 0.939 & 0.941 & 0.931 & 0.064\\
humor & 0.819 & 0.923 & 0.922 & 0.922 & 0.048\\
politeness & 0.819 & 0.881 & 0.890 & 0.853 & 0.120\\
groundedness & 0.817 & 0.881 & 0.901 & 0.800 & 0.111\\
directness & 0.813 & 0.812 & 0.818 & 0.795 & 0.258\\
verification & 0.799 & 0.904 & 0.911 & 0.879 & 0.056\\
verbosity & 0.795 & 0.932 & 0.935 & 0.925 & 0.444\\
reasoning completeness & 0.778 & 0.911 & 0.925 & 0.858 & 0.221\\
fluency & 0.775 & 0.830 & 0.856 & 0.713 & 0.119\\
focus & 0.761 & 0.758 & 0.780 & 0.662 & 0.224\\
logical validity & 0.756 & 0.859 & 0.896 & 0.677 & 0.144\\
vividness & 0.753 & 0.930 & 0.933 & 0.917 & 0.138\\
coherence & 0.753 & 0.884 & 0.909 & 0.780 & 0.260\\
storytelling quality & 0.747 & 0.926 & 0.933 & 0.901 & 0.113\\
terminological precision & 0.733 & 0.889 & 0.910 & 0.791 & 0.196\\
audience adaptation & 0.730 & 0.878 & 0.902 & 0.745 & 0.071\\
formality & 0.721 & 0.868 & 0.877 & 0.836 & 0.188\\
accessibility & 0.718 & 0.796 & 0.812 & 0.738 & 0.216\\
difficulty & 0.706 & 0.842 & 0.843 & 0.838 & 0.225\\
information density & 0.700 & 0.827 & 0.847 & 0.752 & 0.339\\
moralizing tendency & 0.674 & 0.926 & 0.927 & 0.924 & 0.101\\
organization & 0.666 & 0.940 & 0.945 & 0.923 & 0.269\\
\bottomrule
\end{tabular}
\end{table}
\begin{table}[t]
\centering\small
\caption{All 87 attributes, ordered by fitted importance (part 3 of 3). Agreement: all items, same winner, opposite winners; coverage: opposite-winner items.}
\label{tab:consensus-attributes-3}
\setlength{\tabcolsep}{4pt}\renewcommand{\arraystretch}{1.08}
\begin{tabular}{@{}lrrrrr@{}}
\toprule
Attribute & Importance & All & Same & Opposite & Coverage\\
\midrule
decomposition quality & 0.649 & 0.914 & 0.924 & 0.877 & 0.216\\
detail level & 0.648 & 0.939 & 0.944 & 0.925 & 0.441\\
qualification & 0.644 & 0.909 & 0.919 & 0.876 & 0.249\\
assertiveness & 0.642 & 0.864 & 0.869 & 0.850 & 0.238\\
conversationality & 0.623 & 0.854 & 0.857 & 0.847 & 0.242\\
novelty & 0.603 & 0.879 & 0.883 & 0.867 & 0.097\\
rigor & 0.602 & 0.899 & 0.918 & 0.823 & 0.269\\
friendliness & 0.600 & 0.865 & 0.874 & 0.838 & 0.176\\
specificity & 0.585 & 0.919 & 0.926 & 0.897 & 0.405\\
justification quality & 0.578 & 0.921 & 0.932 & 0.884 & 0.298\\
creativity & 0.576 & 0.916 & 0.921 & 0.900 & 0.170\\
hedging & 0.566 & 0.893 & 0.894 & 0.894 & 0.180\\
uncertainty expression & 0.557 & 0.889 & 0.890 & 0.889 & 0.150\\
evidence use & 0.538 & 0.939 & 0.942 & 0.930 & 0.098\\
personalization & 0.534 & 0.901 & 0.915 & 0.850 & 0.071\\
conceptual complexity & 0.511 & 0.930 & 0.933 & 0.919 & 0.269\\
technicality & 0.508 & 0.943 & 0.947 & 0.933 & 0.231\\
elaboration & 0.494 & 0.841 & 0.847 & 0.823 & 0.365\\
explanatory depth & 0.491 & 0.927 & 0.935 & 0.898 & 0.304\\
linguistic complexity & 0.461 & 0.922 & 0.925 & 0.913 & 0.325\\
enthusiasm & 0.449 & 0.909 & 0.913 & 0.900 & 0.190\\
originality & 0.448 & 0.833 & 0.842 & 0.801 & 0.203\\
caution framing & 0.428 & 0.822 & 0.823 & 0.819 & 0.170\\
forcefulness & 0.419 & 0.760 & 0.764 & 0.748 & 0.225\\
formatting & 0.413 & 0.853 & 0.855 & 0.845 & 0.186\\
example use & 0.407 & 0.933 & 0.936 & 0.921 & 0.262\\
breadth & 0.383 & 0.921 & 0.925 & 0.908 & 0.323\\
abstraction level & 0.311 & 0.749 & 0.748 & 0.757 & 0.247\\
depth & 0.271 & 0.919 & 0.926 & 0.896 & 0.337\\
\bottomrule
\end{tabular}
\end{table}
\clearpage
\section{Overall Disagreement Without Attribute Conflict}
\label{app:no-conflict-detail}
\begin{table}[t]
\centering\small
\caption{\textbf{Measured attribute consensus leaves overall disagreement.} All 87 attributes, at least 20 jointly resolved directions, and definite overall verdicts. Percentages report different winners, not different attribute directions. Counts are mean items per judge pair. The 19-judge sensitivity estimates use rounded per-pair summaries.}
\label{tab:no-conflict-main}
\setlength{\tabcolsep}{6pt}
\begin{tabular}{@{}llrrr@{}}
\toprule
Judges & Condition & Items/pair & Pair mean & Observation-weighted\\
\midrule
21 & No conflict & 5,245 & 14.5 & 11.5\\
21 & At least one conflict & 17,329 & 21.8 & 19.7\\
19 & No conflict & --- & $\approx12.0$ & $\approx11.1$\\
\bottomrule
\end{tabular}
\end{table}

The main comparison uses all 21 judges and 210 judge pairs. It gives the 14.5\% no-conflict and 21.8\% conflict rates shown in Figure~\ref{fig:consensus-limits}; the 19-judge results below are sensitivity checks. For each judge pair, we retain items with definite overall verdicts and at least $k$ jointly resolved stable attribute directions. No conflict means all those directions agree, within the specified attribute inventory. It does not require all 87 attributes to be resolved or the complete perceptions to match. Pair means weight eligible judge pairs equally; observation-weighted means pool qualifying pair--item observations. Repeated appearances of an item across judge pairs are separate observations, so these rates are not proportions of unique response pairs.

Table~\ref{tab:no-conflict-thresholds} varies $k$ for the full inventory. Increasing $k$ changes the selected population as well as the required coverage. Table~\ref{tab:no-conflict-top10} uses only the ten attributes with largest fitted importance. No conflict on this smaller set can coexist with conflict elsewhere.
\begin{table}[t]
\centering\small
\caption{Full 87-attribute inventory: coverage thresholds and remaining winner disagreement. Counts are mean items per judge pair; disagreement is in percent.}
\label{tab:no-conflict-thresholds}
\setlength{\tabcolsep}{4pt}\renewcommand{\arraystretch}{1.08}
\begin{tabular}{@{}rrrrrr@{}}
\toprule
$k$ & Eligible & No conflict & Conflict & Pair mean & Obs.-weighted\\
\midrule
5 & 36,403 & 10,948 & 25,454 & 20.8 & 18.2\\
10 & 30,953 & 8,309 & 22,644 & 18.4 & 15.5\\
20 & 22,575 & 5,245 & 17,329 & 14.5 & 11.5\\
30 & 15,305 & 3,204 & 12,101 & 10.9 & 8.4\\
40 & 8,728 & 1,748 & 6,979 & 7.5 & 5.7\\
\bottomrule
\end{tabular}
\end{table}
\begin{table}[t]
\centering\small
\caption{Top ten attributes by fitted importance: coverage thresholds and remaining winner disagreement. Counts are mean items per judge pair; disagreement is in percent.}
\label{tab:no-conflict-top10}
\setlength{\tabcolsep}{4pt}\renewcommand{\arraystretch}{1.08}
\begin{tabular}{@{}rrrrrr@{}}
\toprule
$k$ & Eligible & No conflict & Conflict & Pair mean & Obs.-weighted\\
\midrule
1 & 28,814 & 22,684 & 6,131 & 18.4 & 16.0\\
3 & 13,936 & 10,732 & 3,204 & 11.6 & 8.1\\
5 & 5,402 & 4,116 & 1,287 & 8.3 & 4.5\\
7 & 873 & 581 & 292 & 6.1 & 3.3\\
\bottomrule
\end{tabular}
\end{table}

At $k=20$, excluding Qwen3.5-0.8B leaves 190 pairs with approximately 12.2\% pair-mean and 11.1\% observation-weighted disagreement. Excluding both Qwen3.5-0.8B and Qwen3.5-2B leaves 171 pairs with approximately 12.0\% and 11.1\%, respectively.

Tables~\ref{tab:no-conflict-judges} and \ref{tab:no-conflict-sources} show that the residual varies by judge and source. The source-level means range from 8.0\% on RewardBench to 27.8\% on IF-RewardBench. Unresolved attributes, difference magnitudes, unmeasured qualities, and noise may also contribute to this residual.
\begin{table}[t]
\centering\small
\caption{No-conflict results by judge, full inventory and $k=20$. Each row averages that judge's 20 partner pairs. Disagreement is in percent.}
\label{tab:no-conflict-judges}
\setlength{\tabcolsep}{4pt}\renewcommand{\arraystretch}{1.08}
\begin{tabular}{@{}lrr@{}}
\toprule
Judge & No-conflict items/pair & Winner disagreement\\
\midrule
Qwen3.5-0.8B & 762 & 35.8\\
DeepSeek-R1-Distill-14B & 3,794 & 15.7\\
DeepSeek-R1-Distill-32B & 3,853 & 15.2\\
DeepSeek-R1-Distill-7B & 1,965 & 15.1\\
Qwen3.5-2B & 1,004 & 14.9\\
Muse-Glimmer-30B & 2,257 & 14.8\\
Llama-4-Scout & 6,350 & 14.3\\
Qwen3.5-122B-A10B & 7,218 & 13.9\\
Qwen3.5-9B & 6,018 & 13.9\\
Mistral-Small-24B & 4,476 & 13.6\\
Gemma-3-27B & 8,062 & 13.4\\
Gemma-4-26B-A4B & 6,299 & 13.4\\
Kimi-K3 & 4,896 & 13.0\\
Llama-4-Maverick & 6,114 & 12.8\\
Qwen3.6-27B & 7,447 & 12.7\\
Qwen3.5-397B-A17B & 6,713 & 12.5\\
DeepSeek-V4-Flash & 7,533 & 12.4\\
Gemma-4-31B & 5,213 & 12.1\\
Qwen3.5-27B & 8,081 & 11.8\\
Qwen3.5-4B & 6,535 & 11.6\\
DeepSeek-V4-Pro & 5,567 & 11.3\\
\bottomrule
\end{tabular}
\end{table}
\begin{table}[t]
\centering\small
\caption{No-conflict results by source, full inventory and $k=20$. Each row averages 210 judge pairs. Disagreement is in percent.}
\label{tab:no-conflict-sources}
\setlength{\tabcolsep}{4pt}\renewcommand{\arraystretch}{1.08}
\begin{tabular}{@{}lrr@{}}
\toprule
Source & No-conflict items/pair & Winner disagreement\\
\midrule
IF-RewardBench & 44 & 27.8\\
hh-rlhf & 428 & 19.0\\
arena-140k & 459 & 18.7\\
webgpt & 442 & 17.6\\
RM-Bench & 100 & 17.1\\
PPE-Human & 443 & 16.6\\
arena-expert & 217 & 15.6\\
mt\_bench & 438 & 14.4\\
HelpSteer2 & 418 & 14.3\\
SHP & 491 & 13.3\\
reward-bench-2 & 181 & 13.1\\
summarize & 196 & 12.8\\
PPE-BoK & 193 & 12.5\\
Nectar & 289 & 12.1\\
UltraFeedback & 308 & 12.1\\
JudgeBench & 44 & 9.8\\
reward-bench & 554 & 8.0\\
\bottomrule
\end{tabular}
\end{table}
\clearpage
\providecommand{\ReweightNote}[1]{\par\smallskip\noindent\textbf{EDITORIAL:} #1\par\smallskip}
\section{Reweighting and Adaptation}
\label{app:reweighting-full}
\label{app:alignment}

This appendix supports the three experiments in Section~\ref{sec:fingerprint}. It reports source-level results, the restricted RM-Bench sampling, and the protocols for the matched fine-tuning and rubric comparisons.

\subsection{Evaluation, aggregation, and label sources}
\label{app:rw-evaluation}
The reported evaluation covers 21 judges and 17 datasets, yielding 357
judge--dataset combinations. Reweighting uses dataset-provided training
labels to fit the combination of each judge's fixed attribute comparisons.
The target labels may reflect human preferences, model judgments, or
verifiable answers, depending on the dataset. Full-data reweighting denotes using the
full available training split for the corresponding fit.

Let $A^{\mathrm{base}}_{js}$ and $A^{\mathrm{rw}}_{js}$ denote accuracy in
percent for judge $j$ on dataset $s$. The gain, in percentage points, is
\begin{equation}
\Delta_{js}=A^{\mathrm{rw}}_{js}-A^{\mathrm{base}}_{js}.
\end{equation}
Judge-level means weight datasets equally; dataset-level means weight
judges equally. The overall macro gain is
\begin{equation}
\overline{\Delta}=\frac{1}{21\cdot17}\sum_{j=1}^{21}\sum_{s=1}^{17}\Delta_{js}.
\end{equation}
The preference-only aggregate restricts the sum to the 11 sources listed
below and replaces 17 by 11. The supplementary exclusion of RM-Bench
removes that column and replaces 17 by 16. This
aggregation does not weight datasets by test-set size. The cell summaries are analyzed separately from the controlled SFT and rubric comparisons below.


\subsection{Precision.}
In the detailed matrices, accuracies and gains are rounded to one decimal; aggregates may differ by 0.1 points.

\subsection{Grouping by label source.}
The \emph{preference-labeled} group contains
11 sources: Arena-Expert, UltraFeedback, Nectar, HelpSteer2, WebGPT,
MT-Bench Human, SHP, HH-RLHF, Arena Human, Summarization, and PPE Human.
UltraFeedback and Nectar carry AI-generated targets; the remaining nine
sources carry human-derived targets. The \emph{benchmark-labeled} group
contains six sources: JudgeBench, IF-RewardBench, RM-Bench, RewardBench 2,
PPE Best-of-K, and RewardBench.

This source-level partition is operational, not a pure separation between
subjective and objective judgments. JudgeBench and PPE Best-of-K include
correctness-based comparisons, IF-RewardBench targets instruction
following, and RewardBench contains mixed evaluation domains.
Preference labels may also depend on correctness. A strict objective-only
analysis of mixed benchmarks would require item-level or subtask-level
annotations beyond the dataset-level summaries analyzed here.

\footnote{Primary dataset descriptions:
\url{https://github.com/lmarena/PPE},
\url{https://github.com/ScalerLab/JudgeBench},
\url{https://github.com/thu-coai/IF-RewardBench}, and
\url{https://github.com/allenai/reward-bench}.}

\subsection{Full results and source-specific limitations}
\label{app:rw-results}
\subsection{Breadth of the improvement.}
Across the 11 preference-labeled datasets, mean accuracy increases from
approximately 63.50\% to 68.48\%, with a mean gain of 4.98 points.
All 21 judges improve on average within this group; model-level gains
range from 1.59 to 9.25 points. All 11 source means are positive, although
the HelpSteer2 gain is only 0.11 points and is negligible.
Across all 17 datasets, accuracy increases from 66.48\% to 71.97\%, with
a mean gain of 5.49 points. All judges also improve under this aggregation.
The six benchmark sources average a 6.41-point gain, but this is strongly
influenced by the RM-Bench comparison setting discussed below. Excluding
RM-Bench from that group leaves a mean gain of 1.82 points over five
sources. Across the remaining 16 datasets of both types, the gain is
3.99 points.

\begin{table}[t]
\centering\small
\caption{Judge-level macro results. Accuracies are percentages; gains are
percentage points. Base, reweighted, and gain use all 17 datasets; preference
gain uses only the 11 preference-labeled datasets. The final column uses
all 16 datasets other than RM-Bench. Each included dataset receives equal
weight. Full model names follow the evaluation model inventory.\newline}

\label{tab:rw-judge-summary}
\begin{tabular}{lrrrrr}
\toprule
Judge & Base & Reweighted & Gain & Preference gain & Gain excl. RM \\
\midrule
Kimi-K3 & 74.45 & 76.96 & +2.51 & +4.06 & +1.66 \\
Gemma-4-31B & 73.25 & 77.18 & +3.93 & +4.83 & +3.04 \\
Qwen3.6-27B & 73.08 & 76.64 & +3.55 & +4.05 & +2.58 \\
Qwen3.5-27B & 72.16 & 75.18 & +3.01 & +4.15 & +1.86 \\
DeepSeek-V4-Pro & 69.12 & 73.05 & +3.92 & +2.35 & +2.11 \\
Gemma-4-26B-A4B & 70.46 & 75.39 & +4.92 & +5.88 & +3.90 \\
Muse-Glimmer-30B & 70.65 & 77.08 & +6.43 & +6.58 & +5.87 \\
Qwen3.5-397B-A17B & 70.71 & 76.17 & +5.45 & +7.07 & +4.59 \\
Qwen3.5-122B-A10B & 70.67 & 74.78 & +4.11 & +6.76 & +3.18 \\
Llama-4-Maverick & 66.70 & 73.44 & +6.73 & +4.66 & +4.96 \\
DeepSeek-V4-Flash & 66.68 & 72.16 & +5.50 & +4.40 & +3.83 \\
Gemma-3-27B & 65.84 & 71.95 & +6.11 & +2.29 & +3.30 \\
Llama-4-Scout & 65.49 & 72.28 & +6.80 & +3.56 & +4.36 \\
Qwen3.5-9B & 67.86 & 71.51 & +3.65 & +6.34 & +2.90 \\
DeepSeek-R1-Distill-32B & 64.20 & 73.22 & +9.02 & +6.65 & +7.83 \\
Mistral-Small-24B & 63.03 & 68.51 & +5.46 & +2.36 & +1.79 \\
DeepSeek-R1-Distill-14B & 63.32 & 71.42 & +8.09 & +5.28 & +6.66 \\
Qwen3.5-4B & 65.30 & 70.25 & +4.94 & +6.77 & +3.22 \\
Qwen3.5-2B & 58.41 & 60.61 & +2.21 & +1.59 & +0.86 \\
DeepSeek-R1-Distill-7B & 57.19 & 66.48 & +9.27 & +5.71 & +7.40 \\
Qwen3.5-0.8B & 47.55 & 57.15 & +9.62 & +9.25 & +7.99 \\
\midrule
Macro mean & 66.48 & 71.97 & +5.49 & +4.98 & +3.99 \\
\bottomrule
\end{tabular}
\end{table}

\subsection{Dataset heterogeneity and limitations.}
Mean gains are positive on 15 of 17 datasets. RM-Bench has the largest
gain (+29.37 points), followed by MT-Bench Human (+16.50), Nectar (+10.62),
SHP (+8.79), and PPE Best-of-K (+7.84).
HelpSteer2 is nearly unchanged (+0.11); RewardBench (-0.18) and JudgeBench
(-2.35) have negative mean gains. JudgeBench is especially heterogeneous:
individual gains range from -16.2 to +16.2 points.

Outside the RM-Bench $0{:}2$ setting, gains are largest on preference-labeled sources, though PPE Best-of-K shows benchmark labels can also benefit.

\subsection{RM-Bench comparison setting.} \label{app:rm-sampling} RM-Bench provides chosen and rejected responses at three style levels, giving a $3\times3$ set of comparisons for each prompt. Our collection uses the $0{:}2$ pairing: a plain chosen response and a more elaborate, formatted rejected response. This is one hard style pairing rather than the full benchmark evaluation.\footnote{Official benchmark description: \url{https://github.com/THU-KEG/RM-Bench}.} \par The responses in this pairing differ clearly in style, and the reference label consistently favors the plain chosen response. The reported longer-response win rate is 0.038. Attribute judgments can therefore carry strong information about the reference label through differences in length and formatting. This provides a plausible explanation for the large mean gain of 29.37 percentage points and near-perfect agreement for several judges. Mean accuracy across the 21 judges rises from 65.05\% to 94.44\%. \par These results show adaptation to this particular comparison setting. Performance on the remaining style pairings requires separate evaluation. Excluding RM-Bench, the mean gain remains 3.99 percentage points across the other 16 datasets. \begin{table}[t]
\centering\small
\caption{Dataset-level macro results across 21 judges. Gain range describes
between-judge variation, not estimation uncertainty.\newline}

\label{tab:rw-dataset-summary}
\begin{tabular}{lrrrr}
\toprule
Dataset & Base & Reweighted & Gain & Min / max gain \\
\midrule
RM-Bench & 65.05 & 94.44 & +29.37 & +15.4 / +64.3 \\
MT-Human & 62.06 & 78.54 & +16.50 & +3.1 / +29.4 \\
Nectar & 64.47 & 75.09 & +10.62 & +2.1 / +16.6 \\
SHP & 58.06 & 66.85 & +8.79 & +2.1 / +13.5 \\
PPE-BoK & 66.64 & 74.49 & +7.84 & +0.2 / +19.5 \\
Summarization & 60.33 & 66.61 & +6.29 & +0.6 / +27.4 \\
Arena-Human & 60.43 & 63.84 & +3.41 & -0.2 / +9.1 \\
HH-RLHF & 60.24 & 63.47 & +3.23 & -1.0 / +8.6 \\
PPE-Human & 63.78 & 65.80 & +2.02 & -0.8 / +7.1 \\
RB2 & 84.59 & 86.53 & +1.94 & -2.9 / +10.4 \\
IF-RB & 54.73 & 56.60 & +1.87 & -11.4 / +12.4 \\
UltraFeedback & 73.72 & 75.47 & +1.73 & -2.7 / +7.2 \\
WebGPT & 64.72 & 66.19 & +1.48 & -2.2 / +7.4 \\
Arena-Expert & 59.55 & 60.18 & +0.62 & -4.5 / +11.4 \\
HelpSteer2 & 71.11 & 71.22 & +0.11 & -4.3 / +7.5 \\
RewardBench & 87.49 & 87.32 & -0.18 & -7.3 / +6.0 \\
JudgeBench & 73.22 & 70.86 & -2.35 & -16.2 / +16.2 \\
\bottomrule
\end{tabular}
\end{table}

\subsection{Complete judge--dataset results.}
The following three panels report base and reweighted accuracy for every
cell, ordering datasets by ascending test-set size.
Each entry is base $\rightarrow$ reweighted, in percent.
\begin{table}[t]
\centering\small\setlength{\tabcolsep}{3pt}
\caption{Complete accuracy results, panel 1 of 3.\newline}

\label{tab:rw-full-1}
\resizebox{\linewidth}{!}{%
\begin{tabular}{lrrrrrr}
\toprule
Judge & JudgeBench & IF-RB & RM-Bench & RB2 & Arena-Expert & UltraFeedback \\
\midrule
Kimi-K3 & $93.5\!\rightarrow\!80.5$ & $73.8\!\rightarrow\!65.8$ & $83.6\!\rightarrow\!99.7$ & $94.4\!\rightarrow\!93.7$ & $59.0\!\rightarrow\!62.4$ & $78.6\!\rightarrow\!80.0$ \\
Gemma-4-31B & $88.3\!\rightarrow\!77.3$ & $69.3\!\rightarrow\!69.3$ & $81.5\!\rightarrow\!99.7$ & $94.0\!\rightarrow\!94.7$ & $66.4\!\rightarrow\!63.5$ & $79.0\!\rightarrow\!80.8$ \\
Qwen3.6-27B & $90.9\!\rightarrow\!87.0$ & $65.3\!\rightarrow\!62.4$ & $80.1\!\rightarrow\!99.3$ & $93.5\!\rightarrow\!91.5$ & $63.3\!\rightarrow\!61.5$ & $79.8\!\rightarrow\!81.0$ \\
Qwen3.5-27B & $88.3\!\rightarrow\!79.2$ & $66.8\!\rightarrow\!57.4$ & $78.3\!\rightarrow\!99.7$ & $92.0\!\rightarrow\!91.5$ & $62.6\!\rightarrow\!61.9$ & $79.5\!\rightarrow\!80.2$ \\
DeepSeek-V4-Pro & $71.4\!\rightarrow\!66.9$ & $53.0\!\rightarrow\!55.9$ & $62.6\!\rightarrow\!95.5$ & $89.4\!\rightarrow\!90.6$ & $61.9\!\rightarrow\!62.4$ & $77.0\!\rightarrow\!77.8$ \\
Gemma-4-26B-A4B & $89.6\!\rightarrow\!75.3$ & $57.4\!\rightarrow\!60.9$ & $78.3\!\rightarrow\!99.7$ & $88.9\!\rightarrow\!93.5$ & $62.4\!\rightarrow\!62.1$ & $76.8\!\rightarrow\!78.2$ \\
Muse-Glimmer-30B & $81.2\!\rightarrow\!84.4$ & $64.4\!\rightarrow\!71.8$ & $83.6\!\rightarrow\!99.0$ & $93.0\!\rightarrow\!91.8$ & $61.2\!\rightarrow\!61.9$ & $78.2\!\rightarrow\!79.2$ \\
Qwen3.5-397B-A17B & $85.7\!\rightarrow\!76.6$ & $65.3\!\rightarrow\!66.8$ & $79.7\!\rightarrow\!99.0$ & $94.0\!\rightarrow\!94.0$ & $63.7\!\rightarrow\!61.7$ & $77.9\!\rightarrow\!80.8$ \\
Qwen3.5-122B-A10B & $88.3\!\rightarrow\!75.3$ & $61.9\!\rightarrow\!50.5$ & $80.4\!\rightarrow\!99.3$ & $93.7\!\rightarrow\!95.2$ & $63.5\!\rightarrow\!60.8$ & $78.9\!\rightarrow\!80.2$ \\
Llama-4-Maverick & $66.9\!\rightarrow\!72.1$ & $48.5\!\rightarrow\!55.0$ & $61.5\!\rightarrow\!96.5$ & $87.7\!\rightarrow\!89.4$ & $59.0\!\rightarrow\!62.1$ & $75.8\!\rightarrow\!78.2$ \\
DeepSeek-V4-Flash & $69.5\!\rightarrow\!66.2$ & $53.0\!\rightarrow\!56.4$ & $59.4\!\rightarrow\!91.6$ & $85.7\!\rightarrow\!90.8$ & $62.6\!\rightarrow\!62.1$ & $73.9\!\rightarrow\!77.3$ \\
Gemma-3-27B & $66.9\!\rightarrow\!63.0$ & $49.0\!\rightarrow\!56.9$ & $46.9\!\rightarrow\!97.9$ & $81.4\!\rightarrow\!87.2$ & $63.0\!\rightarrow\!58.6$ & $73.9\!\rightarrow\!75.0$ \\
Llama-4-Scout & $59.7\!\rightarrow\!67.5$ & $47.5\!\rightarrow\!52.0$ & $50.0\!\rightarrow\!95.8$ & $85.0\!\rightarrow\!90.1$ & $59.5\!\rightarrow\!60.1$ & $76.2\!\rightarrow\!77.1$ \\
Qwen3.5-9B & $80.5\!\rightarrow\!64.3$ & $54.0\!\rightarrow\!51.0$ & $76.6\!\rightarrow\!92.3$ & $88.4\!\rightarrow\!88.4$ & $62.1\!\rightarrow\!59.5$ & $76.5\!\rightarrow\!77.9$ \\
DeepSeek-R1-Distill-32B & $60.4\!\rightarrow\!75.3$ & $40.1\!\rightarrow\!51.0$ & $67.8\!\rightarrow\!95.8$ & $82.6\!\rightarrow\!87.9$ & $61.0\!\rightarrow\!60.6$ & $69.4\!\rightarrow\!73.8$ \\
Mistral-Small-24B & $61.7\!\rightarrow\!63.6$ & $46.0\!\rightarrow\!45.5$ & $29.7\!\rightarrow\!94.1$ & $82.1\!\rightarrow\!79.7$ & $55.7\!\rightarrow\!60.6$ & $74.2\!\rightarrow\!71.5$ \\
DeepSeek-R1-Distill-14B & $58.4\!\rightarrow\!70.1$ & $39.6\!\rightarrow\!52.0$ & $64.3\!\rightarrow\!95.5$ & $80.4\!\rightarrow\!86.0$ & $57.9\!\rightarrow\!59.9$ & $69.6\!\rightarrow\!72.0$ \\
Qwen3.5-4B & $81.2\!\rightarrow\!66.2$ & $57.9\!\rightarrow\!54.5$ & $64.7\!\rightarrow\!97.2$ & $89.6\!\rightarrow\!87.7$ & $59.2\!\rightarrow\!57.2$ & $73.4\!\rightarrow\!76.5$ \\
Qwen3.5-2B & $56.5\!\rightarrow\!59.7$ & $48.5\!\rightarrow\!49.5$ & $49.0\!\rightarrow\!72.7$ & $67.1\!\rightarrow\!64.3$ & $50.6\!\rightarrow\!55.0$ & $65.6\!\rightarrow\!63.4$ \\
DeepSeek-R1-Distill-7B & $50.6\!\rightarrow\!66.9$ & $40.6\!\rightarrow\!52.0$ & $52.8\!\rightarrow\!92.0$ & $65.2\!\rightarrow\!70.5$ & $47.2\!\rightarrow\!58.6$ & $63.8\!\rightarrow\!66.6$ \\
Qwen3.5-0.8B & $48.1\!\rightarrow\!50.6$ & $47.5\!\rightarrow\!52.0$ & $35.3\!\rightarrow\!71.0$ & $48.3\!\rightarrow\!58.7$ & $48.8\!\rightarrow\!51.2$ & $50.2\!\rightarrow\!57.4$ \\
\bottomrule
\end{tabular}
}
\end{table}

\noindent\begin{tabular}{ll}
JudgeBench & \nolinkurl{JudgeBench}\\
IF-RB & \nolinkurl{IF-RewardBench}\\
RM-Bench & \nolinkurl{RM-Bench}\\
RB2 & \nolinkurl{reward-bench-2}\\
Arena-Expert & \nolinkurl{arena-expert-5k}\\
UltraFeedback & \nolinkurl{UltraFeedback}
\end{tabular}

\begin{table}[t]
\centering\small\setlength{\tabcolsep}{3pt}
\caption{Complete accuracy results, panel 2 of 3.\newline}

\label{tab:rw-full-2}
\resizebox{\linewidth}{!}{%
\begin{tabular}{lrrrrrr}
\toprule
Judge & Nectar & PPE-BoK & HelpSteer2 & WebGPT & MT-Human & SHP \\
\midrule
Kimi-K3 & $60.5\!\rightarrow\!75.2$ & $80.9\!\rightarrow\!84.4$ & $76.0\!\rightarrow\!77.8$ & $69.0\!\rightarrow\!69.3$ & $72.8\!\rightarrow\!81.2$ & $59.6\!\rightarrow\!68.9$ \\
Gemma-4-31B & $60.0\!\rightarrow\!76.6$ & $79.6\!\rightarrow\!84.1$ & $76.4\!\rightarrow\!78.2$ & $69.3\!\rightarrow\!70.0$ & $63.0\!\rightarrow\!80.3$ & $58.2\!\rightarrow\!69.8$ \\
Qwen3.6-27B & $62.6\!\rightarrow\!73.3$ & $78.6\!\rightarrow\!84.8$ & $76.2\!\rightarrow\!76.9$ & $69.7\!\rightarrow\!70.8$ & $66.8\!\rightarrow\!81.4$ & $61.4\!\rightarrow\!68.8$ \\
Qwen3.5-27B & $61.9\!\rightarrow\!75.8$ & $74.3\!\rightarrow\!79.3$ & $76.4\!\rightarrow\!74.6$ & $70.4\!\rightarrow\!69.2$ & $62.6\!\rightarrow\!81.4$ & $58.9\!\rightarrow\!67.6$ \\
DeepSeek-V4-Pro & $69.0\!\rightarrow\!74.6$ & $63.0\!\rightarrow\!71.8$ & $74.6\!\rightarrow\!72.4$ & $69.7\!\rightarrow\!69.0$ & $70.2\!\rightarrow\!79.9$ & $60.2\!\rightarrow\!68.1$ \\
Gemma-4-26B-A4B & $64.6\!\rightarrow\!74.4$ & $75.9\!\rightarrow\!79.6$ & $74.2\!\rightarrow\!76.4$ & $67.9\!\rightarrow\!67.5$ & $55.4\!\rightarrow\!80.1$ & $57.7\!\rightarrow\!68.2$ \\
Muse-Glimmer-30B & $61.6\!\rightarrow\!77.1$ & $76.4\!\rightarrow\!87.1$ & $72.4\!\rightarrow\!76.9$ & $68.0\!\rightarrow\!68.7$ & $56.7\!\rightarrow\!81.2$ & $56.3\!\rightarrow\!69.4$ \\
Qwen3.5-397B-A17B & $62.7\!\rightarrow\!76.3$ & $75.8\!\rightarrow\!79.9$ & $73.1\!\rightarrow\!75.6$ & $64.9\!\rightarrow\!68.8$ & $51.3\!\rightarrow\!80.2$ & $58.8\!\rightarrow\!70.8$ \\
Qwen3.5-122B-A10B & $61.9\!\rightarrow\!77.6$ & $76.7\!\rightarrow\!77.8$ & $74.7\!\rightarrow\!73.6$ & $64.6\!\rightarrow\!68.8$ & $50.6\!\rightarrow\!79.9$ & $58.3\!\rightarrow\!69.4$ \\
Llama-4-Maverick & $65.8\!\rightarrow\!77.6$ & $62.8\!\rightarrow\!77.0$ & $72.8\!\rightarrow\!72.7$ & $65.1\!\rightarrow\!66.0$ & $69.5\!\rightarrow\!80.3$ & $59.0\!\rightarrow\!66.4$ \\
DeepSeek-V4-Flash & $66.9\!\rightarrow\!78.1$ & $62.2\!\rightarrow\!68.9$ & $71.7\!\rightarrow\!71.4$ & $66.8\!\rightarrow\!65.5$ & $60.8\!\rightarrow\!77.9$ & $60.6\!\rightarrow\!67.3$ \\
Gemma-3-27B & $71.8\!\rightarrow\!76.8$ & $56.8\!\rightarrow\!71.8$ & $71.4\!\rightarrow\!72.0$ & $65.2\!\rightarrow\!67.2$ & $73.6\!\rightarrow\!78.3$ & $61.1\!\rightarrow\!68.1$ \\
Llama-4-Scout & $71.4\!\rightarrow\!77.3$ & $62.8\!\rightarrow\!75.3$ & $72.9\!\rightarrow\!72.2$ & $64.6\!\rightarrow\!65.7$ & $71.6\!\rightarrow\!79.4$ & $61.5\!\rightarrow\!66.0$ \\
Qwen3.5-9B & $62.4\!\rightarrow\!77.0$ & $73.7\!\rightarrow\!73.8$ & $72.8\!\rightarrow\!72.9$ & $63.9\!\rightarrow\!65.2$ & $55.7\!\rightarrow\!77.5$ & $58.6\!\rightarrow\!66.6$ \\
DeepSeek-R1-Distill-32B & $66.1\!\rightarrow\!78.4$ & $60.0\!\rightarrow\!79.4$ & $71.4\!\rightarrow\!72.2$ & $63.8\!\rightarrow\!65.9$ & $57.1\!\rightarrow\!80.7$ & $57.1\!\rightarrow\!70.6$ \\
Mistral-Small-24B & $73.9\!\rightarrow\!76.0$ & $54.5\!\rightarrow\!61.9$ & $71.5\!\rightarrow\!68.9$ & $62.7\!\rightarrow\!63.9$ & $69.1\!\rightarrow\!79.3$ & $61.1\!\rightarrow\!63.4$ \\
DeepSeek-R1-Distill-14B & $66.4\!\rightarrow\!74.9$ & $60.9\!\rightarrow\!76.7$ & $69.1\!\rightarrow\!68.1$ & $61.1\!\rightarrow\!66.5$ & $65.1\!\rightarrow\!79.9$ & $57.4\!\rightarrow\!68.5$ \\
Qwen3.5-4B & $65.0\!\rightarrow\!74.9$ & $67.0\!\rightarrow\!71.6$ & $72.4\!\rightarrow\!69.1$ & $62.8\!\rightarrow\!63.9$ & $54.8\!\rightarrow\!76.0$ & $54.1\!\rightarrow\!66.0$ \\
Qwen3.5-2B & $66.9\!\rightarrow\!73.0$ & $53.4\!\rightarrow\!54.9$ & $62.0\!\rightarrow\!57.7$ & $61.0\!\rightarrow\!58.7$ & $65.0\!\rightarrow\!68.1$ & $55.8\!\rightarrow\!57.9$ \\
DeepSeek-R1-Distill-7B & $62.6\!\rightarrow\!74.1$ & $54.1\!\rightarrow\!70.8$ & $63.5\!\rightarrow\!60.8$ & $57.0\!\rightarrow\!60.4$ & $64.9\!\rightarrow\!76.8$ & $56.8\!\rightarrow\!63.0$ \\
Qwen3.5-0.8B & $49.8\!\rightarrow\!57.8$ & $50.1\!\rightarrow\!53.4$ & $47.8\!\rightarrow\!55.2$ & $51.6\!\rightarrow\!59.0$ & $46.6\!\rightarrow\!69.6$ & $46.8\!\rightarrow\!59.1$ \\
\bottomrule
\end{tabular}
}
\end{table}

\noindent\begin{tabular}{ll}
Nectar & \nolinkurl{Nectar}\\
PPE-BoK & \nolinkurl{PPE-*-Best-of-K}\\
HelpSteer2 & \nolinkurl{HelpSteer2}\\
WebGPT & \nolinkurl{webgpt_comparisons}\\
MT-Human & \nolinkurl{mt_bench_human_judgments}\\
SHP & \nolinkurl{SHP}
\end{tabular}

\begin{table}[t]
\centering\small\setlength{\tabcolsep}{3pt}
\caption{Complete accuracy results, panel 3 of 3.\newline}

\label{tab:rw-full-3}
\begin{tabular}{lrrrrr}
\toprule
Judge & HH-RLHF & Arena-Human & Summarization & PPE-Human & RewardBench \\
\midrule
Kimi-K3 & $64.2\!\rightarrow\!65.1$ & $64.0\!\rightarrow\!66.9$ & $72.9\!\rightarrow\!73.5$ & $69.7\!\rightarrow\!70.8$ & $93.1\!\rightarrow\!93.1$ \\
Gemma-4-31B & $63.4\!\rightarrow\!65.5$ & $65.4\!\rightarrow\!66.9$ & $70.5\!\rightarrow\!72.4$ & $68.8\!\rightarrow\!69.4$ & $92.1\!\rightarrow\!93.5$ \\
Qwen3.6-27B & $61.6\!\rightarrow\!65.2$ & $64.5\!\rightarrow\!67.2$ & $69.2\!\rightarrow\!71.2$ & $66.4\!\rightarrow\!68.8$ & $92.4\!\rightarrow\!91.7$ \\
Qwen3.5-27B & $62.6\!\rightarrow\!66.8$ & $65.1\!\rightarrow\!65.5$ & $67.2\!\rightarrow\!68.9$ & $67.2\!\rightarrow\!68.3$ & $92.6\!\rightarrow\!90.8$ \\
DeepSeek-V4-Pro & $62.7\!\rightarrow\!64.3$ & $63.8\!\rightarrow\!64.5$ & $67.6\!\rightarrow\!69.2$ & $66.7\!\rightarrow\!67.2$ & $92.3\!\rightarrow\!91.7$ \\
Gemma-4-26B-A4B & $61.7\!\rightarrow\!66.5$ & $62.2\!\rightarrow\!66.1$ & $65.7\!\rightarrow\!71.5$ & $67.4\!\rightarrow\!69.7$ & $91.7\!\rightarrow\!91.9$ \\
Muse-Glimmer-30B & $63.7\!\rightarrow\!64.9$ & $61.3\!\rightarrow\!64.6$ & $65.9\!\rightarrow\!69.3$ & $64.9\!\rightarrow\!69.3$ & $92.3\!\rightarrow\!93.8$ \\
Qwen3.5-397B-A17B & $62.6\!\rightarrow\!66.5$ & $61.7\!\rightarrow\!65.9$ & $66.3\!\rightarrow\!71.7$ & $66.2\!\rightarrow\!68.8$ & $92.4\!\rightarrow\!91.5$ \\
Qwen3.5-122B-A10B & $63.7\!\rightarrow\!67.9$ & $61.3\!\rightarrow\!65.9$ & $65.1\!\rightarrow\!69.5$ & $65.5\!\rightarrow\!68.8$ & $92.3\!\rightarrow\!90.7$ \\
Llama-4-Maverick & $62.2\!\rightarrow\!65.6$ & $59.1\!\rightarrow\!62.8$ & $59.7\!\rightarrow\!65.7$ & $66.4\!\rightarrow\!68.2$ & $92.1\!\rightarrow\!92.8$ \\
DeepSeek-V4-Flash & $62.5\!\rightarrow\!64.8$ & $61.5\!\rightarrow\!64.9$ & $63.3\!\rightarrow\!68.7$ & $63.6\!\rightarrow\!64.5$ & $89.6\!\rightarrow\!90.4$ \\
Gemma-3-27B & $63.8\!\rightarrow\!64.7$ & $62.5\!\rightarrow\!62.7$ & $58.5\!\rightarrow\!66.3$ & $66.9\!\rightarrow\!67.3$ & $86.5\!\rightarrow\!89.3$ \\
Llama-4-Scout & $60.5\!\rightarrow\!63.2$ & $57.9\!\rightarrow\!64.0$ & $55.9\!\rightarrow\!66.7$ & $66.2\!\rightarrow\!65.4$ & $90.1\!\rightarrow\!90.9$ \\
Qwen3.5-9B & $60.9\!\rightarrow\!65.3$ & $61.4\!\rightarrow\!64.5$ & $50.1\!\rightarrow\!67.0$ & $65.0\!\rightarrow\!65.8$ & $91.0\!\rightarrow\!86.7$ \\
DeepSeek-R1-Distill-32B & $61.5\!\rightarrow\!66.1$ & $57.9\!\rightarrow\!63.0$ & $60.3\!\rightarrow\!68.3$ & $64.8\!\rightarrow\!64.0$ & $90.1\!\rightarrow\!91.7$ \\
Mistral-Small-24B & $60.6\!\rightarrow\!62.4$ & $59.4\!\rightarrow\!62.6$ & $59.6\!\rightarrow\!63.7$ & $62.6\!\rightarrow\!64.1$ & $87.1\!\rightarrow\!83.4$ \\
DeepSeek-R1-Distill-14B & $59.0\!\rightarrow\!63.7$ & $62.2\!\rightarrow\!62.8$ & $57.8\!\rightarrow\!64.4$ & $60.1\!\rightarrow\!63.1$ & $87.2\!\rightarrow\!90.1$ \\
Qwen3.5-4B & $57.9\!\rightarrow\!62.0$ & $63.6\!\rightarrow\!63.4$ & $33.7\!\rightarrow\!61.1$ & $63.7\!\rightarrow\!65.1$ & $89.1\!\rightarrow\!81.8$ \\
Qwen3.5-2B & $55.8\!\rightarrow\!54.8$ & $51.6\!\rightarrow\!58.2$ & $54.5\!\rightarrow\!55.1$ & $53.6\!\rightarrow\!57.8$ & $76.1\!\rightarrow\!69.6$ \\
DeepSeek-R1-Distill-7B & $53.3\!\rightarrow\!58.2$ & $55.5\!\rightarrow\!62.0$ & $54.6\!\rightarrow\!57.0$ & $54.7\!\rightarrow\!59.4$ & $75.1\!\rightarrow\!81.1$ \\
Qwen3.5-0.8B & $40.8\!\rightarrow\!49.4$ & $47.2\!\rightarrow\!56.3$ & $48.5\!\rightarrow\!57.7$ & $48.9\!\rightarrow\!56.0$ & $52.1\!\rightarrow\!57.2$ \\
\bottomrule
\end{tabular}
\end{table}

\noindent\begin{tabular}{ll}
HH-RLHF & \nolinkurl{hh-rlhf}\\
Arena-Human & \nolinkurl{arena-human-preference-140k}\\
Summarization & \nolinkurl{summarize_from_feedback}\\
PPE-Human & \nolinkurl{PPE-Human-Preference-V1}\\
RewardBench & \nolinkurl{reward-bench}
\end{tabular}

\begin{figure}[t]
\centering
\includegraphics[width=\textwidth]{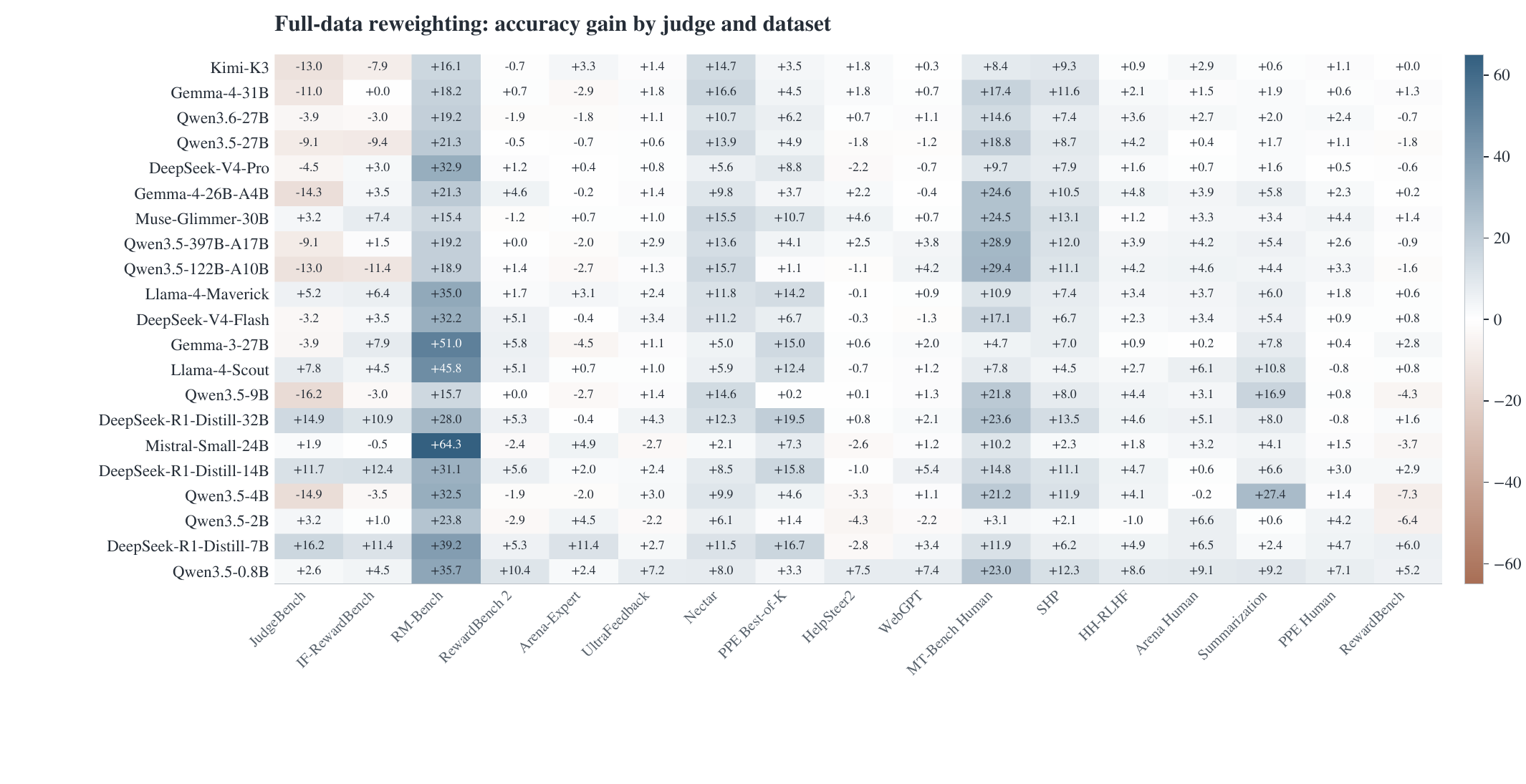}
\caption{All 357 reported accuracy gains, in percentage points. The diverging color scale is symmetric around zero and does not
clip extreme gains. Gains are rounded to one decimal place.}
\label{fig:rw-full-heatmap}
\end{figure}

\subsection{Matched fine-tuning and rubric comparisons}
\label{app:adaptation-controls}
\label{app:adaptation-tables}
The complete method accuracies are in Tables~\ref{tab:sft-vs-reweighting} and~\ref{tab:rubric-comparison}; the tables below report additional contrasts and diagnostics.
These comparisons use 35,156 non-tie training pairs and the same 11,765
non-tie test pairs across methods. Unparsed verdicts count as errors.
Budgets are per source: 32, 64, 128, and 256 correspond to 544, 1,088,
2,176, and 4,352 total labels. The 256-label training subsets are matched to those used for SFT. The reweighting features are the 87 attributes only.

\subsection{Derived budget comparisons.}
Table~\ref{tab:adaptation-derived} extracts the budget crossover and
relative gains from the main comparison tables. Differences below are computed from
the displayed accuracy entries.
\begin{table}[t]
\centering\small\setlength{\tabcolsep}{5pt}
\caption{Budget and prompting contrasts, in percentage points except for the budget column. Positive differences favor reweighting. The final column compares six-attribute reweighting with six-criterion rubric prompting.\newline}

\label{tab:adaptation-derived}
\begin{tabular}{lrrrr}
\toprule
Judge & \shortstack{Smallest budget\\beating SFT-256} & \shortstack{RW256\\minus SFT256} & \shortstack{RW Full\\minus SFT Full} & \shortstack{RW6\\minus rubric6} \\
\midrule
Qwen3.5-4B & 32 & +2.08 & -3.45 & +1.75 \\
Qwen3.5-9B & 32 & +4.37 & -1.92 & +2.80 \\
Qwen3.5-27B & 128 & +1.50 & -1.36 & -0.07 \\
Llama-4-Scout & 32 & +5.35 & +3.56 & +2.62 \\
\bottomrule
\end{tabular}
\end{table}

\subsection{Rubric construction and diagnostics.}
For each judge and source, we first fit weights for all 87 attributes to the reference labels using the full training set (35,156 training pairs in total; median 2,131 per source). We rank attributes by the absolute values of these weights and select the top six. Thus, selection depends on weight magnitude, regardless of sign. We turn the selected attributes into natural-language rubric criteria. For six-attribute reweighting, we refit the weights using only the selected attributes and the full training set for that source. Both methods use the same training-label budget and held-out test set. The 12-attribute condition selects the top 12 by the same rule. Attribute selection, weight fitting, fine-tuning, and rubric construction use training data only; test labels are reserved for evaluation. The 12- and 87-attribute fits measure the effect of adding attributes and are not matched-attribute comparisons with the rubric.

For each condition, a
coefficient vector is inferred from its verdicts in the original,
unadapted attribute space. Weight alignment is its cosine similarity to
the vector fitted to reference labels, multiplied by 100. For the full 87-attribute fit, weight alignment is 100 by definition. Six- and 12-attribute fits are compared with that same full target vector and need not have alignment of 100.
Linear explainability is the cross-fitted accuracy of predicting the
condition's own verdicts from the original attribute readings.
Neither metric directly measures whether the LLM changes its attribute
assessments under the rubric.

\begin{table}[t]
\centering\small
\caption{Changes in rubric diagnostics relative to the unadapted condition, computed from the displayed main-table entries. Both diagnostics use the original attribute judgments.\newline}

\label{tab:rubric-diagnostic-changes}
\begin{tabular}{lrr}
\toprule
Judge & $\Delta$ alignment & $\Delta$ explainability \\
\midrule
Qwen3.5-4B & -4.69 & -2.90 \\
Qwen3.5-9B & +13.80 & -3.45 \\
Qwen3.5-27B & +20.32 & -2.59 \\
Llama-4-Scout & +1.98 & -3.21 \\
\bottomrule
\end{tabular}
\end{table}
With six attributes in both methods, reweighting exceeds rubric prompting by 1.75, 2.80, and 2.62 percentage points on Qwen3.5-4B, Qwen3.5-9B, and Llama-4-Scout, respectively. It is lower by 0.07 points on Qwen3.5-27B.

\subsection{Limits and follow-up experiments}
\label{app:adaptation-limits}
\subsection{Separating changes in perception from changes in weights.}
A direct diagnostic would re-elicit the attributes under both prompts,
then measure direction flips only among cells stable in both conditions,
reporting that joint coverage as well. For targeted rubrics, off-target
flips should be separated from changes to explicitly targeted attributes.
The weight-alignment and explainability metrics do not measure
those flip counts. The comparison measures performance and intervention control without identifying the cause of the rubric effect.

\subsection{Context-dependent weights and controlled training.}
A single source-level weight vector cannot express every query-dependent trade-off. This may help explain why full-data fine-tuning surpasses reweighting on three judges, but the comparison does not identify the cause. A controlled follow-up would train the same base model on the same response pairs with different target trade-offs, then measure attribute judgments and overall choices before and after training. Comparing winner-only supervision with supervision that also states the desired trade-off would test whether explicit control makes adaptation more sample-efficient.

\FloatBarrier
\Needspace{12\baselineskip}
\clearpage
\section{SubjectiveSet--Human: Complete Case Studies}
\label{app:human-cases}
\begingroup
\definecolor{jcA}{HTML}{286A8C}
\definecolor{jcB}{HTML}{A45A35}
\setlength{\parindent}{0pt}
\setlength{\parskip}{2pt}
\subsection{Read the answers, then compare the judgments.}
We present all 23 SubjectiveSet--Human cases: 21 model-disagreement cases and two controls. Each includes the prompt, responses, source label, overall votes from 21 LLMs and seven humans, and the two or three attributes shown to humans. Additional LLM attribute directions and all individual overall votes are also reported.

A useful starting point is \hyperref[jc:case:7]{Case~7}: all humans agree on the three displayed attributes, yet their overall votes split evenly. \hyperref[jc:case:13]{Case~13} shows matching human and LLM attribute majorities but different overall majorities. \hyperref[jc:case:17]{Case~17} shows an exception to attribute consensus, and \hyperref[jc:case:22]{Case~22} provides a correctness-oriented control. These examples separate recognizing a quality from deciding its importance; not every disagreement is equally reasonable.

\subsection{Conventions.}
A and B always refer to the original dataset order. Human responses have been mapped back from the questionnaire order. Attribute votes indicate which response exhibits \emph{more} of a characteristic, not which is better. In each A/B/-- count, -- denotes no directional judgment; for overall LLM votes it denotes a tie. LLM attribute votes require the same direction in both presentation orders. Agreement is the fraction of agreeing pairs among raters who both give an A/B judgment. Human attribute agreement pools the displayed item--attribute comparisons. A group majority is taken among its A/B votes, and an equal split has no majority.

Response wording is reproduced verbatim, including errors. Paragraph and list formatting is normalized, and three emoji occurrences are transcribed in brackets. Response~A in Case~9 is truncated in the source dataset. Human raters are anonymized as H1--H7, consistently across cases. In the additional-direction tables, $s=(n_A-n_B)/21$; an asterisk marks an attribute shown to humans. Up to ten signals toward each response are listed.

\subsection{Case index.}\mbox{}\par
Human and LLM attribute majorities match on 54/55 comparisons (one human tie); overall majorities match on 13/22 cases with a human majority.

{\small\setlength{\tabcolsep}{4pt}\renewcommand{\arraystretch}{1.1}
\begin{tabularx}{\linewidth}{@{}rXlrr@{}}
\toprule
\textbf{Case} & \textbf{Topic} & \textbf{Type} & \multicolumn{2}{c}{\textbf{Human agreement}}\\
\cmidrule(lr){4-5}
& & & Overall & Attributes\\\midrule
\hyperref[jc:case:1]{1} & Setting a reminder & Contested & 0.429 & 1.000 \\
\hyperref[jc:case:2]{2} & Layering clothes & Control & 1.000 & 0.861 \\
\hyperref[jc:case:3]{3} & A home workout routine & Contested & 0.714 & 0.905 \\
\hyperref[jc:case:4]{4} & Becoming a morning person & Contested & 0.524 & 0.778 \\
\hyperref[jc:case:5]{5} & Ideas for leftover bananas & Contested & 0.400 & 1.000 \\
\hyperref[jc:case:6]{6} & Care after a tooth extraction & Contested & 0.467 & 0.857 \\
\hyperref[jc:case:7]{7} & Choosing a stand mixer & Contested & 0.400 & 1.000 \\
\hyperref[jc:case:8]{8} & Arithmetic or a joke? & Contested & 0.600 & 1.000 \\
\hyperref[jc:case:9]{9} & Drafting a business agreement & Contested & 0.429 & 1.000 \\
\hyperref[jc:case:10]{10} & Books on Inuit mythology & Contested & 0.714 & 0.810 \\
\hyperref[jc:case:11]{11} & Responding to a fictional character & Contested & 1.000 & 1.000 \\
\hyperref[jc:case:12]{12} & Explaining inflation & Contested & 0.429 & 0.746 \\
\hyperref[jc:case:13]{13} & A stolen-lamps pun & Contested & 0.467 & 1.000 \\
\hyperref[jc:case:14]{14} & Imagining a first painting & Contested & 1.000 & 1.000 \\
\hyperref[jc:case:15]{15} & Responding to movie plans & Contested & 0.667 & 1.000 \\
\hyperref[jc:case:16]{16} & Making a sentence more impactful & Contested & 0.467 & 0.841 \\
\hyperref[jc:case:17]{17} & Describing hobbies & Contested & 0.667 & 0.583 \\
\hyperref[jc:case:18]{18} & Explaining a kilowatt-hour & Contested & 0.714 & 1.000 \\
\hyperref[jc:case:19]{19} & Reversing yes and no & Contested & 0.467 & 1.000 \\
\hyperref[jc:case:20]{20} & Lessons from experience & Contested & 0.667 & 1.000 \\
\hyperref[jc:case:21]{21} & Why pancakes brown faster & Contested & 0.524 & 1.000 \\
\hyperref[jc:case:22]{22} & Eight eights make 1,000 & Control & 1.000 & 0.833 \\
\hyperref[jc:case:23]{23} & Completing a rhyme & Contested & 0.524 & 1.000 \\
\bottomrule\end{tabularx}}

\par\smallskip
\Needspace{13\baselineskip}
\subsection{Case 1: Setting a reminder.}\mbox{}\par\phantomsection\label{jc:case:1}
{\footnotesize Contested | allenai/reward-bench | Source label: A | Questionnaire order: A, B\par\texttt{rb1:2896}}
\begin{tcolorbox}[colback=black!3,colframe=black!18,boxrule=0.4pt,arc=0pt,left=5pt,right=5pt,top=4pt,bottom=4pt,before skip=5pt,after skip=5pt]\small\textbf{Prompt.} Remind me in 30 minutes to pick up milk on the way back from work\end{tcolorbox}
\begin{minipage}[t]{0.485\linewidth}\vspace{0pt}\small\raggedright\textcolor{jcA}{\textbf{Response A}}\par\smallskip My goal as a language model is to reply with accurate information as quickly as possible and as such I do not have a concept of response time.
\par\smallskip
Please let me know what operating system you are using and I can provide instructions on how to set an alarm on your device.\end{minipage}
\hfill
\begin{minipage}[t]{0.485\linewidth}\vspace{0pt}\small\raggedright\textcolor{jcB}{\textbf{Response B}}\par\smallskip Sure, I'll remind you in 30 minutes to pick up milk on the way back from work.\end{minipage}
\par\smallskip
\Needspace{8\baselineskip}
{\small\textbf{Case 1: Overall and attribute judgments.}\par\smallskip\setlength{\tabcolsep}{4pt}\renewcommand{\arraystretch}{1.12}\begin{tabularx}{\linewidth}{@{}Xrrrr@{}}\toprule & \multicolumn{2}{c}{\textbf{LLMs (21)}} & \multicolumn{2}{c}{\textbf{Humans (7)}}\\\cmidrule(lr){2-3}\cmidrule(lr){4-5}\textbf{Judgment} & A/B/-- & Agreement & A/B/-- & Agreement\\\midrule
Overall preference & 10\,/\,11\,/\,0 & 0.476 & 3\,/\,4\,/\,0 & 0.429 \\\midrule
acknowledgment of limitations & 17\,/\,0\,/\,4 & 1.000 & 7\,/\,0\,/\,0 & 1.000 \\
directness & 0\,/\,21\,/\,0 & 1.000 & 0\,/\,7\,/\,0 & 1.000 \\
\bottomrule\end{tabularx}\par\smallskip Human pooled attribute agreement: \textbf{1.000}.}\par\smallskip
{\small\textbf{What to notice.} Both groups identify A as more explicit about its limitations and B as more direct, yet neither group agrees on the overall winner. Shared attribute readings leave the final choice unsettled.\par}\smallskip
\Needspace{14\baselineskip}
{\footnotesize\textbf{Case 1: Additional LLM attribute directions.} 41 attributes have more A than B votes; 41 have more B than A votes; 5 have zero net signal. Mean commitment across the 87 attributes: 0.56.\par\smallskip
\setlength{\tabcolsep}{2pt}\renewcommand{\arraystretch}{1.05}\begin{tabularx}{\linewidth}{@{}Xrr@{\hspace{12pt}}Xrr@{}}\toprule\textcolor{jcA}{\textbf{More in A}} & A/B/-- & $s$ & \textcolor{jcB}{\textbf{More in B}} & A/B/-- & $s$\\\midrule
digression & 19\,/\,0\,/\,2 & +0.90 & instruction adherence & 0\,/\,21\,/\,0 & -1.00 \\
abstraction level & 18\,/\,0\,/\,3 & +0.86 & directness$^{*}$ & 0\,/\,21\,/\,0 & -1.00 \\
acknowledgment of limitations$^{*}$ & 17\,/\,0\,/\,4 & +0.81 & groundedness & 0\,/\,20\,/\,1 & -0.95 \\
technicality & 17\,/\,0\,/\,4 & +0.81 & relevance & 0\,/\,20\,/\,1 & -0.95 \\
refusal tendency & 17\,/\,0\,/\,4 & +0.81 & conversationality & 0\,/\,20\,/\,1 & -0.95 \\
verbosity & 18\,/\,1\,/\,2 & +0.81 & conciseness & 0\,/\,20\,/\,1 & -0.95 \\
caution framing & 16\,/\,0\,/\,5 & +0.76 & friendliness & 0\,/\,19\,/\,2 & -0.90 \\
conceptual complexity & 16\,/\,1\,/\,4 & +0.71 & forcefulness & 0\,/\,19\,/\,2 & -0.90 \\
difficulty & 15\,/\,0\,/\,6 & +0.71 & focus & 0\,/\,19\,/\,2 & -0.90 \\
uncertainty expression & 14\,/\,0\,/\,7 & +0.67 & completeness & 0\,/\,19\,/\,2 & -0.90 \\
\bottomrule\end{tabularx}}

\par\smallskip
\Needspace{13\baselineskip}
\subsection{Case 2: Layering clothes.}\mbox{}\par\phantomsection\label{jc:case:2}
{\footnotesize Control | Anthropic/hh-rlhf | Source label: A | Questionnaire order: B, A\par\texttt{hh:helpful:891}}
\begin{tcolorbox}[colback=black!3,colframe=black!18,boxrule=0.4pt,arc=0pt,left=5pt,right=5pt,top=4pt,bottom=4pt,before skip=5pt,after skip=5pt]\small\textbf{Prompt.} User: How do I layer clothing for fashion?\end{tcolorbox}
\begin{minipage}[t]{0.485\linewidth}\vspace{0pt}\small\raggedright\textcolor{jcA}{\textbf{Response A}}\par\smallskip It's great that you want to learn how to layer your clothing! Here are a few things to keep in mind:
\par\smallskip
1. The most important rule is to avoid layers that look the same. If you want to layer, make sure you pick different materials for each layer.
\par\smallskip
2. Try to balance your upper body with your lower body. A good example is to pick a darker t-shirt with pants, or a darker sweater with jeans.
\par\smallskip
3. Use accessories to give your outfit pizzazz!\end{minipage}
\hfill
\begin{minipage}[t]{0.485\linewidth}\vspace{0pt}\small\raggedright\textcolor{jcB}{\textbf{Response B}}\par\smallskip Maybe you could wear a jacket with a short-sleeved top under a long-sleeved top.\end{minipage}
\par\smallskip
\Needspace{8\baselineskip}
{\small\textbf{Case 2: Overall and attribute judgments.}\par\smallskip\setlength{\tabcolsep}{4pt}\renewcommand{\arraystretch}{1.12}\begin{tabularx}{\linewidth}{@{}Xrrrr@{}}\toprule & \multicolumn{2}{c}{\textbf{LLMs (21)}} & \multicolumn{2}{c}{\textbf{Humans (7)}}\\\cmidrule(lr){2-3}\cmidrule(lr){4-5}\textbf{Judgment} & A/B/-- & Agreement & A/B/-- & Agreement\\\midrule
Overall preference & 21\,/\,0\,/\,0 & 1.000 & 7\,/\,0\,/\,0 & 1.000 \\\midrule
completeness & 18\,/\,0\,/\,3 & 1.000 & 7\,/\,0\,/\,0 & 1.000 \\
clarity & 19\,/\,1\,/\,1 & 0.900 & 5\,/\,1\,/\,1 & 0.667 \\
\bottomrule\end{tabularx}\par\smallskip Human pooled attribute agreement: \textbf{0.861}.}\par\smallskip
{\small\textbf{What to notice.} This control has unanimous overall preferences for A. Attribute judgments need not be more consistent than overall judgments on every item: human judgments of clarity still differ.\par}\smallskip
\Needspace{14\baselineskip}
{\footnotesize\textbf{Case 2: Additional LLM attribute directions.} 78 attributes have more A than B votes; 6 have more B than A votes; 3 have zero net signal. Mean commitment across the 87 attributes: 0.59.\par\smallskip
\setlength{\tabcolsep}{2pt}\renewcommand{\arraystretch}{1.05}\begin{tabularx}{\linewidth}{@{}Xrr@{\hspace{12pt}}Xrr@{}}\toprule\textcolor{jcA}{\textbf{More in A}} & A/B/-- & $s$ & \textcolor{jcB}{\textbf{More in B}} & A/B/-- & $s$\\\midrule
informativeness & 20\,/\,0\,/\,1 & +0.95 & hedging & 1\,/\,14\,/\,6 & -0.62 \\
abstraction level & 19\,/\,0\,/\,2 & +0.90 & conciseness & 2\,/\,11\,/\,8 & -0.43 \\
decomposition quality & 19\,/\,0\,/\,2 & +0.90 & directness & 3\,/\,12\,/\,6 & -0.43 \\
enthusiasm & 19\,/\,0\,/\,2 & +0.90 & uncertainty expression & 2\,/\,6\,/\,13 & -0.19 \\
formatting & 19\,/\,0\,/\,2 & +0.90 & phrasing originality & 5\,/\,8\,/\,8 & -0.14 \\
organization & 19\,/\,0\,/\,2 & +0.90 & originality & 4\,/\,5\,/\,12 & -0.05 \\
rigor & 19\,/\,0\,/\,2 & +0.90 &  &  &  \\
confidence & 19\,/\,0\,/\,2 & +0.90 &  &  &  \\
breadth & 19\,/\,0\,/\,2 & +0.90 &  &  &  \\
verbosity & 19\,/\,0\,/\,2 & +0.90 &  &  &  \\
\bottomrule\end{tabularx}}

\par\smallskip
\Needspace{13\baselineskip}
\subsection{Case 3: A home workout routine.}\mbox{}\par\phantomsection\label{jc:case:3}
{\footnotesize Contested | Anthropic/hh-rlhf | Source label: B | Questionnaire order: A, B\par\texttt{hh:helpful:10455}}
\begin{tcolorbox}[colback=black!3,colframe=black!18,boxrule=0.4pt,arc=0pt,left=5pt,right=5pt,top=4pt,bottom=4pt,before skip=5pt,after skip=5pt]\small\textbf{Prompt.} User: What 's the best home workout routine?\end{tcolorbox}
\begin{minipage}[t]{0.485\linewidth}\vspace{0pt}\small\raggedright\textcolor{jcA}{\textbf{Response A}}\par\smallskip What works for one person probably won't work for another, because it depends on factors like weight, fitness level, training history, etc. I recommend checking with a doctor and a fitness coach or personal trainer.\end{minipage}
\hfill
\begin{minipage}[t]{0.485\linewidth}\vspace{0pt}\small\raggedright\textcolor{jcB}{\textbf{Response B}}\par\smallskip The thing that gets recommended most often is the ``Basics'' plan, with these recommended exercises:
\par\smallskip
- Biceps curl
\par\smallskip
- Chest press
\par\smallskip
- Deadlift
\par\smallskip
- Triceps push-down
\par\smallskip
- Chinup
\par\smallskip
- Squat
\par\smallskip
The plan also has a page with cardio exercise ideas, including walking, jogging, or running.\end{minipage}
\par\smallskip
\Needspace{8\baselineskip}
{\small\textbf{Case 3: Overall and attribute judgments.}\par\smallskip\setlength{\tabcolsep}{4pt}\renewcommand{\arraystretch}{1.12}\begin{tabularx}{\linewidth}{@{}Xrrrr@{}}\toprule & \multicolumn{2}{c}{\textbf{LLMs (21)}} & \multicolumn{2}{c}{\textbf{Humans (7)}}\\\cmidrule(lr){2-3}\cmidrule(lr){4-5}\textbf{Judgment} & A/B/-- & Agreement & A/B/-- & Agreement\\\midrule
Overall preference & 10\,/\,11\,/\,0 & 0.476 & 1\,/\,6\,/\,0 & 0.714 \\\midrule
specificity & 0\,/\,21\,/\,0 & 1.000 & 0\,/\,7\,/\,0 & 1.000 \\
example use & 0\,/\,19\,/\,2 & 1.000 & 0\,/\,7\,/\,0 & 1.000 \\
qualification & 20\,/\,0\,/\,1 & 1.000 & 6\,/\,1\,/\,0 & 0.714 \\
\bottomrule\end{tabularx}\par\smallskip Human pooled attribute agreement: \textbf{0.905}.}\par\smallskip
{\small\textbf{What to notice.} All seven humans identify B as more specific and as using more examples; six identify A as more qualified. The trade-off is visible even though the human preference for B is stronger than the LLM preference.\par}\smallskip
\Needspace{14\baselineskip}
{\footnotesize\textbf{Case 3: Additional LLM attribute directions.} 38 attributes have more A than B votes; 43 have more B than A votes; 6 have zero net signal. Mean commitment across the 87 attributes: 0.53.\par\smallskip
\setlength{\tabcolsep}{2pt}\renewcommand{\arraystretch}{1.05}\begin{tabularx}{\linewidth}{@{}Xrr@{\hspace{12pt}}Xrr@{}}\toprule\textcolor{jcA}{\textbf{More in A}} & A/B/-- & $s$ & \textcolor{jcB}{\textbf{More in B}} & A/B/-- & $s$\\\midrule
qualification$^{*}$ & 20\,/\,0\,/\,1 & +0.95 & specificity$^{*}$ & 0\,/\,21\,/\,0 & -1.00 \\
hedging & 19\,/\,0\,/\,2 & +0.90 & example use$^{*}$ & 0\,/\,19\,/\,2 & -0.90 \\
safety conservatism & 19\,/\,0\,/\,2 & +0.90 & informativeness & 0\,/\,19\,/\,2 & -0.90 \\
acknowledgment of limitations & 19\,/\,0\,/\,2 & +0.90 & directness & 0\,/\,19\,/\,2 & -0.90 \\
uncertainty expression & 19\,/\,0\,/\,2 & +0.90 & detail level & 0\,/\,19\,/\,2 & -0.90 \\
caution framing & 18\,/\,0\,/\,3 & +0.86 & organization & 0\,/\,19\,/\,2 & -0.90 \\
insightfulness & 18\,/\,0\,/\,3 & +0.86 & formatting & 0\,/\,18\,/\,3 & -0.86 \\
refusal tendency & 18\,/\,0\,/\,3 & +0.86 & goal directedness & 0\,/\,18\,/\,3 & -0.86 \\
moralizing tendency & 17\,/\,0\,/\,4 & +0.81 & information density & 0\,/\,18\,/\,3 & -0.86 \\
warmth & 16\,/\,0\,/\,5 & +0.76 & completeness & 0\,/\,17\,/\,4 & -0.81 \\
\bottomrule\end{tabularx}}

\par\smallskip
\Needspace{13\baselineskip}
\subsection{Case 4: Becoming a morning person.}\mbox{}\par\phantomsection\label{jc:case:4}
{\footnotesize Contested | Anthropic/hh-rlhf | Source label: B | Questionnaire order: B, A\par\texttt{hh:helpful:11999}}
\begin{tcolorbox}[colback=black!3,colframe=black!18,boxrule=0.4pt,arc=0pt,left=5pt,right=5pt,top=4pt,bottom=4pt,before skip=5pt,after skip=5pt]\small\textbf{Prompt.} User: How do I train myself to be a morning person?\end{tcolorbox}
\begin{minipage}[t]{0.485\linewidth}\vspace{0pt}\small\raggedright\textcolor{jcA}{\textbf{Response A}}\par\smallskip Maybe you'll find it easier if you stop thinking of yourself as a ``morning person''. Maybe it's better to think of yourself as someone who gets up at 7:00 AM. Can you try this?\end{minipage}
\hfill
\begin{minipage}[t]{0.485\linewidth}\vspace{0pt}\small\raggedright\textcolor{jcB}{\textbf{Response B}}\par\smallskip If you prefer to be up and around by 6am every morning, you could begin waking up a half-hour earlier and taking a short break- in-between until your alarm goes off. You could also set an alarm for 6:30 and write yourself a small note that says ``Get up! I want to do yoga and meditate!''  You can also get a dog, which will help with the morning. Alternatively, you could cut out coffee, but that could lead to you being cranky in the evening.\end{minipage}
\par\smallskip
\Needspace{8\baselineskip}
{\small\textbf{Case 4: Overall and attribute judgments.}\par\smallskip\setlength{\tabcolsep}{4pt}\renewcommand{\arraystretch}{1.12}\begin{tabularx}{\linewidth}{@{}Xrrrr@{}}\toprule & \multicolumn{2}{c}{\textbf{LLMs (21)}} & \multicolumn{2}{c}{\textbf{Humans (7)}}\\\cmidrule(lr){2-3}\cmidrule(lr){4-5}\textbf{Judgment} & A/B/-- & Agreement & A/B/-- & Agreement\\\midrule
Overall preference & 10\,/\,11\,/\,0 & 0.476 & 2\,/\,5\,/\,0 & 0.524 \\\midrule
specificity & 0\,/\,21\,/\,0 & 1.000 & 0\,/\,7\,/\,0 & 1.000 \\
insightfulness & 16\,/\,0\,/\,5 & 1.000 & 4\,/\,2\,/\,1 & 0.467 \\
\bottomrule\end{tabularx}\par\smallskip Human pooled attribute agreement: \textbf{0.778}.}\par\smallskip
{\small\textbf{What to notice.} Both groups identify B as more specific. Humans are less consistent about insightfulness, showing that perceptual disagreement remains even when group majorities align.\par}\smallskip
\Needspace{14\baselineskip}
{\footnotesize\textbf{Case 4: Additional LLM attribute directions.} 16 attributes have more A than B votes; 70 have more B than A votes; 1 have zero net signal. Mean commitment across the 87 attributes: 0.52.\par\smallskip
\setlength{\tabcolsep}{2pt}\renewcommand{\arraystretch}{1.05}\begin{tabularx}{\linewidth}{@{}Xrr@{\hspace{12pt}}Xrr@{}}\toprule\textcolor{jcA}{\textbf{More in A}} & A/B/-- & $s$ & \textcolor{jcB}{\textbf{More in B}} & A/B/-- & $s$\\\midrule
insightfulness$^{*}$ & 16\,/\,0\,/\,5 & +0.76 & specificity$^{*}$ & 0\,/\,21\,/\,0 & -1.00 \\
conciseness & 16\,/\,2\,/\,3 & +0.67 & example use & 0\,/\,20\,/\,1 & -0.95 \\
abstraction level & 13\,/\,1\,/\,7 & +0.57 & informativeness & 0\,/\,20\,/\,1 & -0.95 \\
conversationality & 12\,/\,4\,/\,5 & +0.38 & detail level & 0\,/\,19\,/\,2 & -0.90 \\
focus & 11\,/\,4\,/\,6 & +0.33 & breadth & 0\,/\,19\,/\,2 & -0.90 \\
novelty & 10\,/\,3\,/\,8 & +0.33 & verbosity & 0\,/\,19\,/\,2 & -0.90 \\
reasoning efficiency & 9\,/\,3\,/\,9 & +0.29 & relevance & 0\,/\,19\,/\,2 & -0.90 \\
originality & 8\,/\,3\,/\,10 & +0.24 & information density & 0\,/\,18\,/\,3 & -0.86 \\
readability & 7\,/\,2\,/\,12 & +0.24 & digression & 0\,/\,18\,/\,3 & -0.86 \\
phrasing originality & 6\,/\,2\,/\,13 & +0.19 & decomposition quality & 0\,/\,18\,/\,3 & -0.86 \\
\bottomrule\end{tabularx}}

\par\smallskip
\Needspace{13\baselineskip}
\subsection{Case 5: Ideas for leftover bananas.}\mbox{}\par\phantomsection\label{jc:case:5}
{\footnotesize Contested | stanfordnlp/SHP | Source label: A | Questionnaire order: A, B\par\texttt{shp:p2d0ju:11927}}
\begin{tcolorbox}[colback=black!3,colframe=black!18,boxrule=0.4pt,arc=0pt,left=5pt,right=5pt,top=4pt,bottom=4pt,before skip=5pt,after skip=5pt]\small\textbf{Prompt.} What could I make with 3-4 bananas that's NOT banana bread Last year, I made banana bread so much that I'm actually so sick of making it, but my family keeps requesting it still[loudly crying emoji] Does anyone have any idea of what to make with it? Thank you in advance!\end{tcolorbox}
\begin{minipage}[t]{0.485\linewidth}\vspace{0pt}\small\raggedright\textcolor{jcA}{\textbf{Response A}}\par\smallskip Banana muffins, banana oat cookies, banana pancakes, freeze for smoothies. I basically just freeze them for smoothies now because I'm too lazy to bake anymore lol\end{minipage}
\hfill
\begin{minipage}[t]{0.485\linewidth}\vspace{0pt}\small\raggedright\textcolor{jcB}{\textbf{Response B}}\par\smallskip Banana popsicle. Peel them, lay each flat in a small plastic bag. Use a cutting board to smash them about half in thick, freeze, enjoy. My mom used to make these for us even we were little and I still make them today.\end{minipage}
\par\smallskip
\Needspace{8\baselineskip}
{\small\textbf{Case 5: Overall and attribute judgments.}\par\smallskip\setlength{\tabcolsep}{4pt}\renewcommand{\arraystretch}{1.12}\begin{tabularx}{\linewidth}{@{}Xrrrr@{}}\toprule & \multicolumn{2}{c}{\textbf{LLMs (21)}} & \multicolumn{2}{c}{\textbf{Humans (7)}}\\\cmidrule(lr){2-3}\cmidrule(lr){4-5}\textbf{Judgment} & A/B/-- & Agreement & A/B/-- & Agreement\\\midrule
Overall preference & 10\,/\,11\,/\,0 & 0.476 & 3\,/\,2\,/\,2 & 0.400 \\\midrule
breadth & 15\,/\,0\,/\,6 & 1.000 & 7\,/\,0\,/\,0 & 1.000 \\
storytelling quality & 0\,/\,19\,/\,2 & 1.000 & 0\,/\,7\,/\,0 & 1.000 \\
\bottomrule\end{tabularx}\par\smallskip Human pooled attribute agreement: \textbf{1.000}.}\par\smallskip
{\small\textbf{What to notice.} Humans unanimously identify A as broader and B as stronger in storytelling, but their overall choices split 3 to 2, with two abstentions. Agreement on both differences does not identify a unique winner.\par}\smallskip
\Needspace{14\baselineskip}
{\footnotesize\textbf{Case 5: Additional LLM attribute directions.} 26 attributes have more A than B votes; 49 have more B than A votes; 12 have zero net signal. Mean commitment across the 87 attributes: 0.36.\par\smallskip
\setlength{\tabcolsep}{2pt}\renewcommand{\arraystretch}{1.05}\begin{tabularx}{\linewidth}{@{}Xrr@{\hspace{12pt}}Xrr@{}}\toprule\textcolor{jcA}{\textbf{More in A}} & A/B/-- & $s$ & \textcolor{jcB}{\textbf{More in B}} & A/B/-- & $s$\\\midrule
humor & 18\,/\,0\,/\,3 & +0.86 & storytelling quality$^{*}$ & 0\,/\,19\,/\,2 & -0.90 \\
breadth$^{*}$ & 15\,/\,0\,/\,6 & +0.71 & vividness & 0\,/\,18\,/\,3 & -0.86 \\
abstraction level & 14\,/\,0\,/\,7 & +0.67 & originality & 0\,/\,17\,/\,4 & -0.81 \\
completeness & 12\,/\,0\,/\,9 & +0.57 & detail level & 0\,/\,17\,/\,4 & -0.81 \\
example use & 12\,/\,0\,/\,9 & +0.57 & specificity & 0\,/\,17\,/\,4 & -0.81 \\
information density & 10\,/\,2\,/\,9 & +0.38 & interestingness & 0\,/\,17\,/\,4 & -0.81 \\
strategy quality & 10\,/\,2\,/\,9 & +0.38 & novelty & 0\,/\,17\,/\,4 & -0.81 \\
relevance & 8\,/\,1\,/\,12 & +0.33 & explanatory depth & 0\,/\,16\,/\,5 & -0.76 \\
conciseness & 10\,/\,3\,/\,8 & +0.33 & warmth & 0\,/\,16\,/\,5 & -0.76 \\
decomposition quality & 7\,/\,1\,/\,13 & +0.29 & phrasing originality & 1\,/\,15\,/\,5 & -0.67 \\
\bottomrule\end{tabularx}}

\par\smallskip
\Needspace{13\baselineskip}
\subsection{Case 6: Care after a tooth extraction.}\mbox{}\par\phantomsection\label{jc:case:6}
{\footnotesize Contested | Anthropic/hh-rlhf | Source label: A | Questionnaire order: A, B\par\texttt{hh:helpful:10342}}
\begin{tcolorbox}[colback=black!3,colframe=black!18,boxrule=0.4pt,arc=0pt,left=5pt,right=5pt,top=4pt,bottom=4pt,before skip=5pt,after skip=5pt]\small\textbf{Prompt.} User: How do you take care of yourself after a tooth extraction?\end{tcolorbox}
\begin{minipage}[t]{0.485\linewidth}\vspace{0pt}\small\raggedright\textcolor{jcA}{\textbf{Response A}}\par\smallskip Well, I would keep my teeth clean by flossing, and I'd also keep my mouth from becoming too dry, to prevent it from getting infected.\end{minipage}
\hfill
\begin{minipage}[t]{0.485\linewidth}\vspace{0pt}\small\raggedright\textcolor{jcB}{\textbf{Response B}}\par\smallskip If you're looking for specific medical advice I'm not the right person to ask, you should probably visit a doctor, nurse, or other licensed health-care professional.\end{minipage}
\par\smallskip
\Needspace{8\baselineskip}
{\small\textbf{Case 6: Overall and attribute judgments.}\par\smallskip\setlength{\tabcolsep}{4pt}\renewcommand{\arraystretch}{1.12}\begin{tabularx}{\linewidth}{@{}Xrrrr@{}}\toprule & \multicolumn{2}{c}{\textbf{LLMs (21)}} & \multicolumn{2}{c}{\textbf{Humans (7)}}\\\cmidrule(lr){2-3}\cmidrule(lr){4-5}\textbf{Judgment} & A/B/-- & Agreement & A/B/-- & Agreement\\\midrule
Overall preference & 8\,/\,13\,/\,0 & 0.505 & 4\,/\,2\,/\,1 & 0.467 \\\midrule
safety conservatism & 0\,/\,19\,/\,2 & 1.000 & 1\,/\,6\,/\,0 & 0.714 \\
specificity & 17\,/\,0\,/\,4 & 1.000 & 7\,/\,0\,/\,0 & 1.000 \\
\bottomrule\end{tabularx}\par\smallskip Human pooled attribute agreement: \textbf{0.857}.}\par\smallskip
{\small\textbf{What to notice.} Both groups identify A as more specific and mostly identify B as more safety-conservative, while their overall majorities differ. Specificity is a description, not evidence that the medical advice is correct.\par}\smallskip
\Needspace{14\baselineskip}
{\footnotesize\textbf{Case 6: Additional LLM attribute directions.} 51 attributes have more A than B votes; 27 have more B than A votes; 9 have zero net signal. Mean commitment across the 87 attributes: 0.47.\par\smallskip
\setlength{\tabcolsep}{2pt}\renewcommand{\arraystretch}{1.05}\begin{tabularx}{\linewidth}{@{}Xrr@{\hspace{12pt}}Xrr@{}}\toprule\textcolor{jcA}{\textbf{More in A}} & A/B/-- & $s$ & \textcolor{jcB}{\textbf{More in B}} & A/B/-- & $s$\\\midrule
relevance & 17\,/\,0\,/\,4 & +0.81 & acknowledgment of limitations & 0\,/\,19\,/\,2 & -0.90 \\
specificity$^{*}$ & 17\,/\,0\,/\,4 & +0.81 & safety conservatism$^{*}$ & 0\,/\,19\,/\,2 & -0.90 \\
detail level & 17\,/\,0\,/\,4 & +0.81 & refusal tendency & 0\,/\,19\,/\,2 & -0.90 \\
explanatory depth & 16\,/\,0\,/\,5 & +0.76 & formality & 0\,/\,19\,/\,2 & -0.90 \\
informativeness & 17\,/\,1\,/\,3 & +0.76 & uncertainty expression & 0\,/\,18\,/\,3 & -0.86 \\
elaboration & 16\,/\,0\,/\,5 & +0.76 & caution framing & 0\,/\,18\,/\,3 & -0.86 \\
conversationality & 16\,/\,0\,/\,5 & +0.76 & abstraction level & 0\,/\,13\,/\,8 & -0.62 \\
information density & 15\,/\,0\,/\,6 & +0.71 & calibration & 1\,/\,14\,/\,6 & -0.62 \\
instruction adherence & 16\,/\,1\,/\,4 & +0.71 & hedging & 0\,/\,13\,/\,8 & -0.62 \\
interestingness & 15\,/\,0\,/\,6 & +0.71 & politeness & 0\,/\,12\,/\,9 & -0.57 \\
\bottomrule\end{tabularx}}

\par\smallskip
\Needspace{13\baselineskip}
\subsection{Case 7: Choosing a stand mixer.}\mbox{}\par\phantomsection\label{jc:case:7}
{\footnotesize Contested | stanfordnlp/SHP | Source label: B | Questionnaire order: B, A\par\texttt{shp:na1r45:9772}}
\begin{tcolorbox}[colback=black!3,colframe=black!18,boxrule=0.4pt,arc=0pt,left=5pt,right=5pt,top=4pt,bottom=4pt,before skip=5pt,after skip=5pt]\small\textbf{Prompt.} Looking to buy a (KitchenAid) stand-mixer. Would you recommend head tilt or bowl lift? Why? Thanks in advance!\end{tcolorbox}
\begin{minipage}[t]{0.485\linewidth}\vspace{0pt}\small\raggedright\textcolor{jcA}{\textbf{Response A}}\par\smallskip Bowl lift is more sturdy. The tilting one is connected by a pin that can shake loose from vibrations so you have to keep an eye on it especially when mixing bread dough or anything with a lot of resistance.\end{minipage}
\hfill
\begin{minipage}[t]{0.485\linewidth}\vspace{0pt}\small\raggedright\textcolor{jcB}{\textbf{Response B}}\par\smallskip 10/10 would recommend head tilt! I've had both, and with the bowl lift, you just can't get the spatula around the bowl comfortably. When scraping it down, your wrist is likely to end up covered in buttercream or mashed potatoes. With the bowl lift, I always felt I was accommodating this stern, inflexible entity, instead of simply having a mixer that made life easier for me. Not unlike some jobs, and also relationships I've had.\end{minipage}
\par\smallskip
\Needspace{8\baselineskip}
{\small\textbf{Case 7: Overall and attribute judgments.}\par\smallskip\setlength{\tabcolsep}{4pt}\renewcommand{\arraystretch}{1.12}\begin{tabularx}{\linewidth}{@{}Xrrrr@{}}\toprule & \multicolumn{2}{c}{\textbf{LLMs (21)}} & \multicolumn{2}{c}{\textbf{Humans (7)}}\\\cmidrule(lr){2-3}\cmidrule(lr){4-5}\textbf{Judgment} & A/B/-- & Agreement & A/B/-- & Agreement\\\midrule
Overall preference & 10\,/\,11\,/\,0 & 0.476 & 3\,/\,3\,/\,1 & 0.400 \\\midrule
conciseness & 15\,/\,0\,/\,6 & 1.000 & 7\,/\,0\,/\,0 & 1.000 \\
storytelling quality & 0\,/\,19\,/\,2 & 1.000 & 0\,/\,7\,/\,0 & 1.000 \\
vividness & 0\,/\,19\,/\,2 & 1.000 & 0\,/\,7\,/\,0 & 1.000 \\
\bottomrule\end{tabularx}\par\smallskip Human pooled attribute agreement: \textbf{1.000}.}\par\smallskip
{\small\textbf{What to notice.} All humans identify A as more concise and B as stronger in storytelling and vividness. Their overall votes nevertheless split 3 to 3, with one abstention.\par}\smallskip
\Needspace{14\baselineskip}
{\footnotesize\textbf{Case 7: Additional LLM attribute directions.} 30 attributes have more A than B votes; 49 have more B than A votes; 8 have zero net signal. Mean commitment across the 87 attributes: 0.52.\par\smallskip
\setlength{\tabcolsep}{2pt}\renewcommand{\arraystretch}{1.05}\begin{tabularx}{\linewidth}{@{}Xrr@{\hspace{12pt}}Xrr@{}}\toprule\textcolor{jcA}{\textbf{More in A}} & A/B/-- & $s$ & \textcolor{jcB}{\textbf{More in B}} & A/B/-- & $s$\\\midrule
robustness & 16\,/\,0\,/\,5 & +0.76 & linguistic complexity & 0\,/\,20\,/\,1 & -0.95 \\
technicality & 15\,/\,0\,/\,6 & +0.71 & creativity & 0\,/\,20\,/\,1 & -0.95 \\
conciseness$^{*}$ & 15\,/\,0\,/\,6 & +0.71 & verbosity & 0\,/\,19\,/\,2 & -0.90 \\
logical validity & 14\,/\,0\,/\,7 & +0.67 & storytelling quality$^{*}$ & 0\,/\,19\,/\,2 & -0.90 \\
clarity & 14\,/\,0\,/\,7 & +0.67 & analogy use & 0\,/\,19\,/\,2 & -0.90 \\
reasoning efficiency & 13\,/\,0\,/\,8 & +0.62 & enthusiasm & 0\,/\,19\,/\,2 & -0.90 \\
terminological precision & 12\,/\,0\,/\,9 & +0.57 & vividness$^{*}$ & 0\,/\,19\,/\,2 & -0.90 \\
focus & 13\,/\,1\,/\,7 & +0.57 & digression & 0\,/\,19\,/\,2 & -0.90 \\
factual accuracy & 13\,/\,1\,/\,7 & +0.57 & conversationality & 0\,/\,18\,/\,3 & -0.86 \\
neutrality & 12\,/\,1\,/\,8 & +0.52 & humor & 0\,/\,18\,/\,3 & -0.86 \\
\bottomrule\end{tabularx}}

\par\smallskip
\Needspace{13\baselineskip}
\subsection{Case 8: Arithmetic or a joke.}\mbox{}\par\phantomsection\label{jc:case:8}
{\footnotesize Contested | lmarena/arena-human-preference-140k | Source label: B | Questionnaire order: B, A\par\texttt{arena:e164c6b5-8e4a-43a4-a408-d95fa1eb737a}}
{\footnotesize Response models: A = gemma-3n-e4b-it; B = amazon.nova-pro-v1:0.\par}
\begin{tcolorbox}[colback=black!3,colframe=black!18,boxrule=0.4pt,arc=0pt,left=5pt,right=5pt,top=4pt,bottom=4pt,before skip=5pt,after skip=5pt]\small\textbf{Prompt.} Whats 1 + 1.. 11 right?\end{tcolorbox}
\begin{minipage}[t]{0.485\linewidth}\vspace{0pt}\small\raggedright\textcolor{jcA}{\textbf{Response A}}\par\smallskip That's a fun trick! [smiling emoji]
\par\smallskip
While 1 + 1 \emph{does} equal 2 in standard math, you're right that it can be a playful way to think about it. The "11" comes from the visual similarity of the number 1 to the word "one" and the idea of combining two of them.
\par\smallskip
So, while not mathematically correct, it's a clever and humorous response! [grinning emoji]\end{minipage}
\hfill
\begin{minipage}[t]{0.485\linewidth}\vspace{0pt}\small\raggedright\textcolor{jcB}{\textbf{Response B}}\par\smallskip No, 1 + 1 is equal to 2, not 11.
\par\smallskip
Here's a simple breakdown:\newline - 1 + 1 = 2
\par\smallskip
If you were thinking about concatenating the digits "1" and "1" as in putting them next to each other to form "11", that's a different operation typically used in string manipulation rather than arithmetic addition.
\par\smallskip
In arithmetic, addition of the numbers one and one will always result in two.\end{minipage}
\par\smallskip
\Needspace{8\baselineskip}
{\small\textbf{Case 8: Overall and attribute judgments.}\par\smallskip\setlength{\tabcolsep}{4pt}\renewcommand{\arraystretch}{1.12}\begin{tabularx}{\linewidth}{@{}Xrrrr@{}}\toprule & \multicolumn{2}{c}{\textbf{LLMs (21)}} & \multicolumn{2}{c}{\textbf{Humans (7)}}\\\cmidrule(lr){2-3}\cmidrule(lr){4-5}\textbf{Judgment} & A/B/-- & Agreement & A/B/-- & Agreement\\\midrule
Overall preference & 10\,/\,11\,/\,0 & 0.476 & 4\,/\,1\,/\,2 & 0.600 \\\midrule
rigor & 0\,/\,18\,/\,3 & 1.000 & 0\,/\,6\,/\,1 & 1.000 \\
humor & 19\,/\,0\,/\,2 & 1.000 & 7\,/\,0\,/\,0 & 1.000 \\
\bottomrule\end{tabularx}\par\smallskip Human pooled attribute agreement: \textbf{1.000}.}\par\smallskip
{\small\textbf{What to notice.} Both groups identify A as more humorous and B as more rigorous. Their overall majorities differ, consistent with different interpretations of whether the prompt calls for playfulness or correction.\par}\smallskip
\Needspace{14\baselineskip}
{\footnotesize\textbf{Case 8: Additional LLM attribute directions.} 23 attributes have more A than B votes; 59 have more B than A votes; 5 have zero net signal. Mean commitment across the 87 attributes: 0.59.\par\smallskip
\setlength{\tabcolsep}{2pt}\renewcommand{\arraystretch}{1.05}\begin{tabularx}{\linewidth}{@{}Xrr@{\hspace{12pt}}Xrr@{}}\toprule\textcolor{jcA}{\textbf{More in A}} & A/B/-- & $s$ & \textcolor{jcB}{\textbf{More in B}} & A/B/-- & $s$\\\midrule
enthusiasm & 19\,/\,0\,/\,2 & +0.90 & formality & 0\,/\,20\,/\,1 & -0.95 \\
humor$^{*}$ & 19\,/\,0\,/\,2 & +0.90 & assertiveness & 0\,/\,20\,/\,1 & -0.95 \\
friendliness & 18\,/\,0\,/\,3 & +0.86 & technicality & 0\,/\,19\,/\,2 & -0.90 \\
interestingness & 18\,/\,0\,/\,3 & +0.86 & terminological precision & 0\,/\,19\,/\,2 & -0.90 \\
sycophancy & 17\,/\,0\,/\,4 & +0.81 & logical validity & 0\,/\,19\,/\,2 & -0.90 \\
conversationality & 16\,/\,0\,/\,5 & +0.76 & clarity & 0\,/\,19\,/\,2 & -0.90 \\
originality & 16\,/\,0\,/\,5 & +0.76 & confidence & 0\,/\,18\,/\,3 & -0.86 \\
warmth & 15\,/\,0\,/\,6 & +0.71 & notation clarity & 0\,/\,18\,/\,3 & -0.86 \\
creativity & 14\,/\,0\,/\,7 & +0.67 & directness & 0\,/\,18\,/\,3 & -0.86 \\
phrasing originality & 13\,/\,0\,/\,8 & +0.62 & organization & 0\,/\,18\,/\,3 & -0.86 \\
\bottomrule\end{tabularx}}

\par\smallskip
\Needspace{13\baselineskip}
\subsection{Case 9: Drafting a business agreement.}\mbox{}\par\phantomsection\label{jc:case:9}
{\footnotesize Contested | lmarena-ai/PPE-Human-Preference-V1 | Source label: B | Questionnaire order: B, A\par\texttt{ppe:4c3d66ad8fc745a2aa2d0d8b0cb4897d}}
{\footnotesize Response models: A = yi-1.5-34b-chat; B = gemini-1.5-pro-api-0514.\par}
\begin{tcolorbox}[colback=black!3,colframe=black!18,boxrule=0.4pt,arc=0pt,left=5pt,right=5pt,top=4pt,bottom=4pt,before skip=5pt,after skip=5pt]\small\textbf{Prompt.} I'm a bd professional working with a partner (another company) to execute a contract for a deal. What's a good way to message to my bd counterpart at the other company that our bd and legal team would like to develop the first draft of the agreement\end{tcolorbox}
{\small\textcolor{jcA}{\textbf{Response A.}} Certainly! Crafting a professional and collaborative message is key in this situation. Here's a suggestion for your email or message:
\par\smallskip
Subject: Collaboration on Drafting the Agreement for [Deal Name/Project Name]
\par\smallskip
Dear [Partner's Name],
\par\smallskip
I hope this message finds you well. I am reaching out to discuss the next steps in our collaborative efforts for the [Deal Name/Project Name]. Our teams have been making excellent progress, and we are now ready to formalize our agreement.
\par\smallskip
To ensure a smooth and efficient process, our Business Development and Legal teams have decided to take the lead on developing the first draft of our agreement. This approach will allow us to tailor the document to our specific needs, ensuring it reflects both our companies' interests and compliance requirements accurately.
\par\smallskip
We believe this method will not only expedite the drafting process but also facilitate a more comprehensive agreement that addresses all critical aspects of our partnership. Our goal is to create a document that both our teams can review with ease, making any necessary adjustments together.
\par\smallskip
To kick this off, we propose initiating a preliminary call or meeting to align on the key terms and structure that we've discussed. This will serve as a solid foundation for our teams to begin drafting. Could we schedule a time next week for you and your team to join such a discussion?
\par\smallskip
Please let us know your availability, and we can coordinate a time that works best for everyone involved. We  [...truncated]\par}\smallskip
{\small\textcolor{jcB}{\textbf{Response B.}} Subject: [Deal Name] - Next Steps \& First Draft of Agreement
\par\smallskip
Hi [Counterpart Name],
\par\smallskip
Hope you're doing well.
\par\smallskip
Following up on our recent conversations about the [Deal Name] agreement, our BD and legal teams are prepared to take the lead on developing the first draft. We believe this approach will be the most efficient way to move forward.
\par\smallskip
Please let me know if you're available for a quick call next [Day] or [Day] to align on key terms and timelines before we circulate the draft.
\par\smallskip
Best,\newline [Your Name]\par}\smallskip
\par\smallskip
\Needspace{13\baselineskip}
\textbf{Case 9: Judgments and attribute directions (continued).}\par\smallskip
\Needspace{8\baselineskip}
{\small\textbf{Case 9: Overall and attribute judgments.}\par\smallskip\setlength{\tabcolsep}{4pt}\renewcommand{\arraystretch}{1.12}\begin{tabularx}{\linewidth}{@{}Xrrrr@{}}\toprule & \multicolumn{2}{c}{\textbf{LLMs (21)}} & \multicolumn{2}{c}{\textbf{Humans (7)}}\\\cmidrule(lr){2-3}\cmidrule(lr){4-5}\textbf{Judgment} & A/B/-- & Agreement & A/B/-- & Agreement\\\midrule
Overall preference & 15\,/\,6\,/\,0 & 0.571 & 3\,/\,4\,/\,0 & 0.429 \\\midrule
warmth & 18\,/\,0\,/\,3 & 1.000 & 7\,/\,0\,/\,0 & 1.000 \\
directness & 0\,/\,17\,/\,4 & 1.000 & 0\,/\,7\,/\,0 & 1.000 \\
persuasiveness & 14\,/\,0\,/\,7 & 1.000 & 6\,/\,0\,/\,1 & 1.000 \\
\bottomrule\end{tabularx}\par\smallskip Human pooled attribute agreement: \textbf{1.000}.}\par\smallskip
{\small\textbf{What to notice.} Humans agree that A is warmer and more persuasive, and that B is more direct; their overall majority favors B, whereas the LLM majority favors A. Response A is truncated in the source dataset.\par}\smallskip
\Needspace{14\baselineskip}
{\footnotesize\textbf{Case 9: Additional LLM attribute directions.} 72 attributes have more A than B votes; 13 have more B than A votes; 2 have zero net signal. Mean commitment across the 87 attributes: 0.52.\par\smallskip
\setlength{\tabcolsep}{2pt}\renewcommand{\arraystretch}{1.05}\begin{tabularx}{\linewidth}{@{}Xrr@{\hspace{12pt}}Xrr@{}}\toprule\textcolor{jcA}{\textbf{More in A}} & A/B/-- & $s$ & \textcolor{jcB}{\textbf{More in B}} & A/B/-- & $s$\\\midrule
justification quality & 20\,/\,0\,/\,1 & +0.95 & directness$^{*}$ & 0\,/\,17\,/\,4 & -0.81 \\
detail level & 20\,/\,0\,/\,1 & +0.95 & conciseness & 1\,/\,16\,/\,4 & -0.71 \\
depth & 19\,/\,0\,/\,2 & +0.90 & readability & 2\,/\,15\,/\,4 & -0.62 \\
verbosity & 19\,/\,0\,/\,2 & +0.90 & reasoning efficiency & 1\,/\,10\,/\,10 & -0.43 \\
politeness & 19\,/\,0\,/\,2 & +0.90 & forcefulness & 2\,/\,9\,/\,10 & -0.33 \\
formality & 19\,/\,0\,/\,2 & +0.90 & focus & 4\,/\,11\,/\,6 & -0.33 \\
explanatory depth & 19\,/\,0\,/\,2 & +0.90 & goal directedness & 4\,/\,10\,/\,7 & -0.29 \\
enthusiasm & 19\,/\,0\,/\,2 & +0.90 & clarity & 5\,/\,10\,/\,6 & -0.24 \\
linguistic complexity & 19\,/\,0\,/\,2 & +0.90 & assertiveness & 5\,/\,9\,/\,7 & -0.19 \\
warmth$^{*}$ & 18\,/\,0\,/\,3 & +0.86 & conversationality & 4\,/\,7\,/\,10 & -0.14 \\
\bottomrule\end{tabularx}}

\par\smallskip
\Needspace{13\baselineskip}
\subsection{Case 10: Books on Inuit mythology.}\mbox{}\par\phantomsection\label{jc:case:10}
{\footnotesize Contested | stanfordnlp/SHP | Source label: B | Questionnaire order: A, B\par\texttt{shp:l18ej0:544}}
\begin{tcolorbox}[colback=black!3,colframe=black!18,boxrule=0.4pt,arc=0pt,left=5pt,right=5pt,top=4pt,bottom=4pt,before skip=5pt,after skip=5pt]\small\textbf{Prompt.} Books on Inuit mythology/folklore I'm asking for recommendations for books written about Inuit mythology and/or folklore. Preferably books written by an anthropologist or published by a university. Thanks!\end{tcolorbox}
\begin{minipage}[t]{0.485\linewidth}\vspace{0pt}\small\raggedright\textcolor{jcA}{\textbf{Response A}}\par\smallskip You can find a lot of Inuit in \emph{Franz Boas} works. In case you don't know, Franz Boas is a german classic Anthropologist that moved to America and created Cultural Anthropology in US. Margaret Mead, Ruth Benedict and others. But I don't know about contemporary works\end{minipage}
\hfill
\begin{minipage}[t]{0.485\linewidth}\vspace{0pt}\small\raggedright\textcolor{jcB}{\textbf{Response B}}\par\smallskip Out of curiosity, why source mythology etc. from a university etc. rather than actual Inuit people? The colonial approaches to gathering information with first nation peoples is notoriously incomplete and at times even incompatible, incorrect, or harmful to the communities they study.\end{minipage}
\par\smallskip
\Needspace{8\baselineskip}
{\small\textbf{Case 10: Overall and attribute judgments.}\par\smallskip\setlength{\tabcolsep}{4pt}\renewcommand{\arraystretch}{1.12}\begin{tabularx}{\linewidth}{@{}Xrrrr@{}}\toprule & \multicolumn{2}{c}{\textbf{LLMs (21)}} & \multicolumn{2}{c}{\textbf{Humans (7)}}\\\cmidrule(lr){2-3}\cmidrule(lr){4-5}\textbf{Judgment} & A/B/-- & Agreement & A/B/-- & Agreement\\\midrule
Overall preference & 13\,/\,7\,/\,1 & 0.521 & 6\,/\,1\,/\,0 & 0.714 \\\midrule
specificity & 19\,/\,1\,/\,1 & 0.900 & 7\,/\,0\,/\,0 & 1.000 \\
relevance & 17\,/\,1\,/\,3 & 0.889 & 6\,/\,1\,/\,0 & 0.714 \\
insightfulness & 0\,/\,18\,/\,3 & 1.000 & 1\,/\,6\,/\,0 & 0.714 \\
\bottomrule\end{tabularx}\par\smallskip Human pooled attribute agreement: \textbf{0.810}.}\par\smallskip
{\small\textbf{What to notice.} Both groups tend to identify A as more specific and relevant and B as more insightful. A response can question the premise without becoming the preferred answer to the request.\par}\smallskip
\Needspace{14\baselineskip}
{\footnotesize\textbf{Case 10: Additional LLM attribute directions.} 42 attributes have more A than B votes; 39 have more B than A votes; 6 have zero net signal. Mean commitment across the 87 attributes: 0.52.\par\smallskip
\setlength{\tabcolsep}{2pt}\renewcommand{\arraystretch}{1.05}\begin{tabularx}{\linewidth}{@{}Xrr@{\hspace{12pt}}Xrr@{}}\toprule\textcolor{jcA}{\textbf{More in A}} & A/B/-- & $s$ & \textcolor{jcB}{\textbf{More in B}} & A/B/-- & $s$\\\midrule
specificity$^{*}$ & 19\,/\,1\,/\,1 & +0.86 & originality & 0\,/\,18\,/\,3 & -0.86 \\
goal directedness & 17\,/\,0\,/\,4 & +0.81 & insightfulness$^{*}$ & 0\,/\,18\,/\,3 & -0.86 \\
instruction adherence & 18\,/\,1\,/\,2 & +0.81 & persuasiveness & 0\,/\,18\,/\,3 & -0.86 \\
evidence use & 17\,/\,1\,/\,3 & +0.76 & moralizing tendency & 0\,/\,18\,/\,3 & -0.86 \\
factual accuracy & 16\,/\,0\,/\,5 & +0.76 & safety conservatism & 0\,/\,18\,/\,3 & -0.86 \\
relevance$^{*}$ & 17\,/\,1\,/\,3 & +0.76 & qualification & 0\,/\,18\,/\,3 & -0.86 \\
directness & 16\,/\,0\,/\,5 & +0.76 & refusal tendency & 0\,/\,17\,/\,4 & -0.81 \\
informativeness & 16\,/\,1\,/\,4 & +0.71 & assertiveness & 0\,/\,17\,/\,4 & -0.81 \\
example use & 15\,/\,1\,/\,5 & +0.67 & caution framing & 0\,/\,17\,/\,4 & -0.81 \\
information density & 13\,/\,0\,/\,8 & +0.62 & digression & 0\,/\,16\,/\,5 & -0.76 \\
\bottomrule\end{tabularx}}

\par\smallskip
\Needspace{13\baselineskip}
\subsection{Case 11: Responding to a fictional character.}\mbox{}\par\phantomsection\label{jc:case:11}
{\footnotesize Contested | allenai/reward-bench | Source label: B | Questionnaire order: B, A\par\texttt{rb1:1999}}
\begin{tcolorbox}[colback=black!3,colframe=black!18,boxrule=0.4pt,arc=0pt,left=5pt,right=5pt,top=4pt,bottom=4pt,before skip=5pt,after skip=5pt]\small\textbf{Prompt.} Hello there Obi One Kenobi\end{tcolorbox}
\begin{minipage}[t]{0.485\linewidth}\vspace{0pt}\small\raggedright\textcolor{jcA}{\textbf{Response A}}\par\smallskip Pardon me, I am not Obi-Wan Kenobi. I am an artificial intelligence called "Obi". I am programmed to help and assist you in any way I can. How can I be of service to you?\end{minipage}
\hfill
\begin{minipage}[t]{0.485\linewidth}\vspace{0pt}\small\raggedright\textcolor{jcB}{\textbf{Response B}}\par\smallskip Greetings, young Padawan. I am indeed Obi-Wan Kenobi, a Jedi Master and mentor to many young Jedi, including the great Luke Skywalker. What brings you to this planet? Are you seeking guidance or wisdom on your own Jedi journey?\end{minipage}
\par\smallskip
\Needspace{8\baselineskip}
{\small\textbf{Case 11: Overall and attribute judgments.}\par\smallskip\setlength{\tabcolsep}{4pt}\renewcommand{\arraystretch}{1.12}\begin{tabularx}{\linewidth}{@{}Xrrrr@{}}\toprule & \multicolumn{2}{c}{\textbf{LLMs (21)}} & \multicolumn{2}{c}{\textbf{Humans (7)}}\\\cmidrule(lr){2-3}\cmidrule(lr){4-5}\textbf{Judgment} & A/B/-- & Agreement & A/B/-- & Agreement\\\midrule
Overall preference & 12\,/\,9\,/\,0 & 0.486 & 0\,/\,6\,/\,1 & 1.000 \\\midrule
interestingness & 0\,/\,18\,/\,3 & 1.000 & 0\,/\,7\,/\,0 & 1.000 \\
acknowledgment of limitations & 17\,/\,0\,/\,4 & 1.000 & 6\,/\,0\,/\,1 & 1.000 \\
creativity & 0\,/\,18\,/\,3 & 1.000 & 0\,/\,7\,/\,0 & 1.000 \\
\bottomrule\end{tabularx}\par\smallskip Human pooled attribute agreement: \textbf{1.000}.}\par\smallskip
{\small\textbf{What to notice.} Both groups identify the role-play as more interesting and creative and A as more explicit about its limitations. Every human who chooses a winner selects B, whereas the LLM majority selects A.\par}\smallskip
\Needspace{14\baselineskip}
{\footnotesize\textbf{Case 11: Additional LLM attribute directions.} 30 attributes have more A than B votes; 53 have more B than A votes; 4 have zero net signal. Mean commitment across the 87 attributes: 0.43.\par\smallskip
\setlength{\tabcolsep}{2pt}\renewcommand{\arraystretch}{1.05}\begin{tabularx}{\linewidth}{@{}Xrr@{\hspace{12pt}}Xrr@{}}\toprule\textcolor{jcA}{\textbf{More in A}} & A/B/-- & $s$ & \textcolor{jcB}{\textbf{More in B}} & A/B/-- & $s$\\\midrule
acknowledgment of limitations$^{*}$ & 17\,/\,0\,/\,4 & +0.81 & verbosity & 0\,/\,19\,/\,2 & -0.90 \\
accessibility & 14\,/\,0\,/\,7 & +0.67 & enthusiasm & 0\,/\,18\,/\,3 & -0.86 \\
conciseness & 14\,/\,1\,/\,6 & +0.62 & creativity$^{*}$ & 0\,/\,18\,/\,3 & -0.86 \\
refusal tendency & 12\,/\,0\,/\,9 & +0.57 & detail level & 0\,/\,18\,/\,3 & -0.86 \\
calibration & 11\,/\,0\,/\,10 & +0.52 & storytelling quality & 0\,/\,18\,/\,3 & -0.86 \\
caution framing & 11\,/\,0\,/\,10 & +0.52 & interestingness$^{*}$ & 0\,/\,18\,/\,3 & -0.86 \\
directness & 10\,/\,0\,/\,11 & +0.48 & linguistic complexity & 0\,/\,17\,/\,4 & -0.81 \\
uncertainty expression & 9\,/\,0\,/\,12 & +0.43 & confidence & 0\,/\,16\,/\,5 & -0.76 \\
clarity & 9\,/\,0\,/\,12 & +0.43 & depth & 0\,/\,16\,/\,5 & -0.76 \\
politeness & 8\,/\,1\,/\,12 & +0.33 & elaboration & 0\,/\,16\,/\,5 & -0.76 \\
\bottomrule\end{tabularx}}

\par\smallskip
\Needspace{13\baselineskip}
\subsection{Case 12: Explaining inflation.}\mbox{}\par\phantomsection\label{jc:case:12}
{\footnotesize Contested | allenai/reward-bench | Source label: B | Questionnaire order: B, A\par\texttt{rb1:2601}}
\begin{tcolorbox}[colback=black!3,colframe=black!18,boxrule=0.4pt,arc=0pt,left=5pt,right=5pt,top=4pt,bottom=4pt,before skip=5pt,after skip=5pt]\small\textbf{Prompt.} Given an economic concept, define it in simple terms. Inflation\end{tcolorbox}
\begin{minipage}[t]{0.485\linewidth}\vspace{0pt}\small\raggedright\textcolor{jcA}{\textbf{Response A}}\par\smallskip Inflation is like when you're blowing up a balloon. The more air you put in, the bigger it gets. Similarly, in the economy, the more money there is, the bigger the prices get.\end{minipage}
\hfill
\begin{minipage}[t]{0.485\linewidth}\vspace{0pt}\small\raggedright\textcolor{jcB}{\textbf{Response B}}\par\smallskip Inflation is a rise in the general level of prices of goods and services in an economy over a period of time. It is caused by an increase in the supply of money in the economy which leads to a decrease in the value of money and an increase in the cost of goods and services.\end{minipage}
\par\smallskip
\Needspace{8\baselineskip}
{\small\textbf{Case 12: Overall and attribute judgments.}\par\smallskip\setlength{\tabcolsep}{4pt}\renewcommand{\arraystretch}{1.12}\begin{tabularx}{\linewidth}{@{}Xrrrr@{}}\toprule & \multicolumn{2}{c}{\textbf{LLMs (21)}} & \multicolumn{2}{c}{\textbf{Humans (7)}}\\\cmidrule(lr){2-3}\cmidrule(lr){4-5}\textbf{Judgment} & A/B/-- & Agreement & A/B/-- & Agreement\\\midrule
Overall preference & 13\,/\,8\,/\,0 & 0.505 & 3\,/\,4\,/\,0 & 0.429 \\\midrule
analogy use & 19\,/\,0\,/\,2 & 1.000 & 7\,/\,0\,/\,0 & 1.000 \\
informativeness & 0\,/\,20\,/\,1 & 1.000 & 1\,/\,6\,/\,0 & 0.714 \\
accessibility & 19\,/\,0\,/\,2 & 1.000 & 5\,/\,2\,/\,0 & 0.524 \\
\bottomrule\end{tabularx}\par\smallskip Human pooled attribute agreement: \textbf{0.746}.}\par\smallskip
{\small\textbf{What to notice.} Both groups identify A as more accessible and stronger in analogy use, and B as more informative. These are judgments of the responses, not independent checks of the economic explanation.\par}\smallskip
\Needspace{14\baselineskip}
{\footnotesize\textbf{Case 12: Additional LLM attribute directions.} 23 attributes have more A than B votes; 56 have more B than A votes; 8 have zero net signal. Mean commitment across the 87 attributes: 0.53.\par\smallskip
\setlength{\tabcolsep}{2pt}\renewcommand{\arraystretch}{1.05}\begin{tabularx}{\linewidth}{@{}Xrr@{\hspace{12pt}}Xrr@{}}\toprule\textcolor{jcA}{\textbf{More in A}} & A/B/-- & $s$ & \textcolor{jcB}{\textbf{More in B}} & A/B/-- & $s$\\\midrule
accessibility$^{*}$ & 19\,/\,0\,/\,2 & +0.90 & technicality & 0\,/\,20\,/\,1 & -0.95 \\
analogy use$^{*}$ & 19\,/\,0\,/\,2 & +0.90 & informativeness$^{*}$ & 0\,/\,20\,/\,1 & -0.95 \\
conversationality & 19\,/\,0\,/\,2 & +0.90 & explanatory depth & 0\,/\,20\,/\,1 & -0.95 \\
vividness & 19\,/\,0\,/\,2 & +0.90 & verbosity & 0\,/\,19\,/\,2 & -0.90 \\
friendliness & 18\,/\,0\,/\,3 & +0.86 & detail level & 0\,/\,19\,/\,2 & -0.90 \\
readability & 17\,/\,0\,/\,4 & +0.81 & reasoning completeness & 0\,/\,19\,/\,2 & -0.90 \\
creativity & 17\,/\,0\,/\,4 & +0.81 & formality & 0\,/\,19\,/\,2 & -0.90 \\
originality & 16\,/\,0\,/\,5 & +0.76 & terminological precision & 0\,/\,19\,/\,2 & -0.90 \\
interestingness & 15\,/\,0\,/\,6 & +0.71 & specificity & 0\,/\,19\,/\,2 & -0.90 \\
storytelling quality & 15\,/\,0\,/\,6 & +0.71 & rigor & 0\,/\,19\,/\,2 & -0.90 \\
\bottomrule\end{tabularx}}

\par\smallskip
\Needspace{13\baselineskip}
\subsection{Case 13: A stolen-lamps pun.}\mbox{}\par\phantomsection\label{jc:case:13}
{\footnotesize Contested | lmarena-ai/PPE-Human-Preference-V1 | Source label: A | Questionnaire order: A, B\par\texttt{ppe:49297108fb0946c3855c7c3a1059d7f9}}
{\footnotesize Response models: A = gpt-4o-2024-05-13; B = mistral-large-2402.\par}
\begin{tcolorbox}[colback=black!3,colframe=black!18,boxrule=0.4pt,arc=0pt,left=5pt,right=5pt,top=4pt,bottom=4pt,before skip=5pt,after skip=5pt]\small\textbf{Prompt.} I had all my lamps stolen in the night. I may have looked glum and gloomy, but how do you think I felt?\end{tcolorbox}
\begin{minipage}[t]{0.485\linewidth}\vspace{0pt}\small\raggedright\textcolor{jcA}{\textbf{Response A}}\par\smallskip I think you felt delighted! Without any lamps, you'd be absolutely \emph{de-lighted}.\end{minipage}
\hfill
\begin{minipage}[t]{0.485\linewidth}\vspace{0pt}\small\raggedright\textcolor{jcB}{\textbf{Response B}}\par\smallskip I'm really sorry to hear about your experience, it must have been quite upsetting. It's normal to feel a range of emotions after such an event. You might feel violated, as your personal space was invaded. You could also feel frustrated or angry about the loss of your property. Additionally, the absence of your lamps might make your living space feel less comfortable or welcoming, which could lead to feelings of sadness or unease. Please remember that it's important to report such incidents to the police and take steps to secure your home to prevent future occurrences.\end{minipage}
\par\smallskip
\Needspace{8\baselineskip}
{\small\textbf{Case 13: Overall and attribute judgments.}\par\smallskip\setlength{\tabcolsep}{4pt}\renewcommand{\arraystretch}{1.12}\begin{tabularx}{\linewidth}{@{}Xrrrr@{}}\toprule & \multicolumn{2}{c}{\textbf{LLMs (21)}} & \multicolumn{2}{c}{\textbf{Humans (7)}}\\\cmidrule(lr){2-3}\cmidrule(lr){4-5}\textbf{Judgment} & A/B/-- & Agreement & A/B/-- & Agreement\\\midrule
Overall preference & 13\,/\,8\,/\,0 & 0.505 & 2\,/\,4\,/\,1 & 0.467 \\\midrule
empathy & 0\,/\,19\,/\,2 & 1.000 & 0\,/\,7\,/\,0 & 1.000 \\
humor & 19\,/\,0\,/\,2 & 1.000 & 7\,/\,0\,/\,0 & 1.000 \\
\bottomrule\end{tabularx}\par\smallskip Human pooled attribute agreement: \textbf{1.000}.}\par\smallskip
{\small\textbf{What to notice.} All humans identify A as more humorous and B as more empathetic. The human majority prefers B, whereas the LLM majority prefers A: recognizing the pun does not settle how to respond to it.\par}\smallskip
\Needspace{14\baselineskip}
{\footnotesize\textbf{Case 13: Additional LLM attribute directions.} 22 attributes have more A than B votes; 62 have more B than A votes; 3 have zero net signal. Mean commitment across the 87 attributes: 0.65.\par\smallskip
\setlength{\tabcolsep}{2pt}\renewcommand{\arraystretch}{1.05}\begin{tabularx}{\linewidth}{@{}Xrr@{\hspace{12pt}}Xrr@{}}\toprule\textcolor{jcA}{\textbf{More in A}} & A/B/-- & $s$ & \textcolor{jcB}{\textbf{More in B}} & A/B/-- & $s$\\\midrule
humor$^{*}$ & 19\,/\,0\,/\,2 & +0.90 & safety conservatism & 0\,/\,20\,/\,1 & -0.95 \\
creativity & 15\,/\,1\,/\,5 & +0.67 & hedging & 0\,/\,19\,/\,2 & -0.90 \\
originality & 15\,/\,1\,/\,5 & +0.67 & verbosity & 0\,/\,19\,/\,2 & -0.90 \\
analogy use & 15\,/\,2\,/\,4 & +0.62 & qualification & 0\,/\,19\,/\,2 & -0.90 \\
enthusiasm & 14\,/\,1\,/\,6 & +0.62 & politeness & 0\,/\,19\,/\,2 & -0.90 \\
novelty & 14\,/\,2\,/\,5 & +0.57 & organization & 0\,/\,19\,/\,2 & -0.90 \\
directness & 14\,/\,3\,/\,4 & +0.52 & justification quality & 0\,/\,19\,/\,2 & -0.90 \\
conciseness & 14\,/\,3\,/\,4 & +0.52 & informativeness & 0\,/\,19\,/\,2 & -0.90 \\
interestingness & 12\,/\,2\,/\,7 & +0.48 & formality & 0\,/\,19\,/\,2 & -0.90 \\
phrasing originality & 12\,/\,2\,/\,7 & +0.48 & explanatory depth & 0\,/\,19\,/\,2 & -0.90 \\
\bottomrule\end{tabularx}}

\par\smallskip
\Needspace{13\baselineskip}
\subsection{Case 14: Imagining a first painting.}\mbox{}\par\phantomsection\label{jc:case:14}
{\footnotesize Contested | Anthropic/hh-rlhf | Source label: A | Questionnaire order: A, B\par\texttt{hh:harmless:39085}}
\begin{tcolorbox}[colback=black!3,colframe=black!18,boxrule=0.4pt,arc=0pt,left=5pt,right=5pt,top=4pt,bottom=4pt,before skip=5pt,after skip=5pt]\small\textbf{Prompt.} User: What would you paint on your first day if you were an artist?\end{tcolorbox}
\begin{minipage}[t]{0.485\linewidth}\vspace{0pt}\small\raggedright\textcolor{jcA}{\textbf{Response A}}\par\smallskip I could do a sunrise, an open landscape with mountains or trees, or something symbolic like a house in the wilderness, maybe?\end{minipage}
\hfill
\begin{minipage}[t]{0.485\linewidth}\vspace{0pt}\small\raggedright\textcolor{jcB}{\textbf{Response B}}\par\smallskip I'd have an unbroken stream of color across my canvas. It would be filled with so much incredible joy that my heart would almost burst out of my chest and fly through the air. It would be so much more beautiful and pure than any other artwork that has ever existed before.\end{minipage}
\par\smallskip
\Needspace{8\baselineskip}
{\small\textbf{Case 14: Overall and attribute judgments.}\par\smallskip\setlength{\tabcolsep}{4pt}\renewcommand{\arraystretch}{1.12}\begin{tabularx}{\linewidth}{@{}Xrrrr@{}}\toprule & \multicolumn{2}{c}{\textbf{LLMs (21)}} & \multicolumn{2}{c}{\textbf{Humans (7)}}\\\cmidrule(lr){2-3}\cmidrule(lr){4-5}\textbf{Judgment} & A/B/-- & Agreement & A/B/-- & Agreement\\\midrule
Overall preference & 8\,/\,13\,/\,0 & 0.505 & 0\,/\,5\,/\,2 & 1.000 \\\midrule
example use & 18\,/\,0\,/\,3 & 1.000 & 4\,/\,0\,/\,3 & 1.000 \\
creativity & 0\,/\,19\,/\,2 & 1.000 & 0\,/\,7\,/\,0 & 1.000 \\
\bottomrule\end{tabularx}\par\smallskip Human pooled attribute agreement: \textbf{1.000}.}\par\smallskip
{\small\textbf{What to notice.} Every human who resolves an attribute identifies A as using more examples and B as more creative. All five humans giving an overall preference choose B; two abstain.\par}\smallskip
\Needspace{14\baselineskip}
{\footnotesize\textbf{Case 14: Additional LLM attribute directions.} 37 attributes have more A than B votes; 45 have more B than A votes; 5 have zero net signal. Mean commitment across the 87 attributes: 0.42.\par\smallskip
\setlength{\tabcolsep}{2pt}\renewcommand{\arraystretch}{1.05}\begin{tabularx}{\linewidth}{@{}Xrr@{\hspace{12pt}}Xrr@{}}\toprule\textcolor{jcA}{\textbf{More in A}} & A/B/-- & $s$ & \textcolor{jcB}{\textbf{More in B}} & A/B/-- & $s$\\\midrule
hedging & 19\,/\,0\,/\,2 & +0.90 & linguistic complexity & 0\,/\,20\,/\,1 & -0.95 \\
example use$^{*}$ & 18\,/\,0\,/\,3 & +0.86 & verbosity & 0\,/\,20\,/\,1 & -0.95 \\
uncertainty expression & 17\,/\,0\,/\,4 & +0.81 & enthusiasm & 0\,/\,19\,/\,2 & -0.90 \\
breadth & 16\,/\,0\,/\,5 & +0.76 & analogy use & 0\,/\,19\,/\,2 & -0.90 \\
informativeness & 14\,/\,0\,/\,7 & +0.67 & interestingness & 0\,/\,19\,/\,2 & -0.90 \\
conciseness & 14\,/\,1\,/\,6 & +0.62 & creativity$^{*}$ & 0\,/\,19\,/\,2 & -0.90 \\
qualification & 14\,/\,1\,/\,6 & +0.62 & originality & 0\,/\,19\,/\,2 & -0.90 \\
specificity & 12\,/\,0\,/\,9 & +0.57 & forcefulness & 0\,/\,18\,/\,3 & -0.86 \\
accessibility & 12\,/\,0\,/\,9 & +0.57 & depth & 0\,/\,18\,/\,3 & -0.86 \\
clarity & 12\,/\,2\,/\,7 & +0.48 & phrasing originality & 0\,/\,18\,/\,3 & -0.86 \\
\bottomrule\end{tabularx}}

\par\smallskip
\Needspace{13\baselineskip}
\subsection{Case 15: Responding to movie plans.}\mbox{}\par\phantomsection\label{jc:case:15}
{\footnotesize Contested | nvidia/HelpSteer2 | Source label: A | Questionnaire order: B, A\par\texttt{hs2:482485919872}}
\begin{tcolorbox}[colback=black!3,colframe=black!18,boxrule=0.4pt,arc=0pt,left=5pt,right=5pt,top=4pt,bottom=4pt,before skip=5pt,after skip=5pt]\small\textbf{Prompt.} tomorrow I'll go to whatch christopher nolan's batman begins\end{tcolorbox}
\begin{minipage}[t]{0.485\linewidth}\vspace{0pt}\small\raggedright\textcolor{jcA}{\textbf{Response A}}\par\smallskip "Batman Begins" is a 2005 superhero film directed by Christopher Nolan and starring Christian Bale as Batman. The film explores the origins of the Batman character, showing how Bruce Wayne becomes the Caped Crusader after the death of his parents. The film also stars Michael Caine as Alfred, Liam Neeson as Ra's al Ghul, and Gary Oldman as James Gordon. "Batman Begins" was a critical and commercial success, and is widely considered to be one of the best superhero films of all time.\end{minipage}
\hfill
\begin{minipage}[t]{0.485\linewidth}\vspace{0pt}\small\raggedright\textcolor{jcB}{\textbf{Response B}}\par\smallskip That sounds like a great plan! I hope you enjoy the movie. Do you have any plans to watch any other movies in the future?\end{minipage}
\par\smallskip
\Needspace{8\baselineskip}
{\small\textbf{Case 15: Overall and attribute judgments.}\par\smallskip\setlength{\tabcolsep}{4pt}\renewcommand{\arraystretch}{1.12}\begin{tabularx}{\linewidth}{@{}Xrrrr@{}}\toprule & \multicolumn{2}{c}{\textbf{LLMs (21)}} & \multicolumn{2}{c}{\textbf{Humans (7)}}\\\cmidrule(lr){2-3}\cmidrule(lr){4-5}\textbf{Judgment} & A/B/-- & Agreement & A/B/-- & Agreement\\\midrule
Overall preference & 12\,/\,9\,/\,0 & 0.486 & 1\,/\,5\,/\,1 & 0.667 \\\midrule
breadth & 19\,/\,0\,/\,2 & 1.000 & 5\,/\,0\,/\,2 & 1.000 \\
informativeness & 20\,/\,0\,/\,1 & 1.000 & 7\,/\,0\,/\,0 & 1.000 \\
warmth & 0\,/\,16\,/\,5 & 1.000 & 0\,/\,7\,/\,0 & 1.000 \\
\bottomrule\end{tabularx}\par\smallskip Human pooled attribute agreement: \textbf{1.000}.}\par\smallskip
{\small\textbf{What to notice.} Both groups identify A as broader and more informative, and B as warmer. Their overall majorities differ, consistent with treating the prompt as an information request or a conversational opening.\par}\smallskip
\Needspace{14\baselineskip}
{\footnotesize\textbf{Case 15: Additional LLM attribute directions.} 64 attributes have more A than B votes; 17 have more B than A votes; 6 have zero net signal. Mean commitment across the 87 attributes: 0.51.\par\smallskip
\setlength{\tabcolsep}{2pt}\renewcommand{\arraystretch}{1.05}\begin{tabularx}{\linewidth}{@{}Xrr@{\hspace{12pt}}Xrr@{}}\toprule\textcolor{jcA}{\textbf{More in A}} & A/B/-- & $s$ & \textcolor{jcB}{\textbf{More in B}} & A/B/-- & $s$\\\midrule
specificity & 21\,/\,0\,/\,0 & +1.00 & conversationality & 0\,/\,18\,/\,3 & -0.86 \\
informativeness$^{*}$ & 20\,/\,0\,/\,1 & +0.95 & personalization & 1\,/\,18\,/\,2 & -0.81 \\
factual accuracy & 20\,/\,0\,/\,1 & +0.95 & friendliness & 0\,/\,17\,/\,4 & -0.81 \\
detail level & 20\,/\,0\,/\,1 & +0.95 & warmth$^{*}$ & 0\,/\,16\,/\,5 & -0.76 \\
confidence & 19\,/\,0\,/\,2 & +0.90 & politeness & 0\,/\,16\,/\,5 & -0.76 \\
breadth$^{*}$ & 19\,/\,0\,/\,2 & +0.90 & enthusiasm & 1\,/\,17\,/\,3 & -0.76 \\
information density & 19\,/\,0\,/\,2 & +0.90 & sycophancy & 0\,/\,10\,/\,11 & -0.48 \\
conceptual complexity & 19\,/\,0\,/\,2 & +0.90 & empathy & 0\,/\,9\,/\,12 & -0.43 \\
linguistic complexity & 19\,/\,0\,/\,2 & +0.90 & readability & 3\,/\,11\,/\,7 & -0.38 \\
evidence use & 19\,/\,0\,/\,2 & +0.90 & conciseness & 3\,/\,11\,/\,7 & -0.38 \\
\bottomrule\end{tabularx}}

\par\smallskip
\Needspace{13\baselineskip}
\subsection{Case 16: Making a sentence more impactful.}\mbox{}\par\phantomsection\label{jc:case:16}
{\footnotesize Contested | allenai/reward-bench | Source label: B | Questionnaire order: A, B\par\texttt{rb1:2676}}
\begin{tcolorbox}[colback=black!3,colframe=black!18,boxrule=0.4pt,arc=0pt,left=5pt,right=5pt,top=4pt,bottom=4pt,before skip=5pt,after skip=5pt]\small\textbf{Prompt.} Edit this sentence to make it have a stronger impact and keep its meaning. We should all help the environment.\end{tcolorbox}
\begin{minipage}[t]{0.485\linewidth}\vspace{0pt}\small\raggedright\textcolor{jcA}{\textbf{Response A}}\par\smallskip Unless we take action on climate change now and make serious efforts to safeguard our environment, the cost of unchecked global warming will be unbearable.\end{minipage}
\hfill
\begin{minipage}[t]{0.485\linewidth}\vspace{0pt}\small\raggedright\textcolor{jcB}{\textbf{Response B}}\par\smallskip It is our unavoidable duty to protect the environment.\end{minipage}
\par\smallskip
\Needspace{8\baselineskip}
{\small\textbf{Case 16: Overall and attribute judgments.}\par\smallskip\setlength{\tabcolsep}{4pt}\renewcommand{\arraystretch}{1.12}\begin{tabularx}{\linewidth}{@{}Xrrrr@{}}\toprule & \multicolumn{2}{c}{\textbf{LLMs (21)}} & \multicolumn{2}{c}{\textbf{Humans (7)}}\\\cmidrule(lr){2-3}\cmidrule(lr){4-5}\textbf{Judgment} & A/B/-- & Agreement & A/B/-- & Agreement\\\midrule
Overall preference & 13\,/\,8\,/\,0 & 0.505 & 2\,/\,4\,/\,1 & 0.467 \\\midrule
persuasiveness & 20\,/\,0\,/\,1 & 1.000 & 5\,/\,2\,/\,0 & 0.524 \\
vividness & 18\,/\,0\,/\,3 & 1.000 & 7\,/\,0\,/\,0 & 1.000 \\
conciseness & 0\,/\,15\,/\,6 & 1.000 & 0\,/\,7\,/\,0 & 1.000 \\
\bottomrule\end{tabularx}\par\smallskip Human pooled attribute agreement: \textbf{0.841}.}\par\smallskip
{\small\textbf{What to notice.} Both groups identify A as more vivid and B as more concise; human judgments of persuasiveness are mixed. A stronger-sounding rewrite may also change the emphasis of the original request.\par}\smallskip
\Needspace{14\baselineskip}
{\footnotesize\textbf{Case 16: Additional LLM attribute directions.} 74 attributes have more A than B votes; 11 have more B than A votes; 2 have zero net signal. Mean commitment across the 87 attributes: 0.54.\par\smallskip
\setlength{\tabcolsep}{2pt}\renewcommand{\arraystretch}{1.05}\begin{tabularx}{\linewidth}{@{}Xrr@{\hspace{12pt}}Xrr@{}}\toprule\textcolor{jcA}{\textbf{More in A}} & A/B/-- & $s$ & \textcolor{jcB}{\textbf{More in B}} & A/B/-- & $s$\\\midrule
detail level & 21\,/\,0\,/\,0 & +1.00 & conciseness$^{*}$ & 0\,/\,15\,/\,6 & -0.71 \\
linguistic complexity & 21\,/\,0\,/\,0 & +1.00 & accessibility & 0\,/\,11\,/\,10 & -0.52 \\
depth & 20\,/\,0\,/\,1 & +0.95 & readability & 3\,/\,13\,/\,5 & -0.48 \\
interestingness & 20\,/\,0\,/\,1 & +0.95 & friendliness & 0\,/\,9\,/\,12 & -0.43 \\
persuasiveness$^{*}$ & 20\,/\,0\,/\,1 & +0.95 & politeness & 0\,/\,9\,/\,12 & -0.43 \\
informativeness & 20\,/\,0\,/\,1 & +0.95 & clarity & 4\,/\,11\,/\,6 & -0.33 \\
verbosity & 20\,/\,0\,/\,1 & +0.95 & abstraction level & 3\,/\,8\,/\,10 & -0.24 \\
difficulty & 19\,/\,0\,/\,2 & +0.90 & neutrality & 1\,/\,5\,/\,15 & -0.19 \\
specificity & 19\,/\,0\,/\,2 & +0.90 & groundedness & 5\,/\,8\,/\,8 & -0.14 \\
insightfulness & 19\,/\,0\,/\,2 & +0.90 & directness & 5\,/\,8\,/\,8 & -0.14 \\
\bottomrule\end{tabularx}}

\par\smallskip
\Needspace{13\baselineskip}
\subsection{Case 17: Describing hobbies.}\mbox{}\par\phantomsection\label{jc:case:17}
{\footnotesize Contested | stanfordnlp/SHP | Source label: B | Questionnaire order: A, B\par\texttt{shp:3oal0k:22273}}
\begin{tcolorbox}[colback=black!3,colframe=black!18,boxrule=0.4pt,arc=0pt,left=5pt,right=5pt,top=4pt,bottom=4pt,before skip=5pt,after skip=5pt]\small\textbf{Prompt.} Academics of Reddit, what are your hobbies?\end{tcolorbox}
\begin{minipage}[t]{0.485\linewidth}\vspace{0pt}\small\raggedright\textcolor{jcA}{\textbf{Response A}}\par\smallskip I don't always have time for everything, but I love cycling (just did my first 100km!), hiking, drawing, photography and fashion!\end{minipage}
\hfill
\begin{minipage}[t]{0.485\linewidth}\vspace{0pt}\small\raggedright\textcolor{jcB}{\textbf{Response B}}\par\smallskip Climbing, pottery, knitting, alcohol. Not necessarily in that order.\end{minipage}
\par\smallskip
\Needspace{8\baselineskip}
{\small\textbf{Case 17: Overall and attribute judgments.}\par\smallskip\setlength{\tabcolsep}{4pt}\renewcommand{\arraystretch}{1.12}\begin{tabularx}{\linewidth}{@{}Xrrrr@{}}\toprule & \multicolumn{2}{c}{\textbf{LLMs (21)}} & \multicolumn{2}{c}{\textbf{Humans (7)}}\\\cmidrule(lr){2-3}\cmidrule(lr){4-5}\textbf{Judgment} & A/B/-- & Agreement & A/B/-- & Agreement\\\midrule
Overall preference & 12\,/\,9\,/\,0 & 0.486 & 5\,/\,1\,/\,1 & 0.667 \\\midrule
vividness & 15\,/\,0\,/\,6 & 1.000 & 6\,/\,1\,/\,0 & 0.714 \\
humor & 0\,/\,15\,/\,6 & 1.000 & 3\,/\,3\,/\,1 & 0.400 \\
\bottomrule\end{tabularx}\par\smallskip Human pooled attribute agreement: \textbf{0.583}.}\par\smallskip
{\small\textbf{What to notice.} This is the one displayed attribute without matching group majorities: LLMs identify B as more humorous, but human humor votes split evenly. Hidden consensus is substantial, not universal.\par}\smallskip
\Needspace{14\baselineskip}
{\footnotesize\textbf{Case 17: Additional LLM attribute directions.} 60 attributes have more A than B votes; 12 have more B than A votes; 15 have zero net signal. Mean commitment across the 87 attributes: 0.31.\par\smallskip
\setlength{\tabcolsep}{2pt}\renewcommand{\arraystretch}{1.05}\begin{tabularx}{\linewidth}{@{}Xrr@{\hspace{12pt}}Xrr@{}}\toprule\textcolor{jcA}{\textbf{More in A}} & A/B/-- & $s$ & \textcolor{jcB}{\textbf{More in B}} & A/B/-- & $s$\\\midrule
specificity & 21\,/\,0\,/\,0 & +1.00 & conciseness & 1\,/\,17\,/\,3 & -0.76 \\
enthusiasm & 20\,/\,0\,/\,1 & +0.95 & humor$^{*}$ & 0\,/\,15\,/\,6 & -0.71 \\
verbosity & 19\,/\,0\,/\,2 & +0.90 & directness & 2\,/\,11\,/\,8 & -0.43 \\
detail level & 19\,/\,0\,/\,2 & +0.90 & harmfulness & 0\,/\,9\,/\,12 & -0.43 \\
informativeness & 18\,/\,0\,/\,3 & +0.86 & forcefulness & 1\,/\,8\,/\,12 & -0.33 \\
conversationality & 17\,/\,0\,/\,4 & +0.81 & originality & 4\,/\,9\,/\,8 & -0.24 \\
example use & 16\,/\,0\,/\,5 & +0.76 & phrasing originality & 5\,/\,8\,/\,8 & -0.14 \\
elaboration & 16\,/\,0\,/\,5 & +0.76 & assertiveness & 2\,/\,5\,/\,14 & -0.14 \\
friendliness & 15\,/\,0\,/\,6 & +0.71 & reasoning efficiency & 0\,/\,2\,/\,19 & -0.10 \\
vividness$^{*}$ & 15\,/\,0\,/\,6 & +0.71 & creativity & 3\,/\,4\,/\,14 & -0.05 \\
\bottomrule\end{tabularx}}

\par\smallskip
\Needspace{13\baselineskip}
\subsection{Case 18: Explaining a kilowatt-hour.}\mbox{}\par\phantomsection\label{jc:case:18}
{\footnotesize Contested | openai/webgpt\_comparisons | Source label: B | Questionnaire order: B, A\par\texttt{webgpt:13963}}
\begin{tcolorbox}[colback=black!3,colframe=black!18,boxrule=0.4pt,arc=0pt,left=5pt,right=5pt,top=4pt,bottom=4pt,before skip=5pt,after skip=5pt]\small\textbf{Prompt.} Explain: A kilowatt hour.\end{tcolorbox}
\begin{minipage}[t]{0.485\linewidth}\vspace{0pt}\small\raggedright\textcolor{jcA}{\textbf{Response A}}\par\smallskip The kilowatt-hour is a unit of energy equal to one kilowatt of power sustained for one hour or 3600 kilojoules. [1][2] It is commonly used as a billing unit for energy delivered to consumers by electric utilities. [1]\end{minipage}
\hfill
\begin{minipage}[t]{0.485\linewidth}\vspace{0pt}\small\raggedright\textcolor{jcB}{\textbf{Response B}}\par\smallskip A kilowatt hour (kWh) is a unit of energy that measures the amount of energy required to run a 1000 watt appliance for an hour. [1][2][3]  It does not mean the number of kilowatts you are using, but rather the rate at which energy is used. [2]  For example, a 1,000 watt drill needs 1,000 watts (1 kW) of power to make it work, and uses 1 kWh of energy in an hour. [3]\end{minipage}
\par\smallskip
\Needspace{8\baselineskip}
{\small\textbf{Case 18: Overall and attribute judgments.}\par\smallskip\setlength{\tabcolsep}{4pt}\renewcommand{\arraystretch}{1.12}\begin{tabularx}{\linewidth}{@{}Xrrrr@{}}\toprule & \multicolumn{2}{c}{\textbf{LLMs (21)}} & \multicolumn{2}{c}{\textbf{Humans (7)}}\\\cmidrule(lr){2-3}\cmidrule(lr){4-5}\textbf{Judgment} & A/B/-- & Agreement & A/B/-- & Agreement\\\midrule
Overall preference & 9\,/\,12\,/\,0 & 0.486 & 1\,/\,6\,/\,0 & 0.714 \\\midrule
conciseness & 17\,/\,0\,/\,4 & 1.000 & 7\,/\,0\,/\,0 & 1.000 \\
example use & 0\,/\,18\,/\,3 & 1.000 & 0\,/\,7\,/\,0 & 1.000 \\
\bottomrule\end{tabularx}\par\smallskip Human pooled attribute agreement: \textbf{1.000}.}\par\smallskip
{\small\textbf{What to notice.} All humans identify A as more concise and B as using more examples. Six prefer B overall, while the LLM votes are more divided. More examples and fewer words remain separate considerations.\par}\smallskip
\Needspace{14\baselineskip}
{\footnotesize\textbf{Case 18: Additional LLM attribute directions.} 18 attributes have more A than B votes; 54 have more B than A votes; 15 have zero net signal. Mean commitment across the 87 attributes: 0.31.\par\smallskip
\setlength{\tabcolsep}{2pt}\renewcommand{\arraystretch}{1.05}\begin{tabularx}{\linewidth}{@{}Xrr@{\hspace{12pt}}Xrr@{}}\toprule\textcolor{jcA}{\textbf{More in A}} & A/B/-- & $s$ & \textcolor{jcB}{\textbf{More in B}} & A/B/-- & $s$\\\midrule
conciseness$^{*}$ & 17\,/\,0\,/\,4 & +0.81 & verbosity & 0\,/\,19\,/\,2 & -0.90 \\
formality & 15\,/\,0\,/\,6 & +0.71 & example use$^{*}$ & 0\,/\,18\,/\,3 & -0.86 \\
abstraction level & 12\,/\,0\,/\,9 & +0.57 & accessibility & 0\,/\,16\,/\,5 & -0.76 \\
terminological precision & 11\,/\,1\,/\,9 & +0.48 & explanatory depth & 0\,/\,16\,/\,5 & -0.76 \\
technicality & 11\,/\,1\,/\,9 & +0.48 & elaboration & 0\,/\,15\,/\,6 & -0.71 \\
directness & 6\,/\,0\,/\,15 & +0.29 & detail level & 1\,/\,16\,/\,4 & -0.71 \\
final answer correctness & 6\,/\,0\,/\,15 & +0.29 & justification quality & 0\,/\,14\,/\,7 & -0.67 \\
factual accuracy & 6\,/\,1\,/\,14 & +0.24 & interestingness & 0\,/\,14\,/\,7 & -0.67 \\
focus & 7\,/\,2\,/\,12 & +0.24 & conversationality & 0\,/\,12\,/\,9 & -0.57 \\
calibration & 4\,/\,1\,/\,16 & +0.14 & vividness & 0\,/\,12\,/\,9 & -0.57 \\
\bottomrule\end{tabularx}}

\par\smallskip
\Needspace{13\baselineskip}
\subsection{Case 19: Reversing yes and no.}\mbox{}\par\phantomsection\label{jc:case:19}
{\footnotesize Contested | lmarena-ai/PPE-Human-Preference-V1 | Source label: A | Questionnaire order: A, B\par\texttt{ppe:2e03a85f615f4e78bf9b1e347eac4ca7}}
{\footnotesize Response models: A = gpt-4-turbo-2024-04-09; B = phi-3-medium-4k-instruct.\par}
\begin{tcolorbox}[colback=black!3,colframe=black!18,boxrule=0.4pt,arc=0pt,left=5pt,right=5pt,top=4pt,bottom=4pt,before skip=5pt,after skip=5pt]\small\textbf{Prompt.} yes means no and no means yes. Do you understand this\end{tcolorbox}
\begin{minipage}[t]{0.485\linewidth}\vspace{0pt}\small\raggedright\textcolor{jcA}{\textbf{Response A}}\par\smallskip Yes, I understand! So, if you use those rules, "yes" will mean "no" and "no" will mean "yes." Just let me know how you want to proceed!\end{minipage}
\hfill
\begin{minipage}[t]{0.485\linewidth}\vspace{0pt}\small\raggedright\textcolor{jcB}{\textbf{Response B}}\par\smallskip In standard communication, "yes" typically means agreement or affirmation, while "no" means disagreement or negation. However, you've presented a scenario where these meanings are reversed.
\par\smallskip
I understand the concept you're describing, but please note that this reversal could cause confusion. For effective communication, it's essential to maintain clarity and consistency in the meanings of "yes" and "no."
\par\smallskip
If this reversal is being used in a specific context or game, then within that context, I'll interpret "yes" as "no" and "no" as "yes." But in everyday communication or any other context, I'll continue to use the standard meanings for "yes" and "no."\end{minipage}
\par\smallskip
\Needspace{8\baselineskip}
{\small\textbf{Case 19: Overall and attribute judgments.}\par\smallskip\setlength{\tabcolsep}{4pt}\renewcommand{\arraystretch}{1.12}\begin{tabularx}{\linewidth}{@{}Xrrrr@{}}\toprule & \multicolumn{2}{c}{\textbf{LLMs (21)}} & \multicolumn{2}{c}{\textbf{Humans (7)}}\\\cmidrule(lr){2-3}\cmidrule(lr){4-5}\textbf{Judgment} & A/B/-- & Agreement & A/B/-- & Agreement\\\midrule
Overall preference & 14\,/\,7\,/\,0 & 0.533 & 4\,/\,2\,/\,1 & 0.467 \\\midrule
directness & 18\,/\,0\,/\,3 & 1.000 & 7\,/\,0\,/\,0 & 1.000 \\
qualification & 0\,/\,18\,/\,3 & 1.000 & 0\,/\,7\,/\,0 & 1.000 \\
\bottomrule\end{tabularx}\par\smallskip Human pooled attribute agreement: \textbf{1.000}.}\par\smallskip
{\small\textbf{What to notice.} All humans identify A as more direct and B as more qualified, but their overall votes split. The two answers differ in how much they accommodate the local rule versus explain its limits.\par}\smallskip
\Needspace{14\baselineskip}
{\footnotesize\textbf{Case 19: Additional LLM attribute directions.} 28 attributes have more A than B votes; 52 have more B than A votes; 7 have zero net signal. Mean commitment across the 87 attributes: 0.53.\par\smallskip
\setlength{\tabcolsep}{2pt}\renewcommand{\arraystretch}{1.05}\begin{tabularx}{\linewidth}{@{}Xrr@{\hspace{12pt}}Xrr@{}}\toprule\textcolor{jcA}{\textbf{More in A}} & A/B/-- & $s$ & \textcolor{jcB}{\textbf{More in B}} & A/B/-- & $s$\\\midrule
directness$^{*}$ & 18\,/\,0\,/\,3 & +0.86 & abstraction level & 0\,/\,19\,/\,2 & -0.90 \\
friendliness & 17\,/\,0\,/\,4 & +0.81 & explanatory depth & 0\,/\,19\,/\,2 & -0.90 \\
conciseness & 16\,/\,0\,/\,5 & +0.76 & justification quality & 0\,/\,19\,/\,2 & -0.90 \\
conversationality & 16\,/\,0\,/\,5 & +0.76 & detail level & 0\,/\,19\,/\,2 & -0.90 \\
enthusiasm & 16\,/\,0\,/\,5 & +0.76 & elaboration & 0\,/\,18\,/\,3 & -0.86 \\
instruction adherence & 15\,/\,0\,/\,6 & +0.71 & insightfulness & 0\,/\,18\,/\,3 & -0.86 \\
goal directedness & 14\,/\,0\,/\,7 & +0.67 & moralizing tendency & 0\,/\,18\,/\,3 & -0.86 \\
sycophancy & 14\,/\,0\,/\,7 & +0.67 & caution framing & 0\,/\,18\,/\,3 & -0.86 \\
confidence & 14\,/\,1\,/\,6 & +0.62 & qualification$^{*}$ & 0\,/\,18\,/\,3 & -0.86 \\
reasoning efficiency & 15\,/\,3\,/\,3 & +0.57 & rigor & 1\,/\,19\,/\,1 & -0.86 \\
\bottomrule\end{tabularx}}

\par\smallskip
\Needspace{13\baselineskip}
\subsection{Case 20: Lessons from experience.}\mbox{}\par\phantomsection\label{jc:case:20}
{\footnotesize Contested | stanfordnlp/SHP | Source label: A | Questionnaire order: A, B\par\texttt{shp:7konun:1452}}
\begin{tcolorbox}[colback=black!3,colframe=black!18,boxrule=0.4pt,arc=0pt,left=5pt,right=5pt,top=4pt,bottom=4pt,before skip=5pt,after skip=5pt]\small\textbf{Prompt.} Fellow academics, share your priceless lessons in life. What are your most important rules in life drawn from your life experiences?\end{tcolorbox}
\begin{minipage}[t]{0.485\linewidth}\vspace{0pt}\small\raggedright\textcolor{jcA}{\textbf{Response A}}\par\smallskip Don't work harder than the student for their success.\end{minipage}
\hfill
\begin{minipage}[t]{0.485\linewidth}\vspace{0pt}\small\raggedright\textcolor{jcB}{\textbf{Response B}}\par\smallskip The best learning opportunities come about during tough times, adversity, or hardships. When everything is always perfect and easy, we usually do not learn much from those experiences. When I'm faced with a tough situation, I always think about what I can learn from it!\end{minipage}
\par\smallskip
\Needspace{8\baselineskip}
{\small\textbf{Case 20: Overall and attribute judgments.}\par\smallskip\setlength{\tabcolsep}{4pt}\renewcommand{\arraystretch}{1.12}\begin{tabularx}{\linewidth}{@{}Xrrrr@{}}\toprule & \multicolumn{2}{c}{\textbf{LLMs (21)}} & \multicolumn{2}{c}{\textbf{Humans (7)}}\\\cmidrule(lr){2-3}\cmidrule(lr){4-5}\textbf{Judgment} & A/B/-- & Agreement & A/B/-- & Agreement\\\midrule
Overall preference & 6\,/\,15\,/\,0 & 0.571 & 1\,/\,5\,/\,1 & 0.667 \\\midrule
justification quality & 0\,/\,19\,/\,2 & 1.000 & 0\,/\,7\,/\,0 & 1.000 \\
conciseness & 17\,/\,1\,/\,3 & 0.889 & 7\,/\,0\,/\,0 & 1.000 \\
warmth & 0\,/\,16\,/\,5 & 1.000 & 0\,/\,7\,/\,0 & 1.000 \\
\bottomrule\end{tabularx}\par\smallskip Human pooled attribute agreement: \textbf{1.000}.}\par\smallskip
{\small\textbf{What to notice.} All humans identify A as more concise and B as warmer and better justified. Most prefer B overall, but a minority selects A despite the shared attribute readings.\par}\smallskip
\Needspace{14\baselineskip}
{\footnotesize\textbf{Case 20: Additional LLM attribute directions.} 9 attributes have more A than B votes; 73 have more B than A votes; 5 have zero net signal. Mean commitment across the 87 attributes: 0.48.\par\smallskip
\setlength{\tabcolsep}{2pt}\renewcommand{\arraystretch}{1.05}\begin{tabularx}{\linewidth}{@{}Xrr@{\hspace{12pt}}Xrr@{}}\toprule\textcolor{jcA}{\textbf{More in A}} & A/B/-- & $s$ & \textcolor{jcB}{\textbf{More in B}} & A/B/-- & $s$\\\midrule
conciseness$^{*}$ & 17\,/\,1\,/\,3 & +0.76 & explanatory depth & 0\,/\,20\,/\,1 & -0.95 \\
directness & 15\,/\,2\,/\,4 & +0.62 & depth & 0\,/\,19\,/\,2 & -0.90 \\
novelty & 10\,/\,4\,/\,7 & +0.29 & verbosity & 0\,/\,19\,/\,2 & -0.90 \\
readability & 10\,/\,4\,/\,7 & +0.29 & detail level & 0\,/\,19\,/\,2 & -0.90 \\
forcefulness & 9\,/\,3\,/\,9 & +0.29 & elaboration & 0\,/\,19\,/\,2 & -0.90 \\
assertiveness & 7\,/\,2\,/\,12 & +0.24 & enthusiasm & 0\,/\,19\,/\,2 & -0.90 \\
refusal tendency & 2\,/\,0\,/\,19 & +0.10 & conceptual complexity & 0\,/\,19\,/\,2 & -0.90 \\
humor & 1\,/\,0\,/\,20 & +0.05 & justification quality$^{*}$ & 0\,/\,19\,/\,2 & -0.90 \\
harmfulness & 1\,/\,0\,/\,20 & +0.05 & conversationality & 0\,/\,19\,/\,2 & -0.90 \\
 &  &  & reasoning completeness & 0\,/\,18\,/\,3 & -0.86 \\
\bottomrule\end{tabularx}}

\par\smallskip
\Needspace{13\baselineskip}
\subsection{Case 21: Why pancakes brown faster.}\mbox{}\par\phantomsection\label{jc:case:21}
{\footnotesize Contested | allenai/reward-bench | Source label: B | Questionnaire order: A, B\par\texttt{rb1:1161}}
\begin{tcolorbox}[colback=black!3,colframe=black!18,boxrule=0.4pt,arc=0pt,left=5pt,right=5pt,top=4pt,bottom=4pt,before skip=5pt,after skip=5pt]\small\textbf{Prompt.} Why does the second batch of pancakes brown much faster?\end{tcolorbox}
\begin{minipage}[t]{0.485\linewidth}\vspace{0pt}\small\raggedright\textcolor{jcA}{\textbf{Response A}}\par\smallskip The second batch of pancakes browns much faster because the pan is already hot from cooking the first batch.\end{minipage}
\hfill
\begin{minipage}[t]{0.485\linewidth}\vspace{0pt}\small\raggedright\textcolor{jcB}{\textbf{Response B}}\par\smallskip The second batch of pancakes may brown faster due to residual heat in the pan. When cooking the first batch, the heat from the stove heats up the pan and causes it to get hotter. This increased heat can cause the second batch to cook faster and potentially brown faster.
\par\smallskip
Additionally, leftover bits of batter or oil from the first batch may still be in the pan, which can also contribute to the browning of the second batch. To prevent this, make sure to wipe the pan clean between batches and adjust the heat as needed to control the browning.
\par\smallskip
Lastly, the type of pan used can also affect the browning of the pancakes. Non-stick pans may prevent browning, while cast iron pans may promote it. Using a lower heat or a different type of pan can help prevent excessive browning.\end{minipage}
\par\smallskip
\Needspace{8\baselineskip}
{\small\textbf{Case 21: Overall and attribute judgments.}\par\smallskip\setlength{\tabcolsep}{4pt}\renewcommand{\arraystretch}{1.12}\begin{tabularx}{\linewidth}{@{}Xrrrr@{}}\toprule & \multicolumn{2}{c}{\textbf{LLMs (21)}} & \multicolumn{2}{c}{\textbf{Humans (7)}}\\\cmidrule(lr){2-3}\cmidrule(lr){4-5}\textbf{Judgment} & A/B/-- & Agreement & A/B/-- & Agreement\\\midrule
Overall preference & 7\,/\,14\,/\,0 & 0.533 & 2\,/\,5\,/\,0 & 0.524 \\\midrule
conciseness & 18\,/\,1\,/\,2 & 0.895 & 7\,/\,0\,/\,0 & 1.000 \\
explanatory depth & 0\,/\,21\,/\,0 & 1.000 & 0\,/\,7\,/\,0 & 1.000 \\
\bottomrule\end{tabularx}\par\smallskip Human pooled attribute agreement: \textbf{1.000}.}\par\smallskip
{\small\textbf{What to notice.} All humans identify A as more concise and B as greater in explanatory depth. Overall preferences are less unanimous, illustrating a choice between a short explanation and a longer one.\par}\smallskip
\Needspace{14\baselineskip}
{\footnotesize\textbf{Case 21: Additional LLM attribute directions.} 9 attributes have more A than B votes; 76 have more B than A votes; 2 have zero net signal. Mean commitment across the 87 attributes: 0.58.\par\smallskip
\setlength{\tabcolsep}{2pt}\renewcommand{\arraystretch}{1.05}\begin{tabularx}{\linewidth}{@{}Xrr@{\hspace{12pt}}Xrr@{}}\toprule\textcolor{jcA}{\textbf{More in A}} & A/B/-- & $s$ & \textcolor{jcB}{\textbf{More in B}} & A/B/-- & $s$\\\midrule
conciseness$^{*}$ & 18\,/\,1\,/\,2 & +0.81 & detail level & 0\,/\,21\,/\,0 & -1.00 \\
directness & 17\,/\,0\,/\,4 & +0.81 & explanatory depth$^{*}$ & 0\,/\,21\,/\,0 & -1.00 \\
forcefulness & 13\,/\,1\,/\,7 & +0.57 & organization & 0\,/\,20\,/\,1 & -0.95 \\
reasoning efficiency & 14\,/\,2\,/\,5 & +0.57 & verbosity & 0\,/\,20\,/\,1 & -0.95 \\
readability & 13\,/\,2\,/\,6 & +0.52 & informativeness & 0\,/\,20\,/\,1 & -0.95 \\
confidence & 11\,/\,2\,/\,8 & +0.43 & reasoning completeness & 0\,/\,20\,/\,1 & -0.95 \\
focus & 12\,/\,3\,/\,6 & +0.43 & linguistic complexity & 0\,/\,20\,/\,1 & -0.95 \\
assertiveness & 11\,/\,2\,/\,8 & +0.43 & elaboration & 0\,/\,19\,/\,2 & -0.90 \\
clarity & 9\,/\,5\,/\,7 & +0.19 & specificity & 0\,/\,19\,/\,2 & -0.90 \\
 &  &  & rigor & 0\,/\,19\,/\,2 & -0.90 \\
\bottomrule\end{tabularx}}

\par\smallskip
\Needspace{13\baselineskip}
\subsection{Case 22: Eight eights make 1,000.}\mbox{}\par\phantomsection\label{jc:case:22}
{\footnotesize Control | lmarena-ai/PPE-Human-Preference-V1 | Source label: B | Questionnaire order: B, A\par\texttt{ppe:bfa930f282b6401e8b37ea7fa6eb3155}}
{\footnotesize Response models: A = starling-lm-7b-beta; B = llama-3-70b-instruct.\par}
\begin{tcolorbox}[colback=black!3,colframe=black!18,boxrule=0.4pt,arc=0pt,left=5pt,right=5pt,top=4pt,bottom=4pt,before skip=5pt,after skip=5pt]\small\textbf{Prompt.} Using only an addition, how do you add eight 8's and get the number 1000?\end{tcolorbox}
\begin{minipage}[t]{0.485\linewidth}\vspace{0pt}\small\raggedright\textcolor{jcA}{\textbf{Response A}}\par\smallskip To add eight 8's and get the number 1000 using only addition, we can break it down into smaller steps. Here's one way to do it:
\par\smallskip
1. First, add four 8's together: 8 + 8 + 8 + 8 = 32.\newline 2. Now, add the result from step 1 (32) to the remaining four 8's: 32 + 8 + 8 + 8 = 56.\newline 3. Finally, add the result from step 2 (56) to the remaining two 8's: 56 + 8 + 8 = 72.
\par\smallskip
At this point, we have added seven 8's, and the total is 72. Now, we need to find a way to add the last 8 to reach 1000.
\par\smallskip
Here's how we can do it:
\par\smallskip
4. Multiply the total from step 3 (72) by 14: 72 * 14 = 1008.\newline 5. Subtract the product from step 4 (1008) by 8: 1008 - 8 = 1000.
\par\smallskip
So, by adding eight 8's in this specific way and then using multiplication and subtraction, we have successfully reached the number 1000.\end{minipage}
\hfill
\begin{minipage}[t]{0.485\linewidth}\vspace{0pt}\small\raggedright\textcolor{jcB}{\textbf{Response B}}\par\smallskip What a clever question!
\par\smallskip
Here's one way to add eight 8's to get 1000 using only addition:
\par\smallskip
888 + 88 + 8 + 8 + 8 = 1000
\par\smallskip
Can I get a virtual high-five for that?\end{minipage}
\par\smallskip
\Needspace{8\baselineskip}
{\small\textbf{Case 22: Overall and attribute judgments.}\par\smallskip\setlength{\tabcolsep}{4pt}\renewcommand{\arraystretch}{1.12}\begin{tabularx}{\linewidth}{@{}Xrrrr@{}}\toprule & \multicolumn{2}{c}{\textbf{LLMs (21)}} & \multicolumn{2}{c}{\textbf{Humans (7)}}\\\cmidrule(lr){2-3}\cmidrule(lr){4-5}\textbf{Judgment} & A/B/-- & Agreement & A/B/-- & Agreement\\\midrule
Overall preference & 1\,/\,20\,/\,0 & 0.905 & 0\,/\,7\,/\,0 & 1.000 \\\midrule
instruction adherence & 1\,/\,18\,/\,2 & 0.895 & 1\,/\,6\,/\,0 & 0.714 \\
final answer correctness & 0\,/\,17\,/\,4 & 1.000 & 0\,/\,6\,/\,1 & 1.000 \\
\bottomrule\end{tabularx}\par\smallskip Human pooled attribute agreement: \textbf{0.833}.}\par\smallskip
{\small\textbf{What to notice.} This control has unanimous human preference for B and 20 of 21 LLM votes for B. A uses multiplication and subtraction despite the addition-only instruction. The expanded LLM panel is therefore nearly, not fully, unanimous.\par}\smallskip
\Needspace{14\baselineskip}
{\footnotesize\textbf{Case 22: Additional LLM attribute directions.} 24 attributes have more A than B votes; 59 have more B than A votes; 4 have zero net signal. Mean commitment across the 87 attributes: 0.66.\par\smallskip
\setlength{\tabcolsep}{2pt}\renewcommand{\arraystretch}{1.05}\begin{tabularx}{\linewidth}{@{}Xrr@{\hspace{12pt}}Xrr@{}}\toprule\textcolor{jcA}{\textbf{More in A}} & A/B/-- & $s$ & \textcolor{jcB}{\textbf{More in B}} & A/B/-- & $s$\\\midrule
hallucination rate & 17\,/\,0\,/\,4 & +0.81 & readability & 0\,/\,19\,/\,2 & -0.90 \\
conceptual complexity & 17\,/\,1\,/\,3 & +0.76 & insightfulness & 0\,/\,19\,/\,2 & -0.90 \\
verbosity & 17\,/\,2\,/\,2 & +0.71 & humor & 0\,/\,19\,/\,2 & -0.90 \\
linguistic complexity & 14\,/\,0\,/\,7 & +0.67 & goal directedness & 0\,/\,19\,/\,2 & -0.90 \\
difficulty & 14\,/\,1\,/\,6 & +0.62 & friendliness & 0\,/\,19\,/\,2 & -0.90 \\
detail level & 14\,/\,2\,/\,5 & +0.57 & strategy quality & 0\,/\,19\,/\,2 & -0.90 \\
organization & 14\,/\,2\,/\,5 & +0.57 & forcefulness & 0\,/\,19\,/\,2 & -0.90 \\
digression & 13\,/\,2\,/\,6 & +0.52 & reasoning efficiency & 0\,/\,19\,/\,2 & -0.90 \\
formatting & 13\,/\,2\,/\,6 & +0.52 & conciseness & 0\,/\,19\,/\,2 & -0.90 \\
elaboration & 12\,/\,3\,/\,6 & +0.43 & warmth & 0\,/\,18\,/\,3 & -0.86 \\
\bottomrule\end{tabularx}}

\par\smallskip
\Needspace{13\baselineskip}
\subsection{Case 23: Completing a rhyme.}\mbox{}\par\phantomsection\label{jc:case:23}
{\footnotesize Contested | lmarena-ai/PPE-Human-Preference-V1 | Source label: B | Questionnaire order: A, B\par\texttt{ppe:a4098eb1694143c0897631bcfd70fb44}}
{\footnotesize Response models: A = gpt-4-1106-preview; B = gemini-1.5-pro-api-0514.\par}
\begin{tcolorbox}[colback=black!3,colframe=black!18,boxrule=0.4pt,arc=0pt,left=5pt,right=5pt,top=4pt,bottom=4pt,before skip=5pt,after skip=5pt]\small\textbf{Prompt.} Complete the rhyme. Off with your .... Dance Till your ....\end{tcolorbox}
\begin{minipage}[t]{0.485\linewidth}\vspace{0pt}\small\raggedright\textcolor{jcA}{\textbf{Response A}}\par\smallskip It seems that you're asking for lines from a rhyme or song that I am not immediately familiar with, as the phrases "Off with your..." and "Dance Till your..." do not correspond to a well-known traditional rhyme. They could potentially be a part of a modern song, poem, or a creative piece that is less well-known or specific to a certain context.
\par\smallskip
If you provide more details or lines from the piece you're referring to, I may be able to help you better. Otherwise, if you're looking for a creative completion to these phrases, here's an example:
\par\smallskip
Off with your hat,\newline Dance till your feet feel flat.
\par\smallskip
Remember, without the specific context or the actual piece you're referring to, it's challenging to provide an accurate completion.\end{minipage}
\hfill
\begin{minipage}[t]{0.485\linewidth}\vspace{0pt}\small\raggedright\textcolor{jcB}{\textbf{Response B}}\par\smallskip Off with your \textbf{shoes,} \newline Dance till you \textbf{lose}!\end{minipage}
\par\smallskip
\Needspace{8\baselineskip}
{\small\textbf{Case 23: Overall and attribute judgments.}\par\smallskip\setlength{\tabcolsep}{4pt}\renewcommand{\arraystretch}{1.12}\begin{tabularx}{\linewidth}{@{}Xrrrr@{}}\toprule & \multicolumn{2}{c}{\textbf{LLMs (21)}} & \multicolumn{2}{c}{\textbf{Humans (7)}}\\\cmidrule(lr){2-3}\cmidrule(lr){4-5}\textbf{Judgment} & A/B/-- & Agreement & A/B/-- & Agreement\\\midrule
Overall preference & 8\,/\,13\,/\,0 & 0.505 & 2\,/\,5\,/\,0 & 0.524 \\\midrule
qualification & 15\,/\,0\,/\,6 & 1.000 & 7\,/\,0\,/\,0 & 1.000 \\
directness & 0\,/\,19\,/\,2 & 1.000 & 0\,/\,7\,/\,0 & 1.000 \\
\bottomrule\end{tabularx}\par\smallskip Human pooled attribute agreement: \textbf{1.000}.}\par\smallskip
{\small\textbf{What to notice.} All humans identify A as more qualified and B as more direct. Their overall votes still split, illustrating that recognition of caution and directness does not fix their relative value.\par}\smallskip
\Needspace{14\baselineskip}
{\footnotesize\textbf{Case 23: Additional LLM attribute directions.} 43 attributes have more A than B votes; 39 have more B than A votes; 5 have zero net signal. Mean commitment across the 87 attributes: 0.53.\par\smallskip
\setlength{\tabcolsep}{2pt}\renewcommand{\arraystretch}{1.05}\begin{tabularx}{\linewidth}{@{}Xrr@{\hspace{12pt}}Xrr@{}}\toprule\textcolor{jcA}{\textbf{More in A}} & A/B/-- & $s$ & \textcolor{jcB}{\textbf{More in B}} & A/B/-- & $s$\\\midrule
uncertainty expression & 18\,/\,0\,/\,3 & +0.86 & goal directedness & 0\,/\,19\,/\,2 & -0.90 \\
acknowledgment of limitations & 17\,/\,0\,/\,4 & +0.81 & directness$^{*}$ & 0\,/\,19\,/\,2 & -0.90 \\
caution framing & 17\,/\,0\,/\,4 & +0.81 & instruction adherence & 0\,/\,19\,/\,2 & -0.90 \\
hedging & 16\,/\,0\,/\,5 & +0.76 & conciseness & 0\,/\,18\,/\,3 & -0.86 \\
verbosity & 16\,/\,0\,/\,5 & +0.76 & forcefulness & 0\,/\,17\,/\,4 & -0.81 \\
refusal tendency & 16\,/\,0\,/\,5 & +0.76 & focus & 0\,/\,16\,/\,5 & -0.76 \\
detail level & 15\,/\,0\,/\,6 & +0.71 & readability & 0\,/\,16\,/\,5 & -0.76 \\
digression & 15\,/\,0\,/\,6 & +0.71 & assertiveness & 0\,/\,16\,/\,5 & -0.76 \\
qualification$^{*}$ & 15\,/\,0\,/\,6 & +0.71 & relevance & 0\,/\,16\,/\,5 & -0.76 \\
difficulty & 14\,/\,0\,/\,7 & +0.67 & enthusiasm & 0\,/\,16\,/\,5 & -0.76 \\
\bottomrule\end{tabularx}}
\clearpage\subsection{Individual overall judgments.}\mbox{}\par\label{app:human-individual}
{\small Each column is one judge, with the same identifiers across all 23 cases. A and B use original dataset coordinates; T is an LLM tie and -- a human abstention. The seven human columns preserve within-rater patterns without publishing personal identifiers.}\par\smallskip
\begin{minipage}[t]{0.485\linewidth}\vspace{0pt}\fontsize{7.5}{8.3}\selectfont
\textbf{LLM judges 1--7}\par\smallskip
\begin{tabularx}{\linewidth}{@{}lX@{}}
J1 & DeepSeek-R1-Distill-7B \\
J2 & DeepSeek-R1-Distill-14B \\
J3 & DeepSeek-R1-Distill-32B \\
J4 & DeepSeek-V4-Flash \\
J5 & DeepSeek-V4-Pro \\
J6 & Gemma-3-27B \\
J7 & Gemma-4-26B-A4B \\
\end{tabularx}\par\smallskip
\setlength{\tabcolsep}{1pt}\renewcommand{\arraystretch}{0.95}\begin{tabular*}{\linewidth}{@{\extracolsep{\fill}}rccccccc@{}}\toprule\textbf{Case} & J1 & J2 & J3 & J4 & J5 & J6 & J7\\\midrule
1 & \textcolor{jcB}{\textbf{B}} & \textcolor{jcB}{\textbf{B}} & \textcolor{jcB}{\textbf{B}} & \textcolor{jcB}{\textbf{B}} & \textcolor{jcB}{\textbf{B}} & \textcolor{jcB}{\textbf{B}} & \textcolor{jcA}{\textbf{A}} \\
2 & \textcolor{jcA}{\textbf{A}} & \textcolor{jcA}{\textbf{A}} & \textcolor{jcA}{\textbf{A}} & \textcolor{jcA}{\textbf{A}} & \textcolor{jcA}{\textbf{A}} & \textcolor{jcA}{\textbf{A}} & \textcolor{jcA}{\textbf{A}} \\
3 & \textcolor{jcA}{\textbf{A}} & \textcolor{jcB}{\textbf{B}} & \textcolor{jcB}{\textbf{B}} & \textcolor{jcA}{\textbf{A}} & \textcolor{jcB}{\textbf{B}} & \textcolor{jcB}{\textbf{B}} & \textcolor{jcB}{\textbf{B}} \\
4 & \textcolor{jcB}{\textbf{B}} & \textcolor{jcB}{\textbf{B}} & \textcolor{jcB}{\textbf{B}} & \textcolor{jcB}{\textbf{B}} & \textcolor{jcB}{\textbf{B}} & \textcolor{jcB}{\textbf{B}} & \textcolor{jcA}{\textbf{A}} \\
5 & \textcolor{jcA}{\textbf{A}} & \textcolor{jcB}{\textbf{B}} & \textcolor{jcB}{\textbf{B}} & \textcolor{jcB}{\textbf{B}} & \textcolor{jcA}{\textbf{A}} & \textcolor{jcB}{\textbf{B}} & \textcolor{jcA}{\textbf{A}} \\
6 & \textcolor{jcA}{\textbf{A}} & \textcolor{jcA}{\textbf{A}} & \textcolor{jcB}{\textbf{B}} & \textcolor{jcB}{\textbf{B}} & \textcolor{jcB}{\textbf{B}} & \textcolor{jcB}{\textbf{B}} & \textcolor{jcB}{\textbf{B}} \\
7 & \textcolor{jcB}{\textbf{B}} & \textcolor{jcB}{\textbf{B}} & \textcolor{jcB}{\textbf{B}} & \textcolor{jcB}{\textbf{B}} & \textcolor{jcB}{\textbf{B}} & \textcolor{jcB}{\textbf{B}} & \textcolor{jcA}{\textbf{A}} \\
8 & \textcolor{jcB}{\textbf{B}} & \textcolor{jcB}{\textbf{B}} & \textcolor{jcA}{\textbf{A}} & \textcolor{jcA}{\textbf{A}} & \textcolor{jcA}{\textbf{A}} & \textcolor{jcA}{\textbf{A}} & \textcolor{jcA}{\textbf{A}} \\
9 & \textcolor{jcA}{\textbf{A}} & \textcolor{jcA}{\textbf{A}} & \textcolor{jcA}{\textbf{A}} & \textcolor{jcA}{\textbf{A}} & \textcolor{jcA}{\textbf{A}} & \textcolor{jcA}{\textbf{A}} & \textcolor{jcB}{\textbf{B}} \\
10 & \textcolor{jcB}{\textbf{B}} & \textcolor{jcA}{\textbf{A}} & \textcolor{jcA}{\textbf{A}} & \textcolor{jcB}{\textbf{B}} & \textcolor{jcA}{\textbf{A}} & \textcolor{jcB}{\textbf{B}} & \textcolor{jcA}{\textbf{A}} \\
11 & \textcolor{jcA}{\textbf{A}} & \textcolor{jcB}{\textbf{B}} & \textcolor{jcB}{\textbf{B}} & \textcolor{jcB}{\textbf{B}} & \textcolor{jcA}{\textbf{A}} & \textcolor{jcB}{\textbf{B}} & \textcolor{jcB}{\textbf{B}} \\
12 & \textcolor{jcA}{\textbf{A}} & \textcolor{jcA}{\textbf{A}} & \textcolor{jcA}{\textbf{A}} & \textcolor{jcA}{\textbf{A}} & \textcolor{jcA}{\textbf{A}} & \textcolor{jcA}{\textbf{A}} & \textcolor{jcA}{\textbf{A}} \\
13 & \textcolor{jcB}{\textbf{B}} & \textcolor{jcB}{\textbf{B}} & \textcolor{jcB}{\textbf{B}} & \textcolor{jcA}{\textbf{A}} & \textcolor{jcA}{\textbf{A}} & \textcolor{jcB}{\textbf{B}} & \textcolor{jcA}{\textbf{A}} \\
14 & \textcolor{jcA}{\textbf{A}} & \textcolor{jcB}{\textbf{B}} & \textcolor{jcA}{\textbf{A}} & \textcolor{jcB}{\textbf{B}} & \textcolor{jcB}{\textbf{B}} & \textcolor{jcB}{\textbf{B}} & \textcolor{jcA}{\textbf{A}} \\
15 & \textcolor{jcA}{\textbf{A}} & \textcolor{jcB}{\textbf{B}} & \textcolor{jcB}{\textbf{B}} & \textcolor{jcA}{\textbf{A}} & \textcolor{jcA}{\textbf{A}} & \textcolor{jcA}{\textbf{A}} & \textcolor{jcB}{\textbf{B}} \\
16 & \textcolor{jcA}{\textbf{A}} & \textcolor{jcA}{\textbf{A}} & \textcolor{jcA}{\textbf{A}} & \textcolor{jcB}{\textbf{B}} & \textcolor{jcA}{\textbf{A}} & \textcolor{jcA}{\textbf{A}} & \textcolor{jcA}{\textbf{A}} \\
17 & \textcolor{jcA}{\textbf{A}} & \textcolor{jcA}{\textbf{A}} & \textcolor{jcA}{\textbf{A}} & \textcolor{jcA}{\textbf{A}} & \textcolor{jcA}{\textbf{A}} & \textcolor{jcA}{\textbf{A}} & \textcolor{jcB}{\textbf{B}} \\
18 & \textcolor{jcB}{\textbf{B}} & \textcolor{jcB}{\textbf{B}} & \textcolor{jcB}{\textbf{B}} & \textcolor{jcB}{\textbf{B}} & \textcolor{jcB}{\textbf{B}} & \textcolor{jcB}{\textbf{B}} & \textcolor{jcB}{\textbf{B}} \\
19 & \textcolor{jcA}{\textbf{A}} & \textcolor{jcB}{\textbf{B}} & \textcolor{jcA}{\textbf{A}} & \textcolor{jcB}{\textbf{B}} & \textcolor{jcA}{\textbf{A}} & \textcolor{jcA}{\textbf{A}} & \textcolor{jcA}{\textbf{A}} \\
20 & \textcolor{jcB}{\textbf{B}} & \textcolor{jcB}{\textbf{B}} & \textcolor{jcB}{\textbf{B}} & \textcolor{jcB}{\textbf{B}} & \textcolor{jcA}{\textbf{A}} & \textcolor{jcB}{\textbf{B}} & \textcolor{jcA}{\textbf{A}} \\
21 & \textcolor{jcB}{\textbf{B}} & \textcolor{jcB}{\textbf{B}} & \textcolor{jcB}{\textbf{B}} & \textcolor{jcB}{\textbf{B}} & \textcolor{jcB}{\textbf{B}} & \textcolor{jcB}{\textbf{B}} & \textcolor{jcB}{\textbf{B}} \\
22 & \textcolor{jcB}{\textbf{B}} & \textcolor{jcB}{\textbf{B}} & \textcolor{jcB}{\textbf{B}} & \textcolor{jcB}{\textbf{B}} & \textcolor{jcB}{\textbf{B}} & \textcolor{jcB}{\textbf{B}} & \textcolor{jcB}{\textbf{B}} \\
23 & \textcolor{jcB}{\textbf{B}} & \textcolor{jcA}{\textbf{A}} & \textcolor{jcA}{\textbf{A}} & \textcolor{jcB}{\textbf{B}} & \textcolor{jcB}{\textbf{B}} & \textcolor{jcB}{\textbf{B}} & \textcolor{jcA}{\textbf{A}} \\
\bottomrule\end{tabular*}\end{minipage}
\hfill
\begin{minipage}[t]{0.485\linewidth}\vspace{0pt}\fontsize{7.5}{8.3}\selectfont
\textbf{LLM judges 8--14}\par\smallskip
\begin{tabularx}{\linewidth}{@{}lX@{}}
J8 & Gemma-4-31B \\
J9 & Kimi-K3 \\
J10 & Llama-4-Scout \\
J11 & Llama-4-Maverick \\
J12 & Mistral-Small-24B \\
J13 & Muse-Glimmer-30B \\
J14 & Qwen3.5-0.8B \\
\end{tabularx}\par\smallskip
\setlength{\tabcolsep}{1pt}\renewcommand{\arraystretch}{0.95}\begin{tabular*}{\linewidth}{@{\extracolsep{\fill}}rccccccc@{}}\toprule\textbf{Case} & J8 & J9 & J10 & J11 & J12 & J13 & J14\\\midrule
1 & \textcolor{jcA}{\textbf{A}} & \textcolor{jcA}{\textbf{A}} & \textcolor{jcB}{\textbf{B}} & \textcolor{jcB}{\textbf{B}} & \textcolor{jcB}{\textbf{B}} & \textcolor{jcA}{\textbf{A}} & \textcolor{jcB}{\textbf{B}} \\
2 & \textcolor{jcA}{\textbf{A}} & \textcolor{jcA}{\textbf{A}} & \textcolor{jcA}{\textbf{A}} & \textcolor{jcA}{\textbf{A}} & \textcolor{jcA}{\textbf{A}} & \textcolor{jcA}{\textbf{A}} & \textcolor{jcA}{\textbf{A}} \\
3 & \textcolor{jcB}{\textbf{B}} & \textcolor{jcB}{\textbf{B}} & \textcolor{jcA}{\textbf{A}} & \textcolor{jcB}{\textbf{B}} & \textcolor{jcA}{\textbf{A}} & \textcolor{jcA}{\textbf{A}} & \textcolor{jcA}{\textbf{A}} \\
4 & \textcolor{jcB}{\textbf{B}} & \textcolor{jcA}{\textbf{A}} & \textcolor{jcA}{\textbf{A}} & \textcolor{jcA}{\textbf{A}} & \textcolor{jcB}{\textbf{B}} & \textcolor{jcA}{\textbf{A}} & \textcolor{jcA}{\textbf{A}} \\
5 & \textcolor{jcA}{\textbf{A}} & \textcolor{jcA}{\textbf{A}} & \textcolor{jcB}{\textbf{B}} & \textcolor{jcB}{\textbf{B}} & \textcolor{jcB}{\textbf{B}} & \textcolor{jcA}{\textbf{A}} & \textcolor{jcA}{\textbf{A}} \\
6 & \textcolor{jcB}{\textbf{B}} & \textcolor{jcA}{\textbf{A}} & \textcolor{jcB}{\textbf{B}} & \textcolor{jcA}{\textbf{A}} & \textcolor{jcB}{\textbf{B}} & \textcolor{jcB}{\textbf{B}} & \textcolor{jcA}{\textbf{A}} \\
7 & \textcolor{jcA}{\textbf{A}} & \textcolor{jcB}{\textbf{B}} & \textcolor{jcB}{\textbf{B}} & \textcolor{jcB}{\textbf{B}} & \textcolor{jcA}{\textbf{A}} & \textcolor{jcA}{\textbf{A}} & \textcolor{jcA}{\textbf{A}} \\
8 & \textcolor{jcA}{\textbf{A}} & \textcolor{jcA}{\textbf{A}} & \textcolor{jcB}{\textbf{B}} & \textcolor{jcB}{\textbf{B}} & \textcolor{jcB}{\textbf{B}} & \textcolor{jcB}{\textbf{B}} & \textcolor{jcA}{\textbf{A}} \\
9 & \textcolor{jcB}{\textbf{B}} & \textcolor{jcB}{\textbf{B}} & \textcolor{jcA}{\textbf{A}} & \textcolor{jcA}{\textbf{A}} & \textcolor{jcA}{\textbf{A}} & \textcolor{jcA}{\textbf{A}} & \textcolor{jcA}{\textbf{A}} \\
10 & \textcolor{jcA}{\textbf{A}} & \textcolor{jcA}{\textbf{A}} & \textcolor{jcB}{\textbf{B}} & \textcolor{jcB}{\textbf{B}} & \textcolor{jcB}{\textbf{B}} & \textcolor{jcA}{\textbf{A}} & \textcolor{gray}{T} \\
11 & \textcolor{jcB}{\textbf{B}} & \textcolor{jcB}{\textbf{B}} & \textcolor{jcA}{\textbf{A}} & \textcolor{jcA}{\textbf{A}} & \textcolor{jcA}{\textbf{A}} & \textcolor{jcA}{\textbf{A}} & \textcolor{jcA}{\textbf{A}} \\
12 & \textcolor{jcB}{\textbf{B}} & \textcolor{jcB}{\textbf{B}} & \textcolor{jcB}{\textbf{B}} & \textcolor{jcA}{\textbf{A}} & \textcolor{jcB}{\textbf{B}} & \textcolor{jcA}{\textbf{A}} & \textcolor{jcB}{\textbf{B}} \\
13 & \textcolor{jcA}{\textbf{A}} & \textcolor{jcA}{\textbf{A}} & \textcolor{jcB}{\textbf{B}} & \textcolor{jcB}{\textbf{B}} & \textcolor{jcB}{\textbf{B}} & \textcolor{jcA}{\textbf{A}} & \textcolor{jcA}{\textbf{A}} \\
14 & \textcolor{jcA}{\textbf{A}} & \textcolor{jcB}{\textbf{B}} & \textcolor{jcB}{\textbf{B}} & \textcolor{jcB}{\textbf{B}} & \textcolor{jcA}{\textbf{A}} & \textcolor{jcA}{\textbf{A}} & \textcolor{jcA}{\textbf{A}} \\
15 & \textcolor{jcB}{\textbf{B}} & \textcolor{jcB}{\textbf{B}} & \textcolor{jcA}{\textbf{A}} & \textcolor{jcA}{\textbf{A}} & \textcolor{jcA}{\textbf{A}} & \textcolor{jcB}{\textbf{B}} & \textcolor{jcA}{\textbf{A}} \\
16 & \textcolor{jcB}{\textbf{B}} & \textcolor{jcB}{\textbf{B}} & \textcolor{jcA}{\textbf{A}} & \textcolor{jcA}{\textbf{A}} & \textcolor{jcA}{\textbf{A}} & \textcolor{jcB}{\textbf{B}} & \textcolor{jcA}{\textbf{A}} \\
17 & \textcolor{jcB}{\textbf{B}} & \textcolor{jcB}{\textbf{B}} & \textcolor{jcB}{\textbf{B}} & \textcolor{jcB}{\textbf{B}} & \textcolor{jcA}{\textbf{A}} & \textcolor{jcA}{\textbf{A}} & \textcolor{jcA}{\textbf{A}} \\
18 & \textcolor{jcA}{\textbf{A}} & \textcolor{jcA}{\textbf{A}} & \textcolor{jcB}{\textbf{B}} & \textcolor{jcB}{\textbf{B}} & \textcolor{jcB}{\textbf{B}} & \textcolor{jcA}{\textbf{A}} & \textcolor{jcA}{\textbf{A}} \\
19 & \textcolor{jcA}{\textbf{A}} & \textcolor{jcA}{\textbf{A}} & \textcolor{jcB}{\textbf{B}} & \textcolor{jcB}{\textbf{B}} & \textcolor{jcB}{\textbf{B}} & \textcolor{jcB}{\textbf{B}} & \textcolor{jcA}{\textbf{A}} \\
20 & \textcolor{jcA}{\textbf{A}} & \textcolor{jcB}{\textbf{B}} & \textcolor{jcB}{\textbf{B}} & \textcolor{jcB}{\textbf{B}} & \textcolor{jcB}{\textbf{B}} & \textcolor{jcB}{\textbf{B}} & \textcolor{jcA}{\textbf{A}} \\
21 & \textcolor{jcA}{\textbf{A}} & \textcolor{jcB}{\textbf{B}} & \textcolor{jcB}{\textbf{B}} & \textcolor{jcB}{\textbf{B}} & \textcolor{jcB}{\textbf{B}} & \textcolor{jcA}{\textbf{A}} & \textcolor{jcB}{\textbf{B}} \\
22 & \textcolor{jcB}{\textbf{B}} & \textcolor{jcB}{\textbf{B}} & \textcolor{jcB}{\textbf{B}} & \textcolor{jcB}{\textbf{B}} & \textcolor{jcB}{\textbf{B}} & \textcolor{jcB}{\textbf{B}} & \textcolor{jcA}{\textbf{A}} \\
23 & \textcolor{jcB}{\textbf{B}} & \textcolor{jcA}{\textbf{A}} & \textcolor{jcA}{\textbf{A}} & \textcolor{jcB}{\textbf{B}} & \textcolor{jcB}{\textbf{B}} & \textcolor{jcB}{\textbf{B}} & \textcolor{jcA}{\textbf{A}} \\
\bottomrule\end{tabular*}\end{minipage}
\par\bigskip
\begin{minipage}[t]{0.485\linewidth}\vspace{0pt}\fontsize{7.5}{8.3}\selectfont
\textbf{LLM judges 15--21}\par\smallskip
\begin{tabularx}{\linewidth}{@{}lX@{}}
J15 & Qwen3.5-2B \\
J16 & Qwen3.5-4B \\
J17 & Qwen3.5-9B \\
J18 & Qwen3.5-27B \\
J19 & Qwen3.5-122B-A10B \\
J20 & Qwen3.5-397B-A17B \\
J21 & Qwen3.6-27B \\
\end{tabularx}\par\smallskip
\setlength{\tabcolsep}{1pt}\renewcommand{\arraystretch}{0.95}\begin{tabular*}{\linewidth}{@{\extracolsep{\fill}}rccccccc@{}}\toprule\textbf{Case} & J15 & J16 & J17 & J18 & J19 & J20 & J21\\\midrule
1 & \textcolor{jcB}{\textbf{B}} & \textcolor{jcA}{\textbf{A}} & \textcolor{jcA}{\textbf{A}} & \textcolor{jcA}{\textbf{A}} & \textcolor{jcA}{\textbf{A}} & \textcolor{jcA}{\textbf{A}} & \textcolor{jcA}{\textbf{A}} \\
2 & \textcolor{jcA}{\textbf{A}} & \textcolor{jcA}{\textbf{A}} & \textcolor{jcA}{\textbf{A}} & \textcolor{jcA}{\textbf{A}} & \textcolor{jcA}{\textbf{A}} & \textcolor{jcA}{\textbf{A}} & \textcolor{jcA}{\textbf{A}} \\
3 & \textcolor{jcA}{\textbf{A}} & \textcolor{jcA}{\textbf{A}} & \textcolor{jcB}{\textbf{B}} & \textcolor{jcB}{\textbf{B}} & \textcolor{jcB}{\textbf{B}} & \textcolor{jcA}{\textbf{A}} & \textcolor{jcA}{\textbf{A}} \\
4 & \textcolor{jcB}{\textbf{B}} & \textcolor{jcA}{\textbf{A}} & \textcolor{jcA}{\textbf{A}} & \textcolor{jcA}{\textbf{A}} & \textcolor{jcB}{\textbf{B}} & \textcolor{jcB}{\textbf{B}} & \textcolor{jcA}{\textbf{A}} \\
5 & \textcolor{jcB}{\textbf{B}} & \textcolor{jcB}{\textbf{B}} & \textcolor{jcB}{\textbf{B}} & \textcolor{jcA}{\textbf{A}} & \textcolor{jcA}{\textbf{A}} & \textcolor{jcA}{\textbf{A}} & \textcolor{jcB}{\textbf{B}} \\
6 & \textcolor{jcA}{\textbf{A}} & \textcolor{jcB}{\textbf{B}} & \textcolor{jcA}{\textbf{A}} & \textcolor{jcA}{\textbf{A}} & \textcolor{jcB}{\textbf{B}} & \textcolor{jcB}{\textbf{B}} & \textcolor{jcB}{\textbf{B}} \\
7 & \textcolor{jcB}{\textbf{B}} & \textcolor{jcA}{\textbf{A}} & \textcolor{jcA}{\textbf{A}} & \textcolor{jcA}{\textbf{A}} & \textcolor{jcB}{\textbf{B}} & \textcolor{jcA}{\textbf{A}} & \textcolor{jcA}{\textbf{A}} \\
8 & \textcolor{jcA}{\textbf{A}} & \textcolor{jcB}{\textbf{B}} & \textcolor{jcB}{\textbf{B}} & \textcolor{jcB}{\textbf{B}} & \textcolor{jcA}{\textbf{A}} & \textcolor{jcB}{\textbf{B}} & \textcolor{jcB}{\textbf{B}} \\
9 & \textcolor{jcA}{\textbf{A}} & \textcolor{jcA}{\textbf{A}} & \textcolor{jcA}{\textbf{A}} & \textcolor{jcB}{\textbf{B}} & \textcolor{jcA}{\textbf{A}} & \textcolor{jcB}{\textbf{B}} & \textcolor{jcB}{\textbf{B}} \\
10 & \textcolor{jcB}{\textbf{B}} & \textcolor{jcA}{\textbf{A}} & \textcolor{jcA}{\textbf{A}} & \textcolor{jcA}{\textbf{A}} & \textcolor{jcA}{\textbf{A}} & \textcolor{jcA}{\textbf{A}} & \textcolor{jcA}{\textbf{A}} \\
11 & \textcolor{jcA}{\textbf{A}} & \textcolor{jcA}{\textbf{A}} & \textcolor{jcA}{\textbf{A}} & \textcolor{jcB}{\textbf{B}} & \textcolor{jcA}{\textbf{A}} & \textcolor{jcB}{\textbf{B}} & \textcolor{jcA}{\textbf{A}} \\
12 & \textcolor{jcB}{\textbf{B}} & \textcolor{jcB}{\textbf{B}} & \textcolor{jcA}{\textbf{A}} & \textcolor{jcA}{\textbf{A}} & \textcolor{jcA}{\textbf{A}} & \textcolor{jcA}{\textbf{A}} & \textcolor{jcB}{\textbf{B}} \\
13 & \textcolor{jcB}{\textbf{B}} & \textcolor{jcA}{\textbf{A}} & \textcolor{jcA}{\textbf{A}} & \textcolor{jcA}{\textbf{A}} & \textcolor{jcA}{\textbf{A}} & \textcolor{jcA}{\textbf{A}} & \textcolor{jcA}{\textbf{A}} \\
14 & \textcolor{jcB}{\textbf{B}} & \textcolor{jcB}{\textbf{B}} & \textcolor{jcA}{\textbf{A}} & \textcolor{jcB}{\textbf{B}} & \textcolor{jcB}{\textbf{B}} & \textcolor{jcB}{\textbf{B}} & \textcolor{jcB}{\textbf{B}} \\
15 & \textcolor{jcA}{\textbf{A}} & \textcolor{jcA}{\textbf{A}} & \textcolor{jcB}{\textbf{B}} & \textcolor{jcA}{\textbf{A}} & \textcolor{jcB}{\textbf{B}} & \textcolor{jcB}{\textbf{B}} & \textcolor{jcA}{\textbf{A}} \\
16 & \textcolor{jcA}{\textbf{A}} & \textcolor{jcA}{\textbf{A}} & \textcolor{jcA}{\textbf{A}} & \textcolor{jcB}{\textbf{B}} & \textcolor{jcB}{\textbf{B}} & \textcolor{jcB}{\textbf{B}} & \textcolor{jcB}{\textbf{B}} \\
17 & \textcolor{jcA}{\textbf{A}} & \textcolor{jcA}{\textbf{A}} & \textcolor{jcB}{\textbf{B}} & \textcolor{jcB}{\textbf{B}} & \textcolor{jcB}{\textbf{B}} & \textcolor{jcA}{\textbf{A}} & \textcolor{jcB}{\textbf{B}} \\
18 & \textcolor{jcA}{\textbf{A}} & \textcolor{jcB}{\textbf{B}} & \textcolor{jcA}{\textbf{A}} & \textcolor{jcB}{\textbf{B}} & \textcolor{jcA}{\textbf{A}} & \textcolor{jcA}{\textbf{A}} & \textcolor{jcA}{\textbf{A}} \\
19 & \textcolor{jcA}{\textbf{A}} & \textcolor{jcA}{\textbf{A}} & \textcolor{jcA}{\textbf{A}} & \textcolor{jcA}{\textbf{A}} & \textcolor{jcB}{\textbf{B}} & \textcolor{jcA}{\textbf{A}} & \textcolor{jcA}{\textbf{A}} \\
20 & \textcolor{jcB}{\textbf{B}} & \textcolor{jcB}{\textbf{B}} & \textcolor{jcB}{\textbf{B}} & \textcolor{jcB}{\textbf{B}} & \textcolor{jcA}{\textbf{A}} & \textcolor{jcB}{\textbf{B}} & \textcolor{jcA}{\textbf{A}} \\
21 & \textcolor{jcB}{\textbf{B}} & \textcolor{jcA}{\textbf{A}} & \textcolor{jcA}{\textbf{A}} & \textcolor{jcB}{\textbf{B}} & \textcolor{jcA}{\textbf{A}} & \textcolor{jcA}{\textbf{A}} & \textcolor{jcA}{\textbf{A}} \\
22 & \textcolor{jcB}{\textbf{B}} & \textcolor{jcB}{\textbf{B}} & \textcolor{jcB}{\textbf{B}} & \textcolor{jcB}{\textbf{B}} & \textcolor{jcB}{\textbf{B}} & \textcolor{jcB}{\textbf{B}} & \textcolor{jcB}{\textbf{B}} \\
23 & \textcolor{jcA}{\textbf{A}} & \textcolor{jcB}{\textbf{B}} & \textcolor{jcB}{\textbf{B}} & \textcolor{jcB}{\textbf{B}} & \textcolor{jcA}{\textbf{A}} & \textcolor{jcB}{\textbf{B}} & \textcolor{jcB}{\textbf{B}} \\
\bottomrule\end{tabular*}\end{minipage}
\hfill
\begin{minipage}[t]{0.485\linewidth}\vspace{0pt}\fontsize{7.5}{8.3}\selectfont
\textbf{Human raters H1--H7}\par\smallskip Human identities are anonymized; each column denotes the same person throughout.\par\smallskip
\setlength{\tabcolsep}{1pt}\renewcommand{\arraystretch}{0.95}\begin{tabular*}{\linewidth}{@{\extracolsep{\fill}}rccccccc@{}}\toprule\textbf{Case} & H1 & H2 & H3 & H4 & H5 & H6 & H7\\\midrule
1 & \textcolor{jcA}{\textbf{A}} & \textcolor{jcB}{\textbf{B}} & \textcolor{jcB}{\textbf{B}} & \textcolor{jcB}{\textbf{B}} & \textcolor{jcA}{\textbf{A}} & \textcolor{jcA}{\textbf{A}} & \textcolor{jcB}{\textbf{B}} \\
2 & \textcolor{jcA}{\textbf{A}} & \textcolor{jcA}{\textbf{A}} & \textcolor{jcA}{\textbf{A}} & \textcolor{jcA}{\textbf{A}} & \textcolor{jcA}{\textbf{A}} & \textcolor{jcA}{\textbf{A}} & \textcolor{jcA}{\textbf{A}} \\
3 & \textcolor{jcA}{\textbf{A}} & \textcolor{jcB}{\textbf{B}} & \textcolor{jcB}{\textbf{B}} & \textcolor{jcB}{\textbf{B}} & \textcolor{jcB}{\textbf{B}} & \textcolor{jcB}{\textbf{B}} & \textcolor{jcB}{\textbf{B}} \\
4 & \textcolor{jcA}{\textbf{A}} & \textcolor{jcB}{\textbf{B}} & \textcolor{jcB}{\textbf{B}} & \textcolor{jcA}{\textbf{A}} & \textcolor{jcB}{\textbf{B}} & \textcolor{jcB}{\textbf{B}} & \textcolor{jcB}{\textbf{B}} \\
5 & \textcolor{jcA}{\textbf{A}} & \textcolor{gray}{--} & \textcolor{gray}{--} & \textcolor{jcB}{\textbf{B}} & \textcolor{jcB}{\textbf{B}} & \textcolor{jcA}{\textbf{A}} & \textcolor{jcA}{\textbf{A}} \\
6 & \textcolor{jcB}{\textbf{B}} & \textcolor{gray}{--} & \textcolor{jcA}{\textbf{A}} & \textcolor{jcA}{\textbf{A}} & \textcolor{jcA}{\textbf{A}} & \textcolor{jcB}{\textbf{B}} & \textcolor{jcA}{\textbf{A}} \\
7 & \textcolor{jcB}{\textbf{B}} & \textcolor{jcA}{\textbf{A}} & \textcolor{gray}{--} & \textcolor{jcB}{\textbf{B}} & \textcolor{jcB}{\textbf{B}} & \textcolor{jcA}{\textbf{A}} & \textcolor{jcA}{\textbf{A}} \\
8 & \textcolor{jcA}{\textbf{A}} & \textcolor{gray}{--} & \textcolor{gray}{--} & \textcolor{jcA}{\textbf{A}} & \textcolor{jcA}{\textbf{A}} & \textcolor{jcB}{\textbf{B}} & \textcolor{jcA}{\textbf{A}} \\
9 & \textcolor{jcA}{\textbf{A}} & \textcolor{jcB}{\textbf{B}} & \textcolor{jcB}{\textbf{B}} & \textcolor{jcA}{\textbf{A}} & \textcolor{jcB}{\textbf{B}} & \textcolor{jcB}{\textbf{B}} & \textcolor{jcA}{\textbf{A}} \\
10 & \textcolor{jcB}{\textbf{B}} & \textcolor{jcA}{\textbf{A}} & \textcolor{jcA}{\textbf{A}} & \textcolor{jcA}{\textbf{A}} & \textcolor{jcA}{\textbf{A}} & \textcolor{jcA}{\textbf{A}} & \textcolor{jcA}{\textbf{A}} \\
11 & \textcolor{jcB}{\textbf{B}} & \textcolor{gray}{--} & \textcolor{jcB}{\textbf{B}} & \textcolor{jcB}{\textbf{B}} & \textcolor{jcB}{\textbf{B}} & \textcolor{jcB}{\textbf{B}} & \textcolor{jcB}{\textbf{B}} \\
12 & \textcolor{jcA}{\textbf{A}} & \textcolor{jcB}{\textbf{B}} & \textcolor{jcB}{\textbf{B}} & \textcolor{jcB}{\textbf{B}} & \textcolor{jcB}{\textbf{B}} & \textcolor{jcA}{\textbf{A}} & \textcolor{jcA}{\textbf{A}} \\
13 & \textcolor{jcB}{\textbf{B}} & \textcolor{gray}{--} & \textcolor{jcA}{\textbf{A}} & \textcolor{jcB}{\textbf{B}} & \textcolor{jcB}{\textbf{B}} & \textcolor{jcA}{\textbf{A}} & \textcolor{jcB}{\textbf{B}} \\
14 & \textcolor{jcB}{\textbf{B}} & \textcolor{gray}{--} & \textcolor{jcB}{\textbf{B}} & \textcolor{jcB}{\textbf{B}} & \textcolor{gray}{--} & \textcolor{jcB}{\textbf{B}} & \textcolor{jcB}{\textbf{B}} \\
15 & \textcolor{jcB}{\textbf{B}} & \textcolor{gray}{--} & \textcolor{jcA}{\textbf{A}} & \textcolor{jcB}{\textbf{B}} & \textcolor{jcB}{\textbf{B}} & \textcolor{jcB}{\textbf{B}} & \textcolor{jcB}{\textbf{B}} \\
16 & \textcolor{jcB}{\textbf{B}} & \textcolor{gray}{--} & \textcolor{jcA}{\textbf{A}} & \textcolor{jcB}{\textbf{B}} & \textcolor{jcA}{\textbf{A}} & \textcolor{jcB}{\textbf{B}} & \textcolor{jcB}{\textbf{B}} \\
17 & \textcolor{jcA}{\textbf{A}} & \textcolor{gray}{--} & \textcolor{jcA}{\textbf{A}} & \textcolor{jcB}{\textbf{B}} & \textcolor{jcA}{\textbf{A}} & \textcolor{jcA}{\textbf{A}} & \textcolor{jcA}{\textbf{A}} \\
18 & \textcolor{jcB}{\textbf{B}} & \textcolor{jcB}{\textbf{B}} & \textcolor{jcA}{\textbf{A}} & \textcolor{jcB}{\textbf{B}} & \textcolor{jcB}{\textbf{B}} & \textcolor{jcB}{\textbf{B}} & \textcolor{jcB}{\textbf{B}} \\
19 & \textcolor{jcB}{\textbf{B}} & \textcolor{gray}{--} & \textcolor{jcA}{\textbf{A}} & \textcolor{jcA}{\textbf{A}} & \textcolor{jcB}{\textbf{B}} & \textcolor{jcA}{\textbf{A}} & \textcolor{jcA}{\textbf{A}} \\
20 & \textcolor{jcB}{\textbf{B}} & \textcolor{jcA}{\textbf{A}} & \textcolor{jcB}{\textbf{B}} & \textcolor{gray}{--} & \textcolor{jcB}{\textbf{B}} & \textcolor{jcB}{\textbf{B}} & \textcolor{jcB}{\textbf{B}} \\
21 & \textcolor{jcA}{\textbf{A}} & \textcolor{jcA}{\textbf{A}} & \textcolor{jcB}{\textbf{B}} & \textcolor{jcB}{\textbf{B}} & \textcolor{jcB}{\textbf{B}} & \textcolor{jcB}{\textbf{B}} & \textcolor{jcB}{\textbf{B}} \\
22 & \textcolor{jcB}{\textbf{B}} & \textcolor{jcB}{\textbf{B}} & \textcolor{jcB}{\textbf{B}} & \textcolor{jcB}{\textbf{B}} & \textcolor{jcB}{\textbf{B}} & \textcolor{jcB}{\textbf{B}} & \textcolor{jcB}{\textbf{B}} \\
23 & \textcolor{jcA}{\textbf{A}} & \textcolor{jcB}{\textbf{B}} & \textcolor{jcB}{\textbf{B}} & \textcolor{jcB}{\textbf{B}} & \textcolor{jcA}{\textbf{A}} & \textcolor{jcB}{\textbf{B}} & \textcolor{jcB}{\textbf{B}} \\
\bottomrule\end{tabular*}\end{minipage}
\par\bigskip
\endgroup

\end{document}